\documentclass[preprint]{elsarticle}      % preprint is be used for submission
\usepackage{amsmath, amssymb, upgreek}    % math stuff
\usepackage{graphicx}                     % for including images
\usepackage{caption, subcaption}          % needed for sub-figures
\usepackage{listings, booktabs, tabularx} % code and tables
\usepackage[colorlinks=true,              % don't have the box around a link
            citecolor=black,              % the number in cite, e.g. [27]
            linkcolor=black,              % the num in ref, eg. Fig. 1b
            urlcolor=black]{hyperref}     % and urls are also
\usepackage[english]{babel}               % hyphenation
\usepackage{natbib}                       % citations/bibliography

\newif\iftrack
\trackfalse % uncomment this to hide comments and deletions

\iftrack
 \usepackage[normalem]{ulem}
 \usepackage{color}
 \newcommand{\dummycite}[1]{[?]}
 \newcommand{\removecite}{\renewcommand\cite\dummycite
                          \renewcommand\citep\dummycite
                          \renewcommand\citet\dummycite}
 
 \newcommand{\deleted}[2]{{\marginpar{\small\color{red}#1}}
                          {\removecite\color{red}\sout{#2}}}
 
 \newcommand{\comment}[2]{{\color{green}
                          \{{\footnote{\color{green}{}#1: #2}}\}}}
\else
 
 \newcommand{\deleted}[2]{}
 
 \newcommand{\comment}[2]{}
\fi

\DeclareMathOperator{\LEFT}{\textsc{left}}
\DeclareMathOperator{\RIGHT}{\textsc{right}}

\DeclareMathOperator{\EXISTS}{{\exists}}

\DeclareMathOperator{\EXIST}{{\exists}}
\DeclareMathOperator{\EXACTLY}{{\#}}
\DeclareMathOperator{\MORE}{{>}}
\DeclareMathOperator{\EQUALNUM}{{=}}

\DeclareMathOperator{\GREATER}{{>}}

\DeclareMathOperator{\GREATERLA}{{>}}
\DeclareMathOperator{\GREATERLLA}{{>}}

\DeclareMathOperator{\EQUAL}{{=}}

\DeclareMathOperator{\MORESIMLA}{{\succ}}
\DeclareMathOperator{\MORESIMLLA}{{\succ}}

\DeclareMathOperator{\CAP}{{\cap}}
\DeclareMathOperator{\CUP}{{\cup}}

\DeclareMathOperator{\SETMINUS}{{\setminus}}

\DeclareMathOperator{\INSIDE}{\textsc{inside}}
\DeclareMathOperator{\CONTAINS}{\textsc{contains}}
\DeclareMathOperator{\ALIGNED}{\textsc{aligned}}

\DeclareMathOperator{\GET}{\textsc{get}}
\DeclareMathOperator{\HIGH}{\textsc{high}}
\DeclareMathOperator{\LOW}{\textsc{low}}

\DeclareMathOperator{\SOLID}{\textsc{solid}}

\DeclareMathOperator{\OUTLINE}{\textsc{outline}}
\DeclareMathOperator{\BIG}{\textsc{large}} %because \LARGE is a latex command
\DeclareMathOperator{\SMALL}{\textsc{small}}

\DeclareMathOperator{\FIGURES}{\textsc{figures}}
\DeclareMathOperator{\OBJECTS}{\textsc{objects}}
\DeclareMathOperator{\CIRCLES}{\textsc{circles}}
\DeclareMathOperator{\TRIANGLES}{\textsc{triangles}}
\DeclareMathOperator{\RECTANGLES}{\textsc{rectangles}}

\DeclareMathOperator{\XPOS}{\textsc{xpos}}
\DeclareMathOperator{\YPOS}{\textsc{ypos}}
\DeclareMathOperator{\ORIENTATION}{\textsc{orientation}}
\DeclareMathOperator{\DISTANCE}{\textsc{distance}}

\DeclareMathOperator{\SIZE}{\textsc{size}}
\DeclareMathOperator{\CONVEXITY}{\textsc{convexity}}
\DeclareMathOperator{\COMPACTNESS}{\textsc{compactness}}
\DeclareMathOperator{\ELONGATION}{\textsc{elongation}}

\DeclareMathOperator{\NCORNERS}{\textsc{ncorners}}

\DeclareMathOperator{\COLOR}{\textsc{color}}

\DeclareMathOperator{\HULLS}{\textsc{hulls}}
\DeclareMathOperator{\HOLES}{\textsc{holes}}

\begin{document}
\title{Solving Bongard Problems with a Visual Language and Pragmatic Reasoning}

\author[uos]{Stefan Depeweg}
\ead{stdepewe@uos.de}
\author[tuda,fias]{Constantin A. Rothkopf}
\ead{rothkopf@psychologie.tu-darmstadt.de}
\author[tuda]{Frank J\"akel}
\ead{jaekel@psychologie.tu-darmstadt.de}

\address[uos]{Institute of Cognitive Science,
              Universit\"at Osnabr\"uck, Germany}
\address[tuda]{Centre for Cognitive Science and Institute of Psychology,
               TU Darmstadt, Germany}
\address[fias]{Frankfurt Institute for Advanced Studies,
               Germany}


\begin{abstract}

More than 50 years ago Bongard introduced 100 visual concept learning problems
as a testbed for intelligent vision systems. These problems are now known as
Bongard problems. Although they are well known in the cognitive science and AI
communities only moderate progress has been made towards building systems that
can solve a substantial subset of them. In the system presented here, visual
features are extracted through image processing and then translated into a
symbolic visual vocabulary. We introduce a formal language that allows
representing complex visual concepts based on this vocabulary. Using this
language and Bayesian inference, complex visual concepts can be induced from the
examples that are provided in each Bongard problem. Contrary to other concept
learning problems the examples from which concepts are induced are not random in
Bongard problems, instead they are carefully chosen to communicate the concept,
hence requiring pragmatic reasoning. Taking pragmatic reasoning into account we
find good agreement between the concepts with high posterior probability and the
solutions formulated by Bongard himself. While this approach is far from solving
all Bongard problems, it solves the biggest fraction yet.

\end{abstract}

\maketitle

%++++++++++++++++++++++++++++++++++++++++++++++++++++++++++++++++++++++++++++++
\section{Introduction}
%++++++++++++++++++++++++++++++++++++++++++++++++++++++++++++++++++++++++++++++

In the last few years computer vision has made tremendous progress on object
recognition and answering the question ``what is where'' in an image
\cite{Marr_1982}. With recent advances in deep learning and big data, robust
solutions finally seem within grasp \cite{Karpathy_etal_2014,lecun2015deep}.
Some even claim that human performance on object recognition tasks has already
been surpassed by machines \cite{he2015delving}.

However, while knowing what is where is a good start, artificial intelligence
still has a long way to go before it can compete with human visual cognition
\cite{hernandez2016computer}. Humans still outperform computers in their ability
to reason about the visual world. They can separate relevant from irrelevant
visual information, relate visual observations to background knowledge, infer
invisible properties, generalize from very little data, interpret novel visual
stimuli, predict what will happen next in a scene, and act accordingly. In order
to mimic these human abilities, artificial vision systems will need to interact
with higher reasoning systems.

In cognitive science there is a long-standing debate about the interaction of
vision and cognition \cite{Pylyshyn_1999b}. There are some aspects of ``visual
cognition'' that are clearly purely visual: they can be conceptualized as image
processing only. Filters in early vision are a prime example
\cite{Campbell_Robson_1968}. At the other end of the spectrum, solving Raven's
progressive matrices
\cite{Carpenter_etal_1990,Strannegard_etal_2013,Kunda_etal_2013}, a popular IQ
test item, trivially requires the visual system because the input is a visual
stimulus. Other than that, however, the task engages general problem solving
mechanisms that, presumably, are independent of vision. These two examples
suggest that the visual system operates on sub-symbolic mechanisms, like filters
and neural networks, whereas reasoning about visual scenes is better described
by a separate symbolic system \cite[but see][]{Smolensky_1988}.

However, there are also examples where it is much harder to tell which side of
the visual/cognitive boundary they are on. Take amodal completion. A square that
is covered by a circle is still perceived as a square, even though the visual
image does not actually contain a square. Does amodal completion invoke
symbolic, non-visual knowledge of squares? Or take the simple concept of a
triangle. There is obviously a strong visual component. But there is also a
symbolic component---a logical definition---that depends on more basic visual
concepts like edges, lines, angles, and corners: A \emph{tri-angle} has three
angles. Such phenomena seem to suggest that there is a tight interaction between
the visual and the conceptual system. Over the years, some cognitive scientists
have insisted that many visual phenomena also show these more cognitive, often
language- or logic-like aspects
\cite[e.g.][]{Biederman_1987,Rock_1983,Savova_etal_2009}. But if the early
visual system is sub-symbolic and the conceptual system is symbolic, how exactly
do these two systems interact in visual cognition and what is their interface?

%Also in the history of artificial intelligence there has always been a tension
%between visual signal processing and more abstract visual representations. After
%the first successes of perceptrons \cite{Rosenblatt_1958} skepticism culminated
%in the famous body of work by Minsky and Papert \cite{Minsky_Papert_1967}. Their
%criticism had a strong logical flavor and the counterexamples are abstract
%visual concepts, like convexity and connectedness (see also \cite{Ullman_1984}).
%With the great success of modern machine learning in the perceptron tradition,
%some recent work continues to caution that many interesting visual concepts,
%like sameness or insideness, can only be learned by heart with many of the most
%popular techniques, e.g. support vector machines \cite{Fleuret_etal_2011}.

Perhaps this question is a red herring. There is, of course, the possibility
that vision and cognition are just two ends of a continuum without a clearcut
boundary between them. Perhaps the same neural mechanisms are applied
hierarchically to achieve ever greater levels of abstraction and there is no
meaningful distinction between vision and cognition
\cite[e.g.][]{Konig_etal_2013}. Cognition is just deeper in the brain. Instead
of two separate modules---one for vision and one for cognition, each operating
on different principles---there might just be one deep, probably recurrent,
neural network.

Although the recent successes of deep learning in computer vision are
impressive, skepticism as to whether deep learning will be able capture higher
aspects of human visual cognition is warranted: For example, the way that neural
networks generalize in computer vision problems
\cite{zeiler2014visualizing,nguyen2015deep,sharif2016accessorize} can be very
different from intuitive human generalizations, although both systems,
artificial neural networks and humans, show a similarly high performance on
benchmarks \cite{szegedy2013intriguing}. Even when only performance is
considered as a criterion, differences between deep networks and humans can go
both ways, in that humans but not deep nets systematically miss very large
objects in visual scenes \cite{eckstein2017humans}, while humans are much more
robust at object detection with respect to a variety of different noise
manipulations \cite{geirhos2017comparing}. Also, a recent series of studies has
focused on demonstrating that human performance in classifying shapes, which are
generated according to a set of rules, are indeed better captured by symbolic
shape inference systems instead of systems utilizing features derived from deep
convolutional networks
\cite{erdogan2015sensory,erdogan2017visual,Overlan_etal_2017,Ellis_etal_2015}.

Hence, it seems that is important to further improve deep neural networks but,
at the same time, to also explore possible ways for sub-symbolic visual systems
to interact with more classic symbolic techniques from artificial intelligence.
Historically, syntactic pattern recognition
\cite{Fu_1974,Gonzalez_Thomason_1978} was an attempt to bring symbolic
techniques from language and logic, i.e. higher cognition, to bear on the
problems of vision. Early attempts were very successful on toy problems
\cite[e.g.][]{Winston_1970} but were unsatisfactory since it was unclear whether
the same techniques could ever be applied to real-world images. Given the
progress that was made on image processing and probabilistic modeling since the
seventies, modern syntactic approaches to visual object recognition cannot be
said to suffer from these restrictions anymore. Stochastic grammars and
graphical models can capture the combinatorial variations and the internal
structure of objects, and probability theory allows us to combine these
high-level representations with noisy low-level image features in a principled
manner \cite[see e.g.][]{Lin_etal_2009,Zhu_Mumford_2006,Lake_etal_2015}.

With object recognition turning into a mature technology, we think the focus of
research in computer vision will increasingly shift to scene understanding with
new benchmark datasets, e.g. for visual reasoning \cite{johnson2017clevr}, and
new criteria for success. We expect vision systems to understand images and
concepts that they were not specifically trained for \cite{Lampert_etal_2014}
and we would like these AI systems to answer natural language questions about
visual scenes, e.g. about the number of drawers of the leftmost filing cabinet
in a picture \cite{Malinowski_Fritz_2014}. Combining computer vision systems,
natural language processing, semantic knowledge-bases, and reasoning systems for
such a visual Turing test starts to become feasible. As the necessary tools have
been developed mostly independently so far, their integration and the
specification of interfaces will become a more prominent topic. In particular,
we will need to specify representations that are sufficiently expressive to
capture the great variations in visual scenes but also allow for efficient
reasoning. In short, the central question in this line of research will be: What
is the language of vision?

\section{Bongard problems}

\begin{figure}[!t]

  \begin{subfigure}[t]{0.425\textwidth}
    \centering
    \includegraphics[width=1\textwidth]{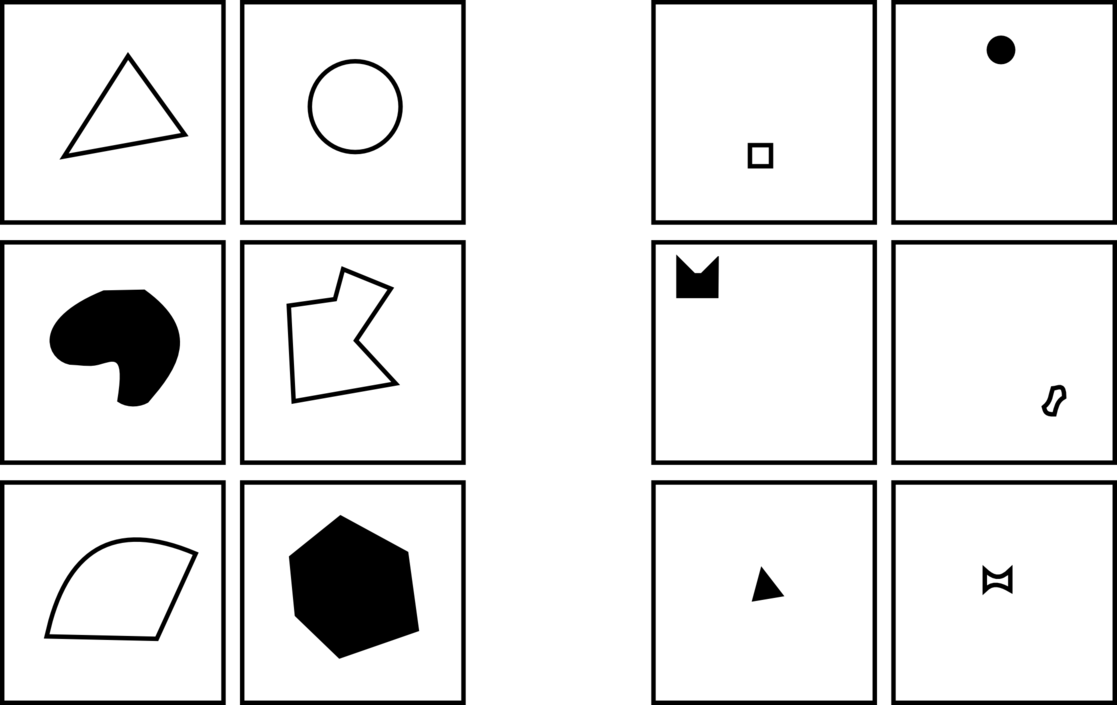}
    \caption{BP \#2. Large figures $\mid$
      Small figures.}
    \label{exbp2}
  \end{subfigure}
  \hfill
  \begin{subfigure}[t]{0.425\textwidth}
    \centering
    \includegraphics[width=1\textwidth]{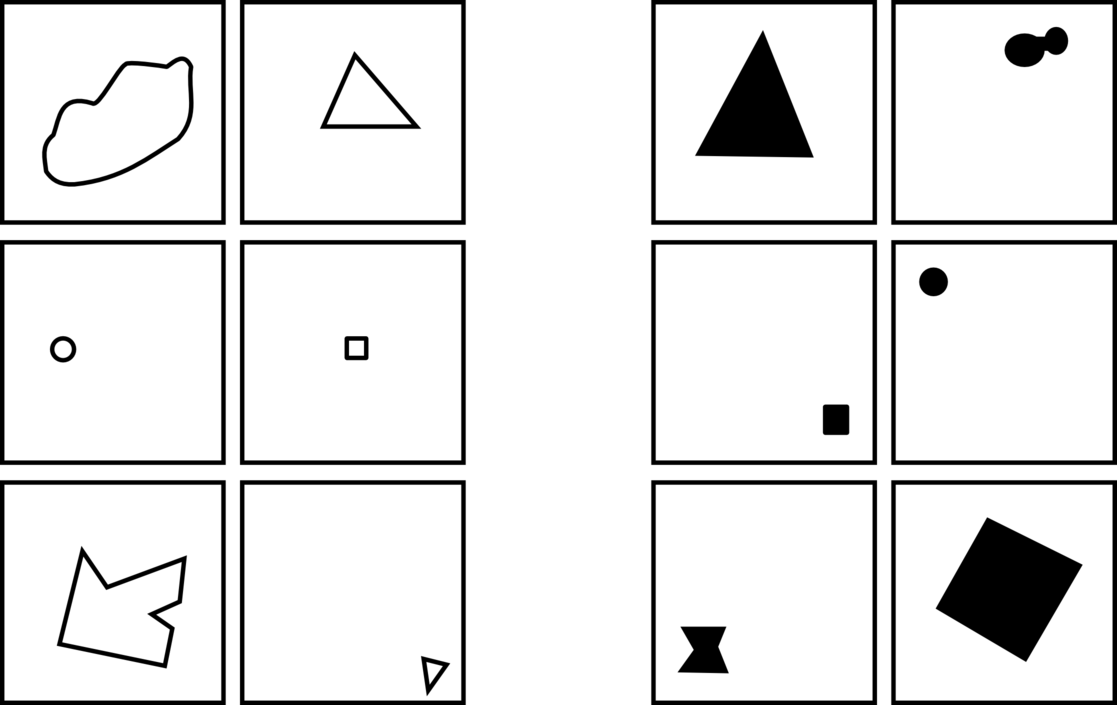}
    \caption{BP \#3. Outline figures $\mid$ Solid
      figures.}
    \label{exbp3}
  \end{subfigure}

  \medskip

  \begin{subfigure}[t]{0.425\textwidth}
    \centering
    \includegraphics[width=1\textwidth]{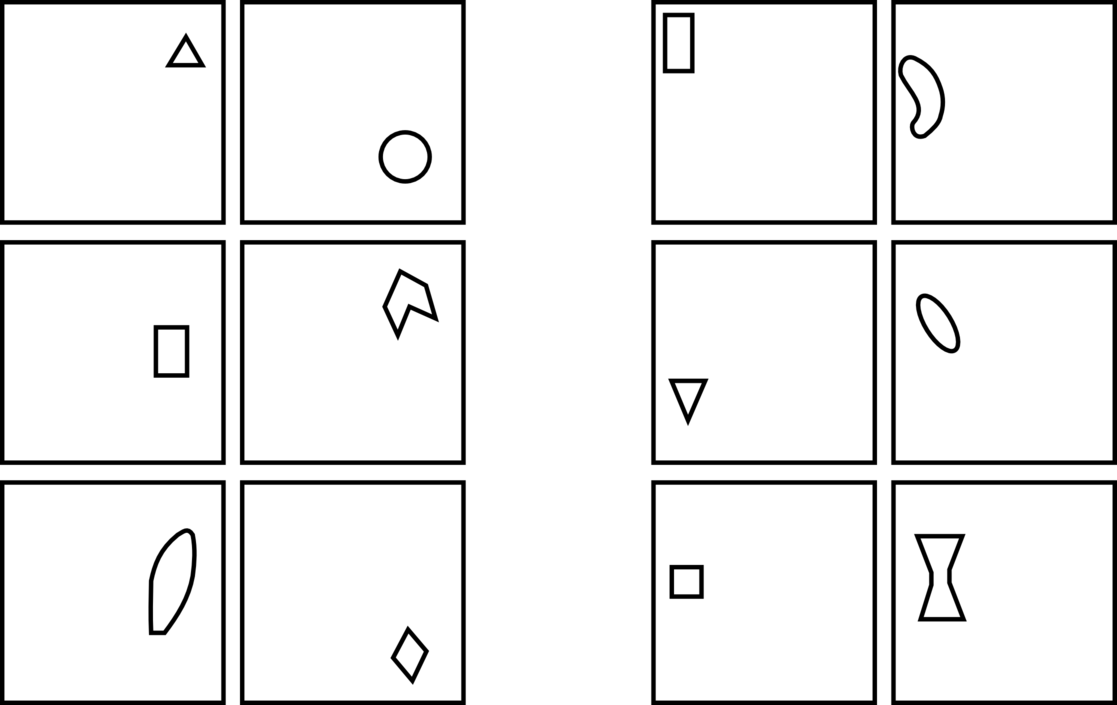}
    \caption{BP \#8. Figures on the right side $\mid$
      Figures on the left side.}
    \label{exbp8}
  \end{subfigure}
  \hfill
  \begin{subfigure}[t]{0.425\textwidth}
    \centering
    \includegraphics[width=1\textwidth]{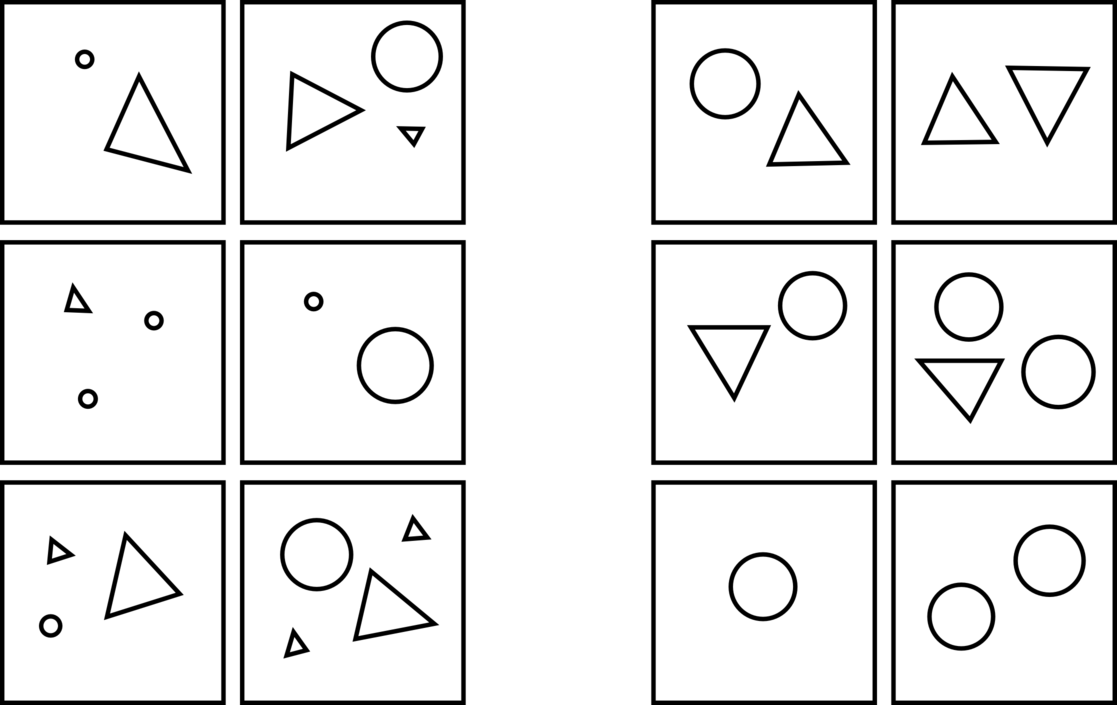}
    \caption{BP \#21. Small figure present $\mid$
      No small figure present.}
    \label{exbp21}
  \end{subfigure}

  \medskip

  \begin{subfigure}[t]{0.425\textwidth}
    \centering
    \includegraphics[width=1\textwidth]{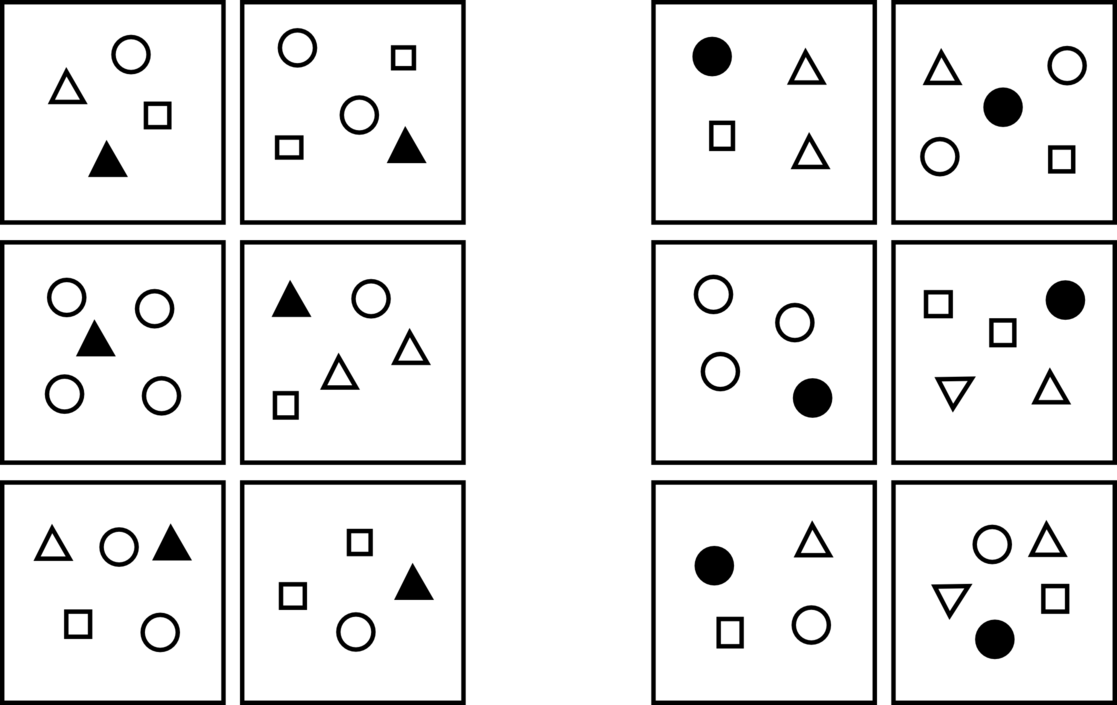}
    \caption{BP \#25. Black figure is a triangle $\mid$
      Black figure is a circle.}
    \label{exbp25}
  \end{subfigure}
  \hfill
  \begin{subfigure}[t]{0.425\textwidth}
    \centering
    \includegraphics[width=1\textwidth]{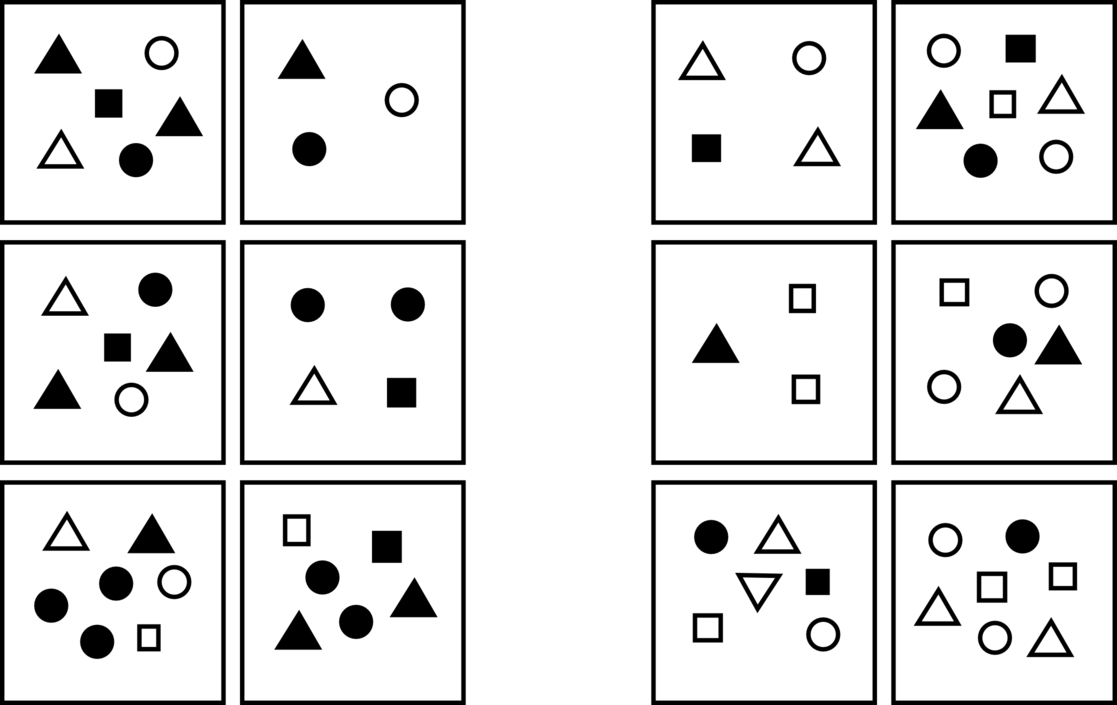}
    \caption{BP \#27. More solid black figures $\mid$
      More outline figures.}
    \label{exbp27}
  \end{subfigure}

  \medskip

  \begin{subfigure}[t]{0.425\textwidth}
    \centering
    \includegraphics[width=1\textwidth]{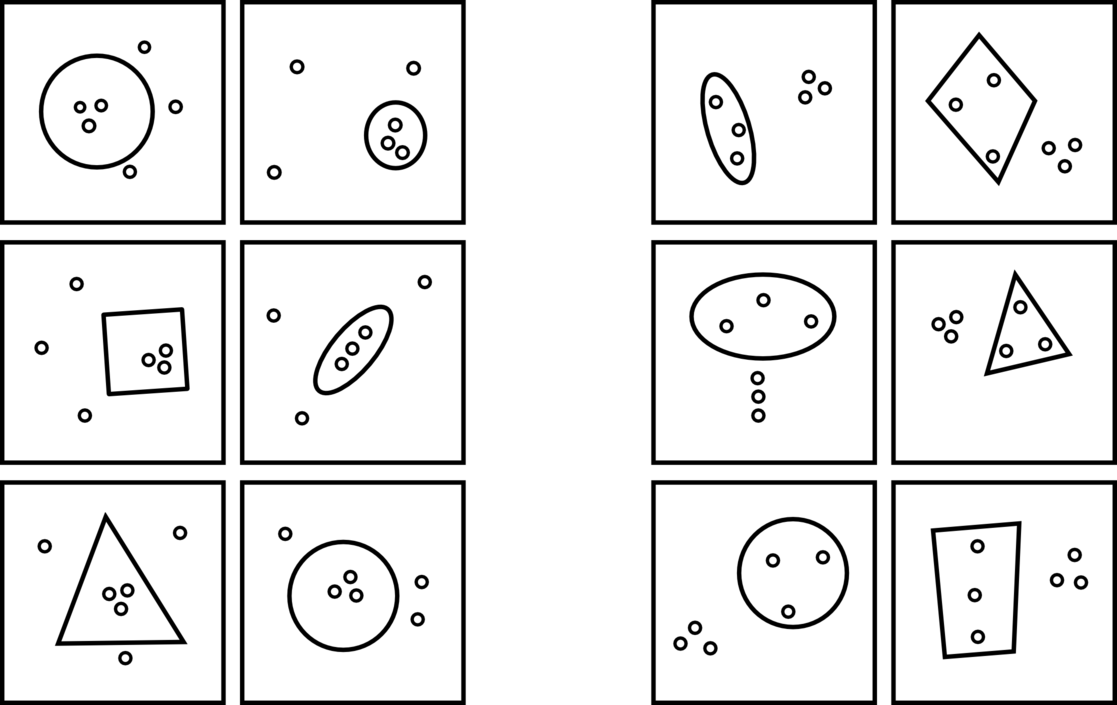}
    \caption{BP \#49. Points inside are grouped more
      densely than outside $\mid$ Points outside are grouped more
      densely than inside.}
    \label{exbp49}
  \end{subfigure}
  \hfill
  \begin{subfigure}[t]{0.425\textwidth}
    \centering
    \includegraphics[width=1\textwidth]{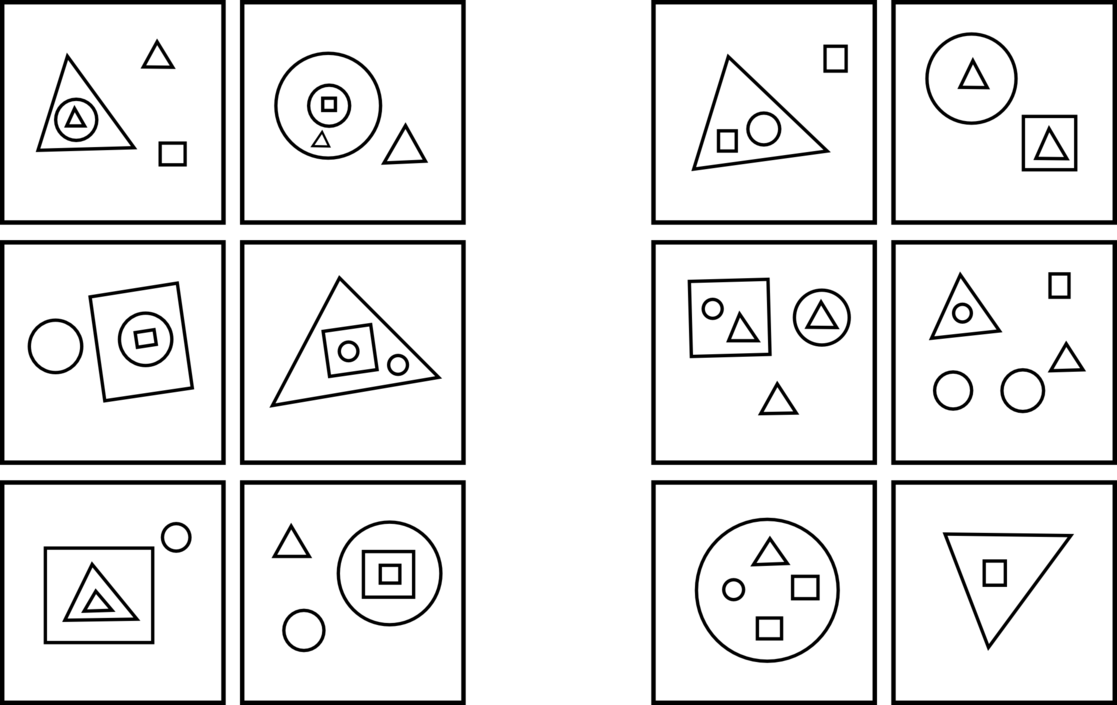}
    \caption{BP \#71. There are inside figures of the
      second order $\mid$ There are no inside figures of the second
      order.}
    \label{exbp71}
  \end{subfigure}

  \caption{A Bongard problem (BP) consists of two sets of six binary
    images. The six images on the left illustrate a visual concept and
    the six images on the right a related visual concept. The task of
    the vision system is to describe the concept based on the
    examples.}

\end{figure}

One of the first researchers to ask this question was Mikhail Moiseevich Bongard
\cite{Bongard_1970}. He introduced the now so-called \emph{Bongard problems}
(BPs), that were later popularized by Hofstadter \cite[ch. 19]{Hofstadter_1977},
in order to demonstrate the inadequacy of the standard pattern recognition tools
of the day for human-level visual cognition \cite[see
also][]{Minsky_Papert_1967,Fleuret_etal_2011}. A simple Bongard problem is shown
in Fig.~\ref{exbp3}. Each Bongard problem consists of a set of twelve example
images. The images are geometric, binary images. They are divided into two
classes. The first class is given by the six images on the left while the second
class is given by the six examples on the right. The task for a vision system is
to find two related concepts that can discriminate between the left and the
right side. For Fig.~\ref{exbp3} the solution that Bongard gives is: ``Outline
figures $\mid$ Solid figures'' \cite[p. 247]{Bongard_1970}. For each Bongard
problem the solution is a logical rule given in natural language that refers to
visual features seen in the examples.

%.....................................................................
%.....................................................................

Bongard problems come in a huge variety and with varying difficulty. Some of the
problems are like IQ test items \cite[cf.][]{hernandez2016computer} and are
relatively hard to solve but many are rather easy for the human visual-cognitive
system. More examples can be seen in Fig.~\ref{exbp2} to Fig.
\ref{exbp71}.\footnote{The original Bongard problems and additional, similar
problems can be found on Harry Foundalis' webpage:
\url{http://www.foundalis.com/res/bps/bpidx.htm}.} The challenge is to create a
vision system that can solve all Bongard problems without using ad-hoc solutions
for each problem. Bongard carefully hand-crafted these problems to illustrate
the challenges that an intelligent vision system needs to overcome. His
problems can therefore be used as a guidance to explore representations for
vision systems \cite{Montalvo_1985}. For example, Bongard problem \#71 (Fig.
\ref{exbp71}) illustrates the recursive use of the inside relation to define a
concept. This use of recursion, like many of the other Bongard problems, again
calls for language-like visual representations
\cite{Bongard_1970,Savova_etal_2009,Savova_Tenenbaum_2008}.

Although Bongard problems are well known among AI researchers and cognitive
scientists and are standardly used as demonstrations in inductive logic
programming \cite[e.g.][]{DeRaedt_2010}, few systems have been built that
systematically try and solve a significant subset of them (in contrast to, e.g.,
Raven's progressive matrices
\cite{Carpenter_etal_1990,Strannegard_etal_2013,Kunda_etal_2013}). RF4
\cite{Saito_Nakano_1996} is an inductive logic programming system that can solve
41 of the 100 Bongard problems, but only because the researchers hand-coded the
images into logical formulas. Thus, the computer vision side of the problem is
completely avoided in this approach. Phaeaco, in contrast, \cite{Foundalis_2006}
uses images as inputs, but its expressivity is rather limited \cite[ch.
11.2]{Foundalis_2006}. Of the original 100 Bongard problems it can only reliably
solve around 10.

Here we present a new method for solving Bongard problems that starts with a set
of image processing functions that extract basic shapes and visual properties.
Using this visual vocabulary we define a formal language, a context-free
grammar, that allows the expression of complex visual concepts. Using Bayesian
inference on the space of concepts we search for sentences in this language that
best describe a given Bongard problem. While this basic idea corresponds to the
intuitions already formulated by Bongard \cite[ch. 9]{Bongard_1970}, progress
towards implementing a working system has been very slow so far.
%Our computational system improves on previous approaches by providing unambiguous
%solution to 35 of the 39 Bongard problems that we considered.

\subsection{Preliminary Observations}

Before we start explaining our system, a few preliminary observations are in
order. First, automated solvers for Bongard problems should not fail because of
problems in image processing. Bongard problems are just black and white line
drawings and they are a lot easier to process than natural images. If the image
is of sufficient quality and resolution, standard computer vision tools for
segmentation and feature extraction will work extremely well. While humans can
certainly deal with low quality images, image processing problems are not our
central concern here. Hence, we have created a clean subset of Bongard problems
with high resolution that makes the image processing side easy \cite[in contrast
to][]{Foundalis_2006}.\footnote{These images are made available together with
the publication as supplementary material.}

\begin{figure}[t]
  \begin{subfigure}[t]{0.45\textwidth}
    \centering
    \includegraphics[width=1\textwidth]{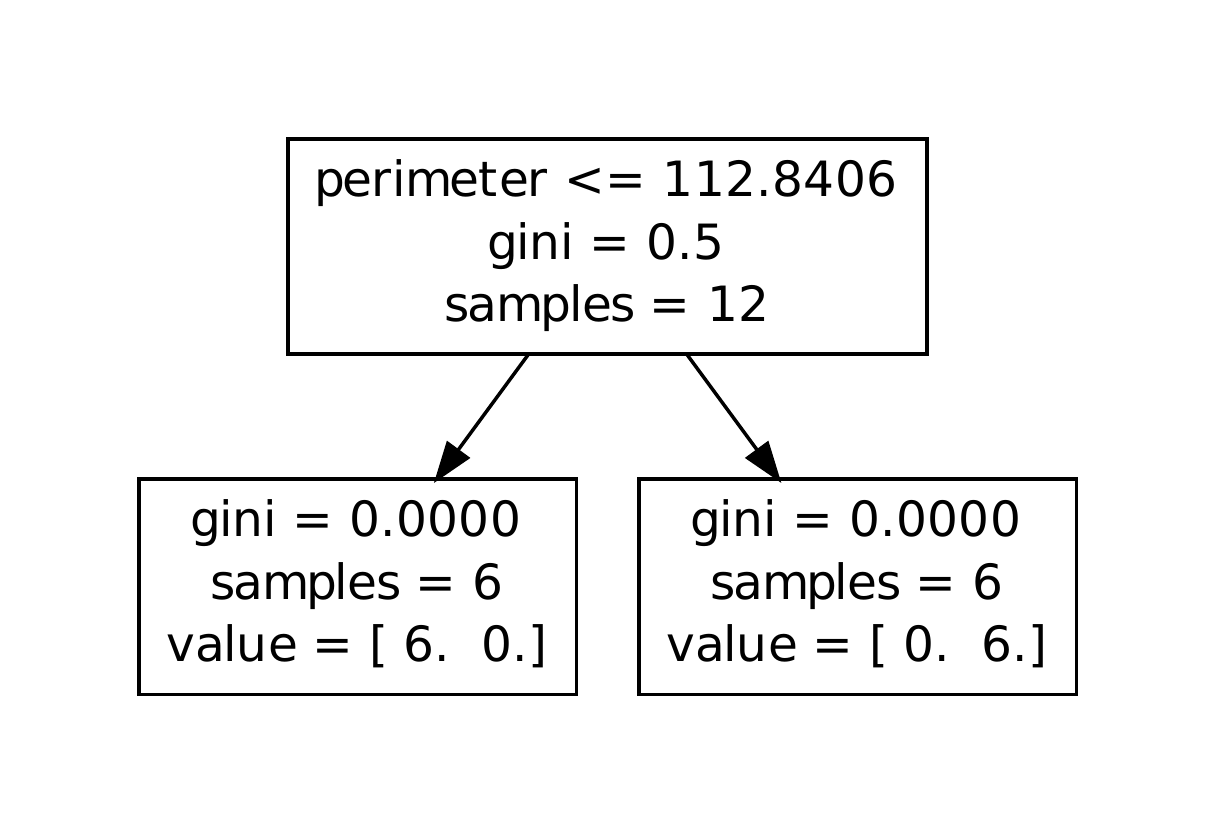}
    \caption{Bongard problem \#02. Large figures $\mid$ Small figures
             (see Fig.~\ref{exbp2}).}
    \label{dtbp02}
  \end{subfigure}
  \hfill
  \begin{subfigure}[t]{0.45\textwidth}
    \centering
    \includegraphics[width=1\textwidth]{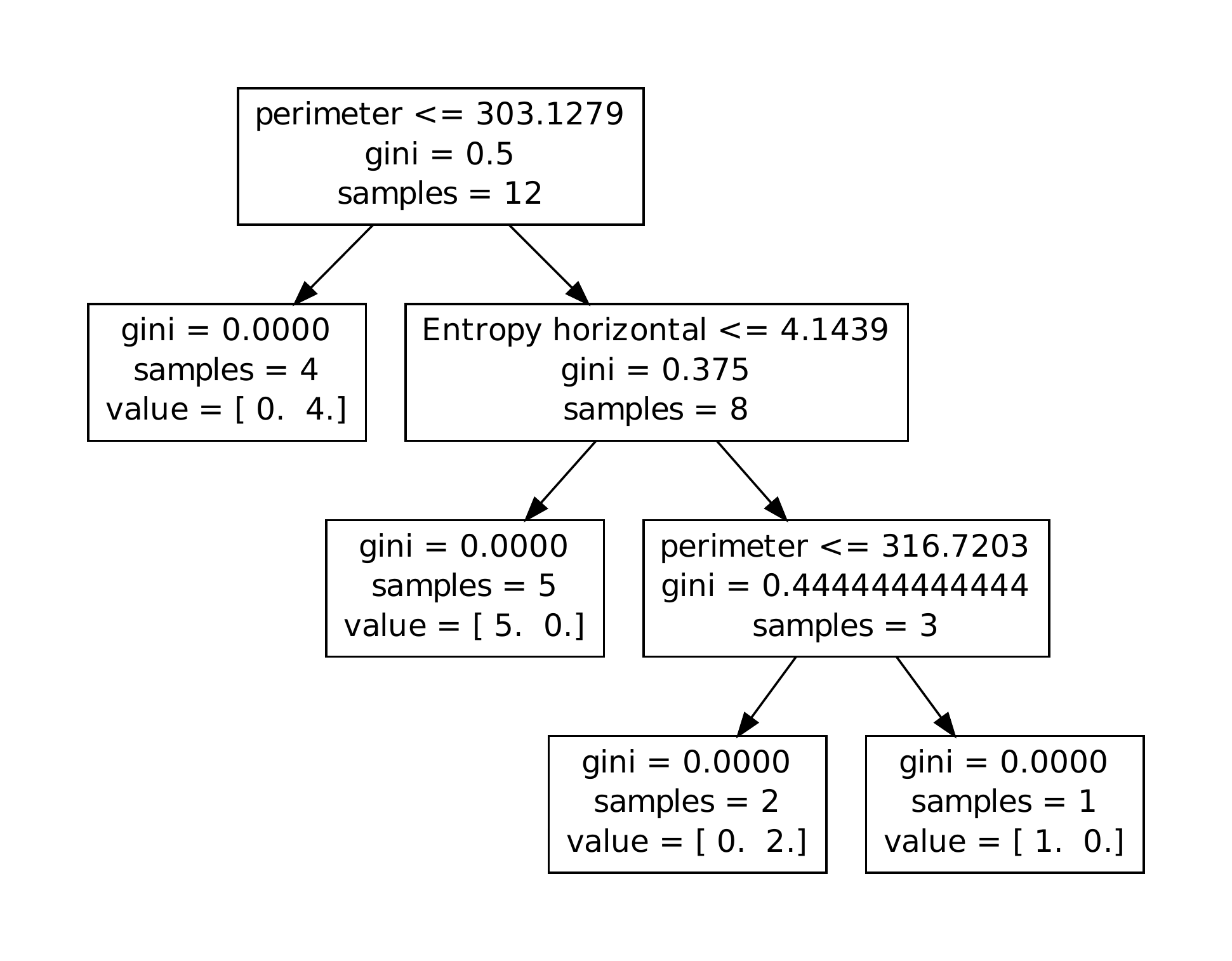}
    \caption{Bongard problem \#71. There are inside figures of the
      second order $\mid$ There are no inside figures of the second
      order (see Fig.~\ref{exbp71}).}
    \label{dtbp71}
  \end{subfigure}

  \caption{Applying decision trees to the Bongard problems. Two examples of
  using decision trees to separate the two visual concepts. Left: The two
  concepts are distinguished based on the perimeter. Right: Successive
  applications of decisions based on incidental feature differences allow
  separating the left and right side images but does not correspond to the
  intended visual concept.}

\end{figure}

Second, Bongard problems are not easily solved by standard classification
methods. In fact, they were designed to illustrate how standard methods fell
short at the time. And they still do. The standard approach would be to extract
features from the images and then apply a classifier to the 12 examples. To make
this discussion more concrete we have extracted a number of standard computer
vision features from the example images (like the position, area, or perimeter
of the figures) and applied a standard decision tree algorithm (with the Gini
index as criterion) to separate the two classes. In some cases this seemed to
work well, e.g. for BP \#2 (Fig.~\ref{exbp2}) that merely distinguishes between
small and large objects. Fig.~\ref{dtbp02} shows that the decision tree
algorithm distinguishes the objects on the left and on the right based on their
perimeter. However, for more complicated Bongard problems the classifier usually
overfits the small number of samples. This is illustrated for BP \#71 (see Fig.
\ref{exbp71}) in Fig.~\ref{dtbp71}. The learning algorithm capitalizes on
incidental small differences in perimeter between the few examples in the two
classes. In this way the algorithm can easily separate the two classes but does
not learn the intended concept (``inside figures of the second order'').

A common method to avoid overfitting is cross-validation. However, Bongard
problems are not like your off-the-shelf classification problem. There is only a
very small number of examples and, what is worse, these examples are not random
samples. For the most part Bongard carefully constructed the examples in order
to convey the concept he had in mind. For example, in BP \#71
(Fig.~\ref{exbp71}) the shapes and number of objects are purposely varied to
rule out more specific hypotheses. Sometimes leaving out just one of the
examples can even make the intended concept ambiguous. For example, in BP \#21
(Fig.~\ref{exbp21}), leaving out the example where all figures are small might
suggest that there also have to exist large figures or that the variability in
size between the figures is important. Bongard problems are therefore best
understood as communication problems: Bongard is trying to communicate a concept
through carefully chosen examples. Hence, solving a Bongard problem is not a
problem of generalizing well from random samples but a problem of inferring the
intended concept in a communicative situation \cite[as
in][]{Shafto_Goodman_2008}.

In order to infer the intended concept, as a first step, we need to understand
the space of hypotheses that Bongard (or any other human) entertained. Which
visual concepts are ``natural'' enough to be considered? Even though Bongard
problems come in a huge variety, there are some recurring themes. For example,
some basic shapes, like circles, triangles, and squares occur in many problems.
Some shape properties, like size, position, orientation, and color (i.e. outline
or black) also frequently play a role. Very often, the relations between objects
are crucial, as in BP \#71 (Fig.~\ref{exbp71}) that hinges on the inside
relation. Our aim was not to build a general vision system, instead we decided
to focus on these visual shapes, properties and relations that appear in many of
the Bongard problems and hence seem to be likely candidates for a natural
vocabulary from which relevant visual concepts can be built.

For example, in BP \#25 (Fig.~\ref{exbp25}) figures have to have a certain shape
and a certain color: black triangles vs. black circles. Hence, a language to
express visual concepts will at the very least need logical operators to combine
basic visual properties and shapes. We will also need some form of
quantification as can clearly be seen in  BP \#1 (Fig.~\ref{exbp21}): ``Small
figure present'' vs. ``No small figure present''. Apart from such standard
ingredients of logic, there are other recurring patterns that we will need to
express. For BP \#27 (Fig.~\ref{exbp27}), for instance, we need to be able to
count and state that there are more solid than outlined figures in each image.

Last but not least, all solutions have the form of stating one concept for the
images on the left and one concept for the images on the right. The left-concept
has to apply to all images on the left and must not be true of any of the images
on the right, and conversely for the right-concept. Unfortunately, it is not
always clear how exactly the left-concept and the right-concept are related.
Sometimes they are negations of each other (e.g. BP \#21, Fig.~\ref{exbp21}),
but most often they are not (e.g. BP \#25, Fig.~\ref{exbp25}). As, by
definition, the negation of the concept for one side is a valid concept for the
other side we did not allow for negation, assuming that people will strongly
prefer the non-negated concepts. This is consistent with concept attainment
studies where subjects usually prefer the positive examples of boolean concepts
\cite{Feldman_2000}. In this way, focusing attention on one side or the other
can lead to genuinely different concepts.

%++++++++++++++++++++++++++++++++++++++++++++++++++++++++++++++++++++++++++++++
\section{A visual language for solving Bongard problems}
%++++++++++++++++++++++++++++++++++++++++++++++++++++++++++++++++++++++++++++++

\begin{figure}[t]
  \centering
  \includegraphics[width=0.5\textwidth]{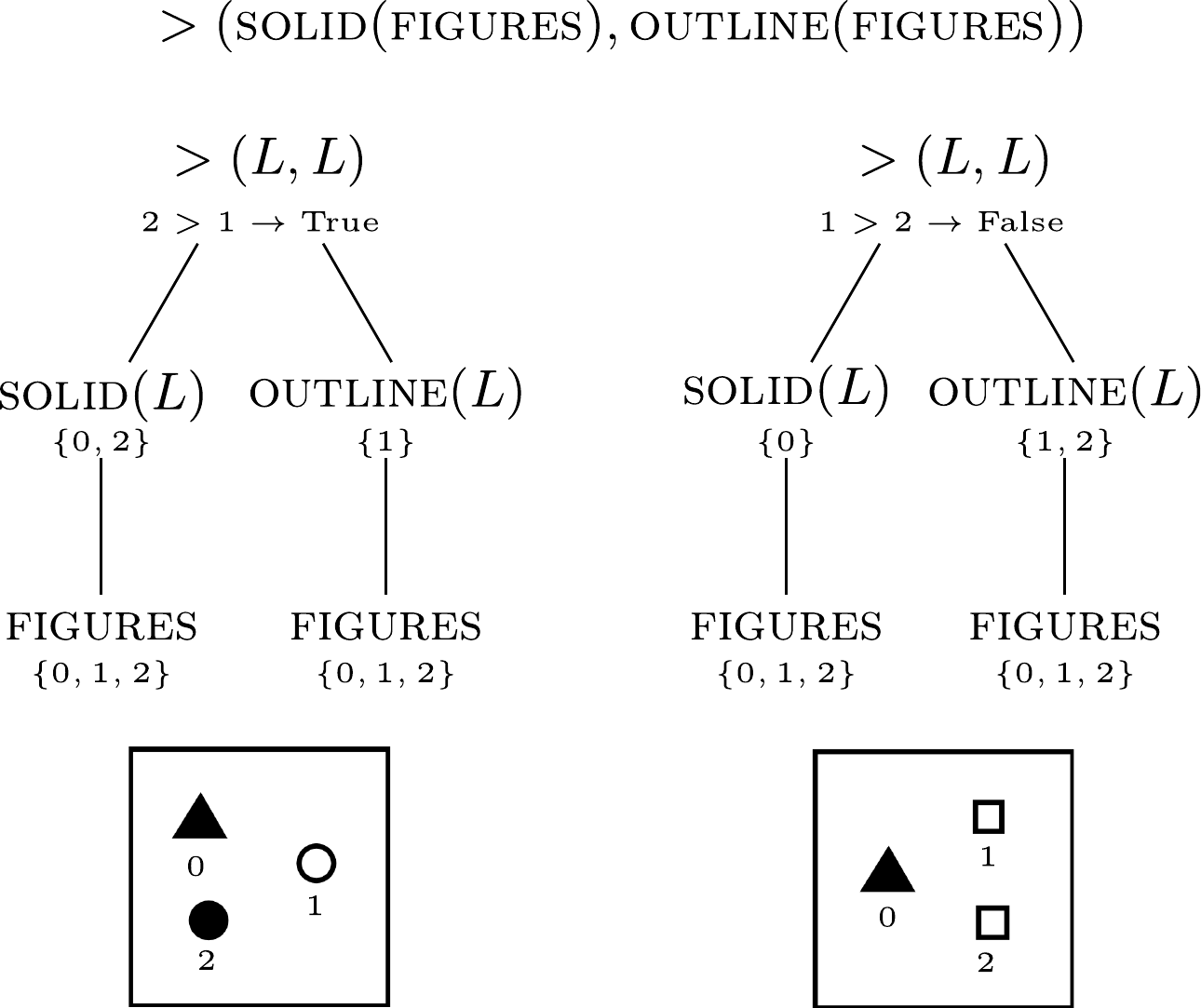}
  \caption{\label{parseTree} Whether a sentence is true of an example image is
  checked by first identifying all objects in an example image through connected
  component labeling and numbering them. In the two example images shown, the
  the evaluation of the expression starts with with all figures. Then the solid
  and the outlined figures are filtered out. Finally, it is checked wether there
  are more outline figures than solid figures in each example figure.}
\end{figure}

In the following we will describe the formal language in which solutions to
Bongard problems will be expressed. To anticipate what sentences in this formal
language will look like, consider the following solution to BP \#27 (Fig.\
\ref{exbp27})
$$\RIGHT : \MORE(\OUTLINE(\FIGURES),\SOLID(\FIGURES))$$
and its intended semantics. The expression after the colon applies to the
example images on the right, as indicated by the symbol $\RIGHT$, i.e. the
logical expression is true for all example images on the right and false for all
example images on the left. The symbol $\MORE$ indicates a comparison in number
of objects of two sets in an example image. It is true iff the set in the first
argument has more elements than the set in the second argument. $\FIGURES$ is
the set of all figures (or objects) in an example image. $\SOLID$ and $\OUTLINE$
act like filters that select all solid and outline figures. Hence, the meaning
of the full sentence is: On the right, there are more outline figures than solid
figures. Fig.~\ref{parseTree} illustrates how this expression is evaluated on
two example images. To explain the formal language and its semantics in more
detail, we will proceed from the bottom up, starting with the basic visual
vocabulary and ending with the solutions that are given in full sentences.

\subsection{Extracting visual properties from Bonagard's images}\label{Sec:Features}

The visual language we will describe is based on objects that are extracted
using standard connected-component labeling. Bongard calls each visual object a
figure. A sentence in the language specifies a logical function that checks
whether certain visual properties of the figures are true for the images at
hand. The most common figure types in Bongard problems are circles, rectangles,
and triangles. We therefore implemented functions that can recognize each of
these categories in binary images. In addition to these categorical features we
also have functions that extract quantitative features from the objects, namely
position, size (logarithm of the area in pixels), orientation (angle of the
largest eigenvector), convexity (ratio of area to area of convex hull),
compactness (isoperimetric quotient), and elongation
\cite{Stojmenovic_Zunic_2008}. From these features, specific boolean properties
of the figures can be computed, e.g. whether a figure is solid or outlined, or
whether a figure is small or large compared to the other figures. Solving
Bongard problems also requires relational features and we implemented functions
that compute the distance between two figures, the angle of one figure relative
to another, and whether a figure is inside another one. A last set of functions
can transform the binary pixel images of the figures into new images: one
returns the filled convex hull of the input image and another one inverts it,
thereby exposing the holes in it. For all these low-level visual computations we
used standard implementations whenever possible.\footnote{\url{http://scikit-image.org}}

%..........................................................................................................
\subsection{Context free grammar for representing visual concepts}\label{Sec:cfg}
%..........................................................................................................
\begin{table}
\small
\centering
\begin{tabular}{llll}
                       & $R \to \LEFT: S$       & $R \to \RIGHT: S$          &                           \\
                       &                        &                            &                           \\
$S \to \EXISTS(L)$     & $S \to \EQUALNUM(L,L)$ & $S \to \GREATERLA(L,A)$    & $S \to \MORESIMLA(L,A)$   \\
$S \to \EXACTLY(N,L)$  & $S \to \MORE(L,L)$     & $S \to \GREATERLLA(L,L,A)$ & $S \to \MORESIMLLA(L,L,A)$\\
                       &                        &                            &                           \\
$L \to \CAP(L,L)$      & $L \to \INSIDE(L)$     & $L \to \SOLID(L)$          & $L \to \FIGURES$          \\
$L \to \CUP(L,L)$      & $L \to \CONTAINS(L)$   & $L \to \OUTLINE(L)$        & $L \to \CIRCLES$          \\
$L \to \SETMINUS(L,L)$ & $L \to \ALIGNED(L)$    & $L \to \BIG(L)$            & $L \to \TRIANGLES$        \\
                       & $L \to \GET(L,T)$      & $L \to \SMALL(L)$          & $L \to \RECTANGLES$       \\
                       &                        & $L \to \HIGH(L,A)$         &                           \\
                       &                        & $L \to \LOW(L,A)$          &                           \\
                       &                        &                            &                           \\
$A \to \XPOS$          & $A \to \ORIENTATION$   & $A \to \SIZE$              & $A \to \ELONGATION$       \\
$A \to \YPOS$          & $A \to \NCORNERS$      & $A \to \COMPACTNESS$       & $T \to \HULLS$            \\
$A \to \DISTANCE$      & $A \to \COLOR$         & $A \to \CONVEXITY$         & $T \to \HOLES$            \\
                       &                        &                            &                           \\
$N \to 1$              & $N \to 2$              & $N \to 3$                  & $N \to 4$                 \\
\end{tabular}
\caption{The context-free grammar $\mathcal{G}$ that is used to
  describe solutions to Bongard problems\label{Tab:Grammar}}
\end{table}
%..........................................................................................................

With these basic visual features in mind we define a context-free grammar
$\mathcal{G}$ that allows us to express complex visual concepts.

\newdefinition{definition}{Definition}

\begin{definition}
The visual grammar consists of a 4-tuple
$\mathcal{G}=(\mathcal{NT},\mathcal{T},\mathcal{R},R)$. The grammar has six
types of non-terminals, $\mathcal{NT}=\{R, S, L, A, T, N\}$. The production
rules $\mathcal{R}$ of the grammar are given in Table \ref{Tab:Grammar}. The
terminals $\mathcal{T}$ are the parentheses, the colon and all strings in small
caps in Table \ref{Tab:Grammar}. The generation of every sentence of
$\mathcal{G}$ starts with the startsymbol $R$.
\end{definition}

A rule $R$ is a description that applies either to the six images on the left
($\LEFT$) or the six images on the right ($\RIGHT$) side of a Bongard problem. A
sentence $S$ is a logical function that can be evaluated on an image to check
whether it is true for this image. For example, consider BP \#3 again (Fig.
\ref{exbp3}). A rule that describes this problem is
$\LEFT:\EXISTS(\OUTLINE(\FIGURES))$: there are outline figures on the left. Part
of the semantics of this rule is also that the negation applies to the six
images on the right, namely that there are no outline figures. As outlined and
solid figures are mutually exclusive, the same problem could also equivalently
be described by the rule $\RIGHT:\EXISTS(\SOLID(\FIGURES))$. We consider each of
the two rules to be a solution to BP \#3 and thereby together capture the
solution that was given by Bongard: ``Outline figures $\mid$ Solid figures.''
Remember that sometimes, for example for BP \#71 (Fig.~\ref{exbp71}), one side
is explicitly the negation of the other side, but most often this is not the
case (see Preliminary Observations above). In these other cases, like BP \#3,
one side is the negation of the other side only under certain presuppositions
that are implicit to the problem. For BP \#3 such an implicit presupposition is
that there is only one figure in each image and that each figure is either
outlined or solid.

For the full semantics behind the grammar consider the $L$-symbols next. The
terminals $\FIGURES$, $\CIRCLES$, $\TRIANGLES$, and $\RECTANGLES$ stand for
functions that for a given image return a set with all the figures of the
respective kind. The first three $L$-rules are standard set-theoretic
operations: intersection, union, and set-difference. Using just these and the
terminals an example for an $L$-subclause of a sentence is
$\CUP(\CIRCLES,\TRIANGLES)$ which returns a set with all the circles and all the
triangles in an image. If there are none the set will be empty.

%TODO: Make picture BP with labelled components at bottom, tree above and show
%how sets are combined/split as we traverse the tree.

The symbols $\INSIDE$ and $\CONTAINS$ refer to binary relations that are
implemented as monadic predicates, akin to some description logics
\cite{Baader_etal_2007}. Hence, $\INSIDE(L)$ returns the set containing each
object that lies inside of any object in $L$. Suppose there is a triangle inside
of a rectangle and nothing else. Then $\INSIDE(\RECTANGLES)$ will return a set
containing only this triangle. Similarly, $\CONTAINS(\RECTANGLES)$ will return
the triangle. The symbol $\ALIGNED$ stands for a function that returns the
biggest subset of $L$ with the property that all the objects lie on a straight
line. It does so by enumerating the power-set of $L$ and computing pairwise
angles between the centroids. If these angles do not differ more than a defined
threshold a subset is declared as being on a straight line.
%\footnote{As the number of subsets grows quickly with the number of
%  objects we have limited this function to 8 objects. If it is called
%  on more than that it returns the empty list.}

For the meaning of $\GET(L,T)$ we turn to the non-terminal $T$. The function
$\GET(L,T)$ evaluates the non-terminal $L$ first and passes the resulting set of
objects to the transformation process determined by $T$. Then the transformation
is applied to all objects in the set to either get their filled convex hull
($\HULLS$) or the holes ($\HOLES$) of the image of an object. $\GET(L,T)$ thus
returns a set of new objects that were not originally in the image. These
transformations are needed for some Bongard problems where the objects that the
rules relate to cannot be found through connected component labeling, but have
to be constructed (e.g. BP \#12, \#34, \#35, not shown).

The functions for the symbols $\SOLID$, $\OUTLINE$, $\BIG$, and $\SMALL$ filter
out the objects with the respective property from a set of objects. For the
semantics of $\HIGH(L,A)$ and $\LOW(L,A)$ we first turn to the non-terminals
$A$. The $A$-rules represent numerical attributes that can be extracted from
objects, such as size, position, or color (degree of blackness). Orientation is
the only circular variable and hence we need to be careful for comparisons to
make sense (e.g. we cannot compute means or variances on angles). The only
attribute that does not have an obvious semantics is $\DISTANCE$. When
$\DISTANCE$ is applied to a set of objects it computes for each object the
minimum distance to all the other objects in the set. As an example for the
semantics of $\HIGH$ consider $\HIGH(\FIGURES,\XPOS)$ that appears in the
solution to BP \#8 (Fig. \ref{exbp8}). It returns all the figures in a given
image that have a high x-position relative to all figures in all images of the
Bongard problem under consideration. In order for $\HIGH$ and $\LOW$ to be
meaningful there should be a noticeable difference between the figures with a
high and those with a low x-position and we search for the best split that
maximizes the difference between the $\HIGH-$ and the $\LOW$-class, ensuring
that the difference is large enough to be perceptually relevant. Note that,
$\SMALL(\TRIANGLES)$ is simply a shorthand for $\LOW(\TRIANGLES,\SIZE)$ and this
is also how it is implemented (see Discussion section).

The non-terminal $S$ connects all these operations on sets of objects into a
logical statement about the images. Each function that can be built by the
$S$-rules returns a truth value (as opposed to the $L$-functions that return
sets of objects). $\EXISTS(L)$ means the set $L$ is not empty and
$\EXACTLY(N,L)$ evaluates to true if there are exactly $N$ objects (where we
assume that people can only subitize up to four). To be able to check, for
example, whether there are as many or more triangles as circles we have the
relations $\EQUALNUM(L,L)$ and $\MORE(L,L)$.

We can also compare two sets of objects with respect to any of the visual
attributes $A$ using $\GREATERLLA(L,L,A)$, e.g.
$\GREATERLLA(\TRIANGLES,\CIRCLES,\XPOS)$ checks whether all triangles are right
of (have a greater x-position than) all circles. The two argument version
$\GREATERLA(L,A)$ compares the attribute on all the relevant figures for the
left and the right side of the Bongard problem, e.g.
$\LEFT:\GREATERLA(\FIGURES,\XPOS)$ means that the figures in the $\LEFT$-class
have a greater x-position than the figures in the $\RIGHT$-class (also solving
Bongard problem \#8).

To compare the similarity of objects with respect to an attribute we use
$\MORESIMLLA(L,L,A)$. Imagine an image where the triangles have the same size
but the circles vary in size, then $\MORESIMLLA(\TRIANGLES,\CIRCLES,\SIZE)$ will
compute the size of the circles and the triangles and return true if the
variance in size is smaller for the triangles than for the circles. It is a
little tricky to decide whether the differences in variance are perceptually
meaningful. We do this by checking that that the difference in log variance
between the classes is bigger than the differences in log variance within the
classes. Finally, $\MORESIMLA(L,A)$ checks whether the $L$-objects in the class
specified by the top rule $R$ are more similar to each other with respect to $A$
than the $L$-objects in the other class. For example, if in the left class every
image contains two objects and they are of the same size and in the right class
there are two objects in each image but with different sizes, then $\LEFT:
\MORESIMLA(\OBJECTS,\SIZE)$ will be true.

One complication for the $S$-rules is that for some of them the truth value is
\emph{undefined} if they involve empty sets, e.g.
$\GREATER(\TRIANGLES,\CIRCLES,\XPOS)$ is neither true nor false if there are no
triangles in the image. The condition that checks for equal numbers is also
problematic if called on two empty sets. While mathematically true it expands
the hypothesis space dramatically with hypotheses that humans will almost
certainly not entertain. For example, say, that the concept on the left is
``there does not exist a triangle'' (and some images also contain circles). Then
we can form sentences like $\EQUAL(\TRIANGLES,\CAP(\CIRCLES,\TRIANGLES))$, that
are highly unlikely to be considered by humans. Hence, all such cases involving
the ``unintuitive'' empty set are taken as undefined and will be excluded as
valid hypotheses (see Eq.~\ref{lik} below).

%++++++++++++++++++++++++++++++++++++++++++++++++++++++++++++++++++++++++++++++
\section{Probabilistic model and inference}\label{Sec:Inference}
%++++++++++++++++++++++++++++++++++++++++++++++++++++++++++++++++++++++++++++++

Given the above language for visual concepts, there are many ways to define a
score for how well a concept can solve a Bongard problem and many search
algorithms that could be used to find the best solution. Here, we frame the
problem as one of Bayesian inference and use standard sampling methods. This
approach has previously been very successful for other human concept learning
tasks \cite{Goodman_etal_2008,Overlan_etal_2017}. As there will usually be
several potential solutions, the system, and humans as well, should entertain
several of them and state their relative plausibility. Using sampling the full
posterior distribution over potential expressions is approximated instead of
just finding the one best solution through search.

\subsection{Problem formulation}

We assume, that a visual concept can be expressed by a rule using the above
context free grammar. For each Bongard problem two disjoint sets of six images
$E_{\LEFT}=\{E_1,...,E_6\}$ and $E_{\RIGHT}=\{E_7,...,E_{12}\}$ are given. The
six examples in $E_{\LEFT}$ belong to the class on the left while the other six
in $E_{\RIGHT}$ belong to class on the right, and $E=(E_{\LEFT},E_{\RIGHT})$ is
the tuple of both sets. We need to specify a likelihood that expresses the
probabilistic relationship between the images $E$ and a particular rule $R$ and
a prior probability distribution over rules $R$ within the grammar $G$. Then we
can compute the posterior distribution over rules $R$ from the grammar
$\mathcal{G}$ given the twelve examples $E$:
\begin{equation*}
  P(R|E,\mathcal{G}) \propto P(E|R)P(R|\mathcal{G}).
\end{equation*}

\subsection{Model}

We take the same prior as the Rational-Rules model that has previously been
applied successfully to human concept learning tasks \cite{Goodman_etal_2008}.
Every rule in our Grammar $G$ can be expressed as a sentence starting with one
of the six types of non-terminals $\{R, S, L, A, T, N\}$ defined above. More
complex rules are obtained by successively applying the production rules of our
Grammar $G$. This allows defining a prior over rules by assigning a probability
to each production resulting in the product of probabilities of all individual
production rules involved in generating a particular rule. By including
uncertainty over the production probabilities one can derive a prior over rules
obtained through our context-free grammar by integrating out the production
probabilities using a flat Dirichlet prior. This integration leads to the
following expression for the rule prior \cite[see][for
details]{Goodman_etal_2008}:
\begin{equation*}
  P(R|\mathcal{G}) = \prod_{Y \in \mathcal{NT}}
                     \frac{\upbeta(\textbf{C}_Y(R) + \textbf{1})}
                          {\upbeta(\textbf{1})},
\end{equation*}
where $\mathcal{NT}$ is the set of non-terminals in the grammar
$\mathcal{G}$, $\upbeta(\textbf{c})$ is the multinomial beta function, and
$\textbf{C}_Y(R)$ counts for non-terminal $Y$ the uses of each production in
$R$. Hence, $\textbf{C}_Y(R)$ has dimensionality equal to the number of
productions for $Y$ and $\textbf{1}$ is a vector of equal length.\footnote{Just
like for normal PCFGs \cite{Chi_1999} it is interesting to ask under which
conditions the distribution is proper.}

This prior has two key properties that are relevant in the context of solving
Bongard problems. First, this prior favors shorter rules compared to longer
rules because of the monotonicity property of the multinomial Beta function in
the number of counts of non-terminals in a rule. Accordingly, less complex
formulas are preferred. Second, this prior favors rules that contain identical
subtrees, thereby encouraging the reuse of productions in their generation. This
is because for the multinomial beta distribution the following relation holds:
\begin{equation*}
  \upbeta(i,j, ...)<\upbeta(i+1,j-1,...) \text{ if } i\geq j.
\end{equation*}
Thus a formula containing a non-terminal $Y$ twice is more probable than a
formula of equal length with two different productions of $Y$.

For the likelihood we choose:
\begin{equation*}
  P(E|R) \propto
  \begin{cases}
    1 &\text{if $E$ is compatible with $R$}\\
    0 &\text{otherwise.}
  \end{cases}
  \label{lik}
\end{equation*}
$E$ is compatible with $R$ iff all of the following conditions are met:
\begin{enumerate}
\item if $R$ refers to the $\LEFT$ side then all images in $E_{\LEFT}$
  fulfill it and no image in $E_{\RIGHT}$,
\item if $R$ refers to the $\RIGHT$ side then all images in
  $E_{\RIGHT}$ fulfill it and no image in $E_{\LEFT}$,
\item no image in $E_{\LEFT}$ or $E_{\RIGHT}$ returns \emph{undefined}
  when $R$ is applied to it,
\item the images $E$ are informative about the rule $R$.
\end{enumerate}
The last condition requires some explanation. To be \emph{informative} the
following pragmatic reasoning is applied.

\subsubsection{Pragmatic reasoning}

Bongard constructed each problem very carefully to illustrate the rule in
question. As we already emphasized in the introduction, we conceptualize Bongard
problems as having to infer the intended concept in a communicative situation.
As usual in communication, understanding Bongard's message requires pragmatic
reasoning. Say, for example, Bongard wanted to find examples for the rule
``there are circles or triangles'' on the left side of the problem:
$\LEFT:\EXISTS(\CUP(\CIRCLES,\TRIANGLES))$. Not all the images have to have
circles in them. However, if Bongard wants us to be able to solve the problem,
he will include some circles in the problem. If no circles appeared in any of
the images then we would not even consider $\CIRCLES$ when trying to solve the
problem. Hence, we assume in the likelihood that Bongard chose the examples to
be informative about the rule to account for this pragmatic effect.
% cite Piantadosi Quantifier once paper comes out?
How the informativeness condition is spelled out in detail can greatly improve
both the quality of the  found solutions and the time efficiency of our
inference mechanism. Taking this pragmatic effect into account is one of the
main innovations of our approach. Concretely, we reject $E$ if it does not
contain at least once each of the zero-arity $L$-terminals ($\OBJECTS$,
$\SOLID$, $\TRIANGLES$, ...) that are mentioned in the rule $R$. A number of
other symbols ($\INSIDE$, $\CONTAINS$, $\HIGH$, $\LOW$, $\DISTANCE$) require
that at least two (or even three in the case of $\ALIGNED$) objects are present
to be informative. Similarly, if $\INSIDE$ or $\CONTAINS$ are in the rule, at
least one image should have one object inside another object.  the
informativeness condition in the likelihood licenses pruning the search space
dramatically: If $E$ is not informative about $R$ (e.g., there are no triangles
even though $\TRIANGLES$ is mentioned in $R$) then $R$ has likelihood zero and
we do not need to explore it.

\subsection{Inference algorithm}

Sampling from the posterior distribution is done by a Metropolis-Hastings
algorithm with sub-tree regeneration as a proposal distribution
\cite{Goodman_etal_2008}, with the additional pragmatic pruning described in the
previous paragraph. To improve the convergence of the chain the likelihood in
the sampler is a soft version of the above likelihood that allows for some
mistakes in the rules (increasing the ``temperature''). If $n$ is the number of
mistakes that a rule makes on the 12 images then the likelihood was
$\varepsilon^n$, but all samples with $n>0$ were discarded at the end. We run
the model for each Bongard problem independently and for each posterior
distribution we set up 6 independent chains. For each chain we collect 50,000
samples (after thinning by a factor of ten and a burn-in of at least 100.000).
How many of the 300,000 samples fulfill the hard likelihood condition (number of
mistakes $n=0$) depends on how difficult the problem is to solve.

A few implementational details are worth mentioning. Sub-tree regeneration is
done with a PCFG and setting the probabilities with which each rule fires has an
influence on the performance of the algorithm. First, the probabilities need to
be set in a way that the PCFG does not produce infinite trees (i.e. it is proper
\cite{Chi_1999}). This can be achieved by not giving recursive rules too much
probability mass. Second, the informativeness constraint in the likelihood
licences setting the production probabilities for some rules to zero because
they will be rejected later anyway and hence need not be generated (note,
however, that this pruning interacts with the ``temperature'' parameter
$\varepsilon$). In order to make the algorithm run in acceptable time, it was
necessary to make extensive use of caching to avoid re-computations of
likelihoods and priors whenever possible.
% @Stefan: mehr Details? Noch was anderes, das wichitg wäre?

%++++++++++++++++++++++++++++++++++++++++++++++++++++++++++++++++++++++++++++++
\section{Inferring visual concepts for Bongard problems}\label{Sec:Results}
%++++++++++++++++++++++++++++++++++++++++++++++++++++++++++++++++++++++++++++++

% THE FOLLOWING DATA ARE BASED ON SIMULATION V6f
%.....................................................................
\begin{figure}[hp]

  % first subfigure
  \begin{minipage}[t]{1\textwidth}
    \centering
    \footnotesize
    % picture of Bongard problem
    \begin{minipage}[t]{0.35\textwidth}
      \vspace{0pt}
      \includegraphics[width=1\textwidth]{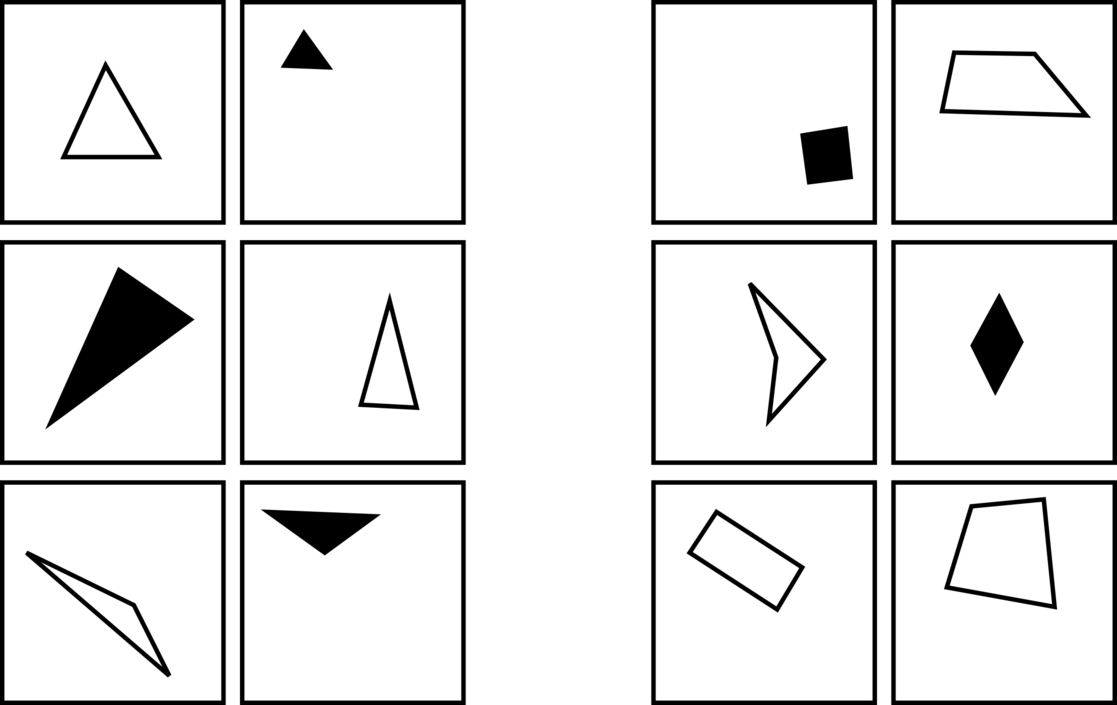}
    \end{minipage}
    \hfill
    % table with results
    \begin{minipage}[t]{0.6\textwidth}
      \vspace{0pt}
      {\noindent\scriptsize
        \begin{tabularx}{\textwidth}{Xr}
          \hline%---------------------------------------------------
          $\LEFT$                                           & $p$ \\
          \hline%---------------------------------------------------
          $\EXISTS(\TRIANGLES)$                             & .69 \\
          $\EXACTLY(1,\TRIANGLES)$                          & .17 \\
          $\EXISTS(\GET(\TRIANGLES,\HULLS))$                & .02 \\
          $\EXISTS(\CUP(\TRIANGLES,\TRIANGLES))$            & .00 \\
          $\EXACTLY(1,\GET(\TRIANGLES,\HULLS))$             & .00 \\
          \hline%---------------------------------------------------
          $\RIGHT$                                          & $p$ \\
          \hline%---------------------------------------------------
          $\GREATERLA(\FIGURES,\NCORNERS)$                  & .07 \\
          $\EXISTS(\HIGH(\FIGURES,\NCORNERS))$              & .00 \\
          $\GREATERLA(\CUP(\FIGURES,\FIGURES),\NCORNERS)$   & .00 \\
          $\GREATERLA(\CAP(\FIGURES,\FIGURES),\NCORNERS)$   & .00 \\
          $\GREATERLA(\CUP(\TRIANGLES,\FIGURES),\NCORNERS)$ & .00 \\
          \hline%---------------------------------------------------
          Remaining rules                                   & .04 \\
          \hline%---------------------------------------------------
      \end{tabularx}}
    \end{minipage}

    \medskip (a) BP \#6. Triangles $\mid$ Quadrangles.
  \end{minipage}

  \bigskip\medskip

  % second subfigure
  \begin{minipage}[t]{1\textwidth}
    \centering
    \footnotesize
    % picture of Bongard problem
    \begin{minipage}[t]{0.35\textwidth}
      \vspace{0pt}
      \includegraphics[width=1\textwidth]{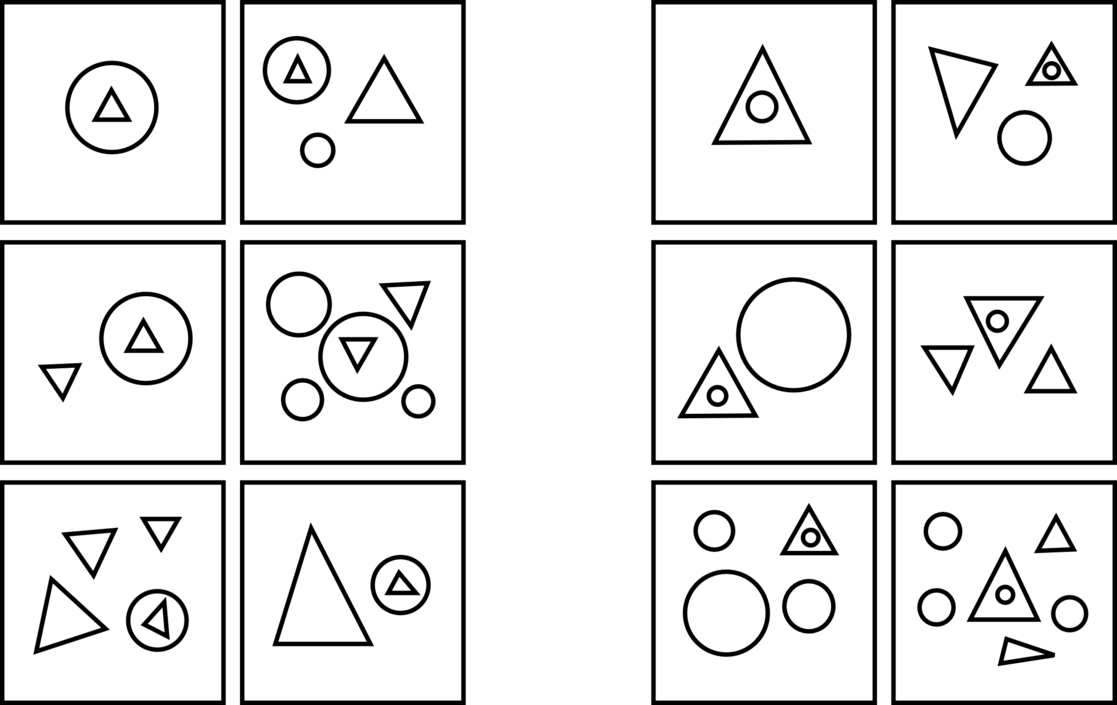}
    \end{minipage}
    \hfill
     % table with results
    \begin{minipage}[t]{0.6\textwidth}
      \vspace{0pt}
      {\noindent\scriptsize
        \begin{tabularx}{\textwidth}{Xr}
          \hline%--------------------------------------------
          $\LEFT$                                    & $p$ \\
          \hline%--------------------------------------------
          $\EXISTS(\CONTAINS(\TRIANGLES))$           & .10 \\
          $\EXISTS(\INSIDE(\CIRCLES))$               & .10 \\
          $\EXACTLY(1,\INSIDE(\CIRCLES))$            & .03 \\
          $\EXACTLY(1,\CONTAINS(\TRIANGLES))$        & .03 \\
          $\EXISTS(\OUTLINE(\CONTAINS(\TRIANGLES)))$ & .01 \\
          \hline%--------------------------------------------
          $\RIGHT$                                   & $p$ \\
          \hline%--------------------------------------------
          $\EXISTS(\CONTAINS(\CIRCLES))$             & .12 \\
          $\EXISTS(\INSIDE(\TRIANGLES))$             & .12 \\
          $\EXACTLY(1,\INSIDE(\TRIANGLES))$          & .03 \\
          $\EXACTLY(1,\CONTAINS(\CIRCLES))$          & .03 \\
          $\EXISTS(\CONTAINS(\INSIDE(\TRIANGLES)))$  & .01 \\
          \hline%--------------------------------------------
          Remaining rules                            & .43 \\
          \hline%--------------------------------------------
      \end{tabularx}}
    \end{minipage}

    \medskip (b) BP \#47. Triangle inside of circle
    $\mid$ Circle inside of triangle.
  \end{minipage}

  \bigskip\medskip

  % third subfigure
  \begin{minipage}[t]{1\textwidth}
    \centering
    \footnotesize
    %% picture of Bongard problem
    \begin{minipage}[t]{0.35\textwidth}
      \vspace{0pt}
      \includegraphics[width=1\textwidth]{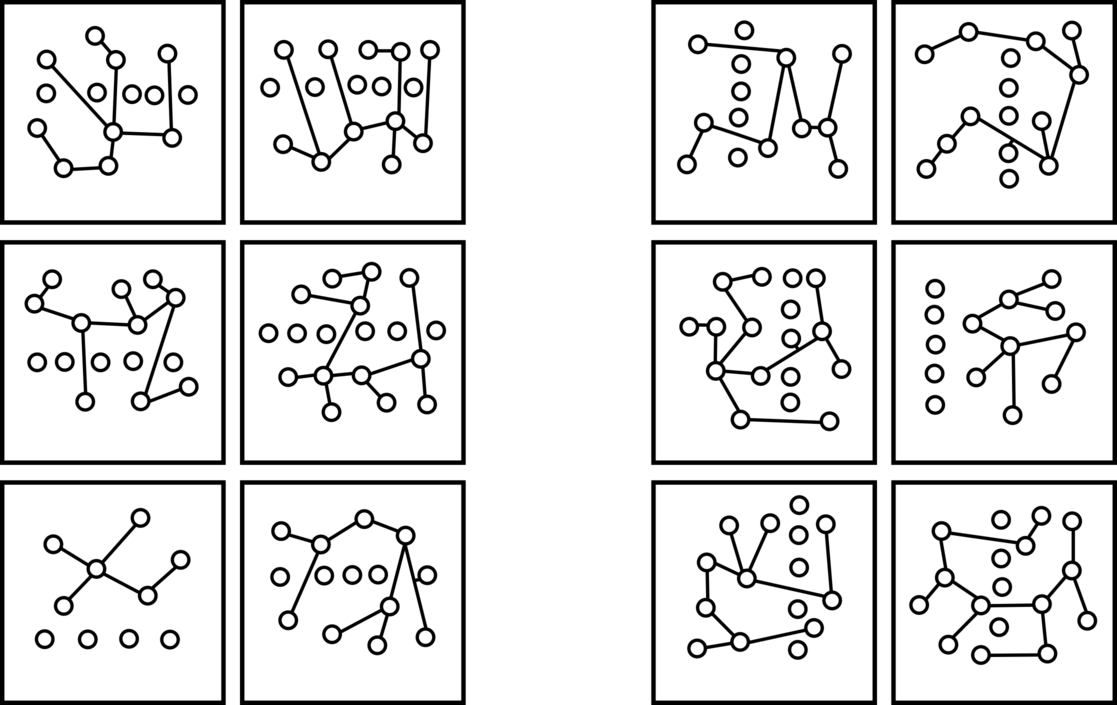}
    \end{minipage}
    \hfill
     % table with results
    \begin{minipage}[t]{0.6\textwidth}
      \vspace{0pt}
      {\noindent\scriptsize
        \begin{tabularx}{\textwidth}{Xr}
          \hline%----------------------------------------
          $\LEFT$                                & $p$ \\
          \hline%----------------------------------------
          $\MORESIMLA(\FIGURES,\YPOS)$           & .24 \\
          $\MORESIMLA(\CIRCLES,\YPOS)$           & .23 \\
          $\MORESIMLA(\SMALL(\FIGURES),\YPOS)$   & .01 \\
          $\MORESIMLA(\ALIGNED(\FIGURES),\YPOS)$ & .01 \\
          $\MORESIMLA(\OUTLINE(\CIRCLES),\YPOS)$ & .01 \\
          \hline%----------------------------------------
          $\RIGHT$                               & $p$ \\
          \hline%----------------------------------------
          $\MORESIMLA(\CIRCLES,\XPOS)$           & .23 \\
          $\MORESIMLLA(\CIRCLES,\FIGURES,\XPOS)$ & .01 \\
          $\MORESIMLA(\OUTLINE(\CIRCLES),\XPOS)$ & .01 \\
          $\MORESIMLA(\ALIGNED(\CIRCLES),\XPOS)$ & .01 \\
          $\MORESIMLA(\SMALL(\FIGURES),\XPOS)$   & .01 \\
          \hline%----------------------------------------
          Remaining rules                        & .21 \\
          \hline%----------------------------------------
      \end{tabularx}}
    \end{minipage}

    \medskip (c) BP \#66. Unconnected circles on a
    horizontal line $\mid$ Unconnected circles on a vertical line.
  \end{minipage}

  \bigskip\medskip

  % fourth subfigure
  \begin{minipage}[t]{1\textwidth}
    \centering
    \footnotesize
    %% picture of Bongard problem
    \begin{minipage}[t]{0.35\textwidth}
      \vspace{0pt}
      \includegraphics[width=1\textwidth]{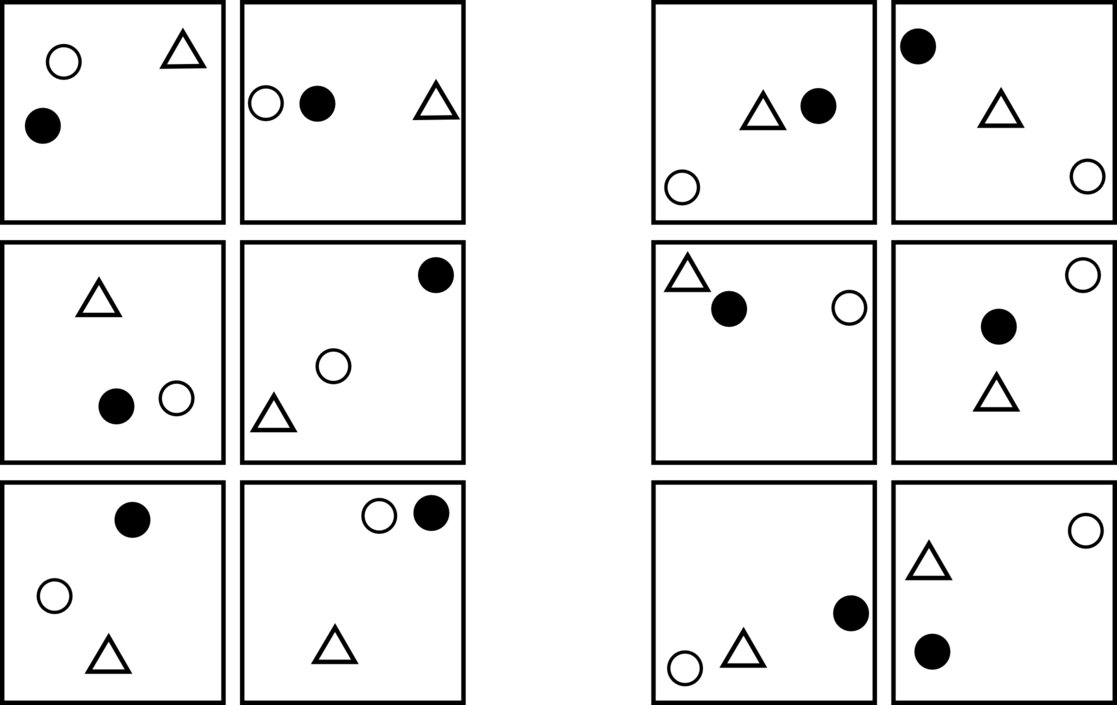}
    \end{minipage}
    \hfill
     % table with results
    \begin{minipage}[t]{0.6\textwidth}
      \vspace{0pt}
      {\noindent\scriptsize
        \begin{tabularx}{\textwidth}{Xr}
          \hline%--------------------------------------------------------------------------
          $\LEFT$                                                                  & $p$ \\
          \hline%--------------------------------------------------------------------------
          No rules were found                                                      &     \\
          %                                                                         &     \\
          %                                                                         &     \\
          %                                                                         &     \\
          %                                                                         &     \\
          \hline%--------------------------------------------------------------------------
          $\RIGHT$                                                                 & $p$ \\
          \hline%--------------------------------------------------------------------------
          $\GREATERLLA(\CIRCLES,\CUP(\SOLID(\CIRCLES),\TRIANGLES),\DISTANCE)$      & .12 \\
          $\GREATERLLA(\CIRCLES,\CUP(\TRIANGLES,\SOLID(\CIRCLES)),\DISTANCE)$      & .09 \\
          $\GREATERLLA(\CIRCLES,\SETMINUS(\FIGURES,\OUTLINE(\CIRCLES)),\DISTANCE)$ & .06 \\
          $\GREATERLLA(\CIRCLES,\CUP(\SOLID(\FIGURES),\TRIANGLES),\DISTANCE)$      & .06 \\
          $\GREATERLLA(\CIRCLES,\CUP(\TRIANGLES,\SOLID(\FIGURES)),\DISTANCE)$      & .05 \\
          \hline%--------------------------------------------------------------------------
          Remaining rules                                                          & .61 \\
          \hline%--------------------------------------------------------------------------
      \end{tabularx}}
    \end{minipage}

    \medskip (d) BP \#79. A dark circle is closer to the
    triangle than to the outline circle $\mid$ A dark circle is closer
    to the outline circle than to the triangle.
  \end{minipage}

  \medskip

  \caption{Example solutions for Bongard problems found by a
    Metropolis-Hastings algorithm. The four tables give the five most
    probable rules for each side of the respective Bongard problem
    ($\LEFT$ and $\RIGHT$). The last column shows the proportion of
    samples $p$ for each rule.}
  \label{results}
\end{figure}
%.....................................................................

We evaluated the system on a subset of Bongard problems. The selection of
problems was dependent on three criteria. First, a problem should not require
highly advanced computer vision techniques. We deliberately wanted to keep the
traditional computer vision side simple and focus on the cognitive, symbolic
representations. For this reason we have, for example, excluded all problems
that involve occlusion. While we expect that for occlusion the interaction
between visual and cognitive modules is particularly important, we will leave
this for future work. Second, a Bongard problem should not require specific
world knowledge or sophisticated object recognition. BP \#100, for example,
shows different letters on both sides with varying fonts and was therefore
excluded. And third, the problem has a solution that can be expressed in our
visual language. In many cases we could have extended the language to be able to
deal with specific Bongard problems but we tried to avoid the introduction of
ad-hoc constructs that will only appear in the solution of one problem. With
these restrictions we were left with a set of 39 Bongard problems.\footnote{BP
1, 2, 3, 4, 6, 7, 8, 9, 11, 12, 21, 22, 23, 24, 25, 26, 27, 28, 29, 34, 35, 36,
37, 38, 39, 40, 41, 42, 47, 48, 49, 51, 53, 56, 58, 65, 66, 71, 79} For the
remaining problems we already know that our system will not be able to solve
them because our visual language cannot express reasonable solutions. Note,
however, that even if the language can express a solution to a BP, it is not
obvious whether the solution will be found.

Fig.~\ref{results} shows the solutions for four exemplary Bongard problems. Fig.
\ref{results}a depicts a problem that is solved very well, in the sense that it
assigns a high probability to reasonable rules. The highest probability rule was
that there is a triangle on the left side, $\EXIST(\TRIANGLES)$, as it should
be. For the right side the highest probability rule was that it has more corners
than the left side, $\GREATERLA(\FIGURES,\NCORNERS)$. This is not the solution
that Bongard gives. His solution refers to ``quadrangles''. But even though the
system does not have a word for ``quadrangle'', arguably, it still found a
reasonable solution for the right side. Often the posterior was less peaked than
in Fig.~\ref{results}a, but this is to be expected with many logically
equivalent rules. The system can also find relatively complex rules as shown by
Fig.~\ref{results}d, although the fact that only rules for the right side were
found shows that this rule was hard to find and the Markov chain did not
converge. For most of the problems it was similarly obvious to all three authors
that our system found reasonable solutions as the ones with the highest
posterior probability. For instance, for the problem shown in
Fig.~\ref{results}b the ``correct'' solution is ``triangle inside of circle
$\mid$ circle inside of triangle.'' The system finds that on the left there is a
triangle inside another object and that there is a circle that contains another
object but not that one is in the other. As we decided, for reasons of
efficiency, to represent binary relations by monadic predicates this is the best
the language can do.

As these examples show, whether a rule expressed in a formal language is a
reasonable solution to a Bongard problem, very often requires a subjective
judgement. Is the expression close enough to Bongard's solution that is given in
natural language? If not, it may still be a solution that many people would
accept. While we think that the solutions that our system finds are all
reasonable, they are not always ideal. One systematic problem that we have
encountered is, again, of a pragmatic nature. In BP \#41 (not shown) the left
side shows ``outline circles on a straight line''. The system finds
$\EXISTS(\ALIGNED(\OUTLINE(\FIGURES)))$ and
$\EXISTS(\ALIGNED(\OUTLINE(\CIRCLES)))$ with the same probability. Bongard gives
the more specific solution, referring to circles instead of figures, but for the
system the two expressions have the same length and therefore the same prior
probability. For the system there is no reason to prefer one over the other. One
could try and include this pragmatic effect to prefer specificity in the
likelihood. However, in this particular problem all figures are circles and
hence with this presupposition there is no need to be more specific. We
therefore think that both solutions are reasonable.

Unclear presuppositions are an issue in many Bongard problems. Take BP \#6
(Fig.~\ref{results}a), again; each image contains exactly one figure
irrespective of whether they are on the left or on the right. For the left side
Bongard gives the solution ``triangles'' instead of ``there is only one figure
and it is a triangle'', presumably because there being exactly one figure does
not help discriminate between the left and the right side. Sometimes, however,
the presuppositions are reflected in Bongard's rule. The solution for BP \#25
(Fig.~\ref{exbp25}) is ``black figure is a triangle | black figure is a
circle.'' Here, it is made explicit that there is exactly one black figure on
both sides. Our system finds $\EXISTS(\SOLID(\TRIANGLES))$, there is a solid
triangle, as the most probable solution for the left side. If also finds the
slightly longer and therefore less probable $\EXACTLY(1,\SOLID(\TRIANGLES))$,
there is exactly one solid triangle. But it does not specify the full
presupposition that there is only one solid figure irrespective what the side
is. As our current implementation treats the solution for the left side and the
right side independently, it cannot express these presuppositions, and we
therefore also consider these problems reasonably solved.\footnote{We have tried
to learn one rule for the presuppositions that apply to both sides and two
separate rules that have to hold additionally for each side. So far, however,
the results of this approach have been disappointing. As the prior prefers
shorter rules and more specific rules are longer, the presuppositions tend to be
too unspecific (``there exist figures'' instead of ``there exists exactly one
solid figure''). Using the size principle \cite{Tenenbaum_Griffiths_2001} to
prefer more specific solutions in the likelihood seems like a good idea to solve
this problem, but it is not immediately clear how it can be applied here. \nopagebreak}

Overall, we consider 35 of the 39 problems to be solved, i.e. all three authors
agree that the solutions are reasonable. However, not all readers might agree
with our assessment and we have therefore included all results in the
supplementary material (just 4 of the 39 problems are shown in Fig.
\ref{results}). For the remaining 4 problems the system found the intended
solution among the top five solutions on both sides, but it also found another,
shorter solution that received a higher probability. In two cases one could
argue that the solution is as valid as the one given by Bongard. For BP \#27
(Fig.~\ref{exbp27}) the system finds:
$\RIGHT:\MORE(\OUTLINE(\FIGURES),\CIRCLES$), on the right side there are more
outline figures as circles. And for BP \#49 (Fig.~\ref{exbp49}) it finds:
$\LEFT:\MORESIMLA(\FIGURES,\DISTANCE)$, the similarity between the distances
among the figures is more similar on the right side than on the left side. In BP
\#66 (Fig.~\ref{results}c) the system finds reliably that the y-positions of the
circles are more similar on the left than on the right,
$\MORESIMLA(\CIRCLES,\YPOS)$, but it also finds $\MORESIMLA(\FIGURES,\YPOS)$
with equal probability. Because of the connected-component labeling the system
treats all the connected circles as one object and as we compute the position as
the mean of an object the system also finds that actually all figures (not just
the circles) have a more similar y-position. As the system cannot see the
circles in the connected circles, it does give a reasonable solution---but for
the wrong reasons. Finally, BP \#48 finds a rule that exploits the brittleness
of the visual preprocessing where we have set thresholds to decide whether
logical predicates apply and the rule that is found is just above this
threshold.
% only 27, 49, and 66 remain problematic, but okay. 48 is still very
% brittle and apparently on some grid nodes scikit evaluates a bad
% rule as true... this will need tweaking of the parameters that we
% could not do in time anymore

The above comparison between the rules provided by Bongard and the rules found
by our system point to a fundamental difficulty regarding the evaluation of
cognitive vision systems beyond object recognition. Unless one runs an
experiment with human participants who judge the reasonableness of a solution,
it is hard to quantitatively evaluate the performance of a system that solves
Bongard problems. Of course, the evaluation of object recognition systems also
relies on humans providing ground-truth labels, but to benchmark artificial
systems you only need to count how often their labels agree \emph{exactly} with
the human labels. Here, the comparison is not as simple. Hence, while we think
that Bongard problems provide an interesting testbed for visual cognition, they
do not easily allow us to benchmark different systems. Having said that, our
system clearly performs much better than Phaeaco \cite{Foundalis_2006} that can
only solve around 10 problems. RF4 \cite{Saito_Nakano_1996} solves roughly the
same number of problems as our system but the inputs are hand-translated logical
rules and not images. Presumably the researchers did not write down the large
number of correct, but irrelevant facts about each image that our system needs
to deal with. Therefore, even though we cannot easily measure progress, our
system clearly does outperform previous systems.

%------------------------------------------------------------------------------
\section{Discussion}
%------------------------------------------------------------------------------

% At the moment the system works on clean binary images and hand-crafted detectors
% for each basic term. The visual feature extraction did require some tweaking of
% threshold parameters. As the images were relatively clean, it was not a big
% problem to find parameters that work well for all the images in our set.
% Ultimately one would like to replace the visual side of our system with more
% sophisticated and robust methods for object recognition and feature extraction.
% If these methods are also formulated as probabilistic models they can, in
% principle, be integrated smoothly into the overall architecture.

Bongard problems were originally constructed to illustrate that human visual
cognition is much more flexible and language-like than the first pattern
recognition algorithms. Interestingly, Bongard problems still pose a challenge
for modern day artificial intelligence and progress has been surprisingly slow.
We have constructed a visual language that allows us to solve a significant
number of Bongard problems, while still using images as inputs. Like in other
human concept learning tasks, using Bayesian inference and the Rational-Rules
prior proved to be a fruitful starting point \cite{Goodman_etal_2008}.
Additionally, we found it helpful to conceptualize Bongard problems as concept
communication problems \cite{Shafto_Goodman_2008}. This allowed us to prune
large portions of the search space through pragmatic reasoning.

As mentioned in the introduction, it is an interesting question to ask where
vision ends and cognition starts \cite{Pylyshyn_1999b}. Here, we have
concentrated on the cognitive side, even though the inputs were actual images.
In a language for vision the interface between the visual and the cognitive
system is given by the symbols that can be grounded out directly in operations
on the visual image. In our language, some of the terminals seem far removed
from the visual input, however, and therefore one might worry that our language
is not very domain general. For example, we have included the predicates
$\SOLID$ and $\OUTLINE$ as primitives for the language because they not only
help in problem \#3 but occur in many other Bongard problems, too. For the basic
vocabulary that we have chosen one can argue that a mature human
visual-cognitive system comes equipped with these concepts and that for solving
Bongard problems we want to focus on how known visual concepts are combined to
form new concepts, rather than explaining how the basic terms were learned in
the first place. It is, in principle, easy to replace our simple visual system
with a more advanced and robust visual system that is capable of learning (such
as a neural network) and thereby explain, e.g., how the concept for $\SOLID$ is
learned from more realistic images. Nevertheless, learning might also happen at
the symbolic level. Note that our language has a predicate $\COLOR$ that will
return a high value if an object is black and a low values if an object is
white. We can therefore express $\SOLID(\TRIANGLES)$ equivalently as
$\HIGH(\TRIANGLES, \COLOR)$, namely those triangles that have a high value on
the color attribute. Similarly we can paraphrase $\BIG(\TRIANGLES)$ as
$\HIGH(\TRIANGLES,\SIZE)$. The shorter expressions are preferred in the prior
and therefore adding these redundant abbreviations make a difference for the
performance of the system. It is straightforward to imagine why and how these
redundant shorter representations might have been learned. If
$\HIGH(\TRIANGLES,\SIZE)$ is used very frequently then we would like to make one
chunk out of it. So far we have treated each Bongard problem separately but
using a sequence of Bongard problems and re-using previously learned concepts
for new problems seems very desirable for explaining how some of the basic
concepts are learned. In future work, this could be achieved by using fragment
grammars as a prior \cite{ODonnell_etal_2009}.

Our primary goal in this paper was to develop a useful language for representing
simple, geometric, visual scenes. While our language is useful for Bongard
problems, it is certainly not a domain-general language for vision. What is the
right language of vision? Taking triangles and circles as basic concepts is
different than taking lines and corners as basic concepts. Whether one judges a
system that solves Bongard problems to be successful crucially depends on
knowing how the system works internally and whether one considers the
representations and mechanisms that lead to the solution as cognitively
interesting. The more problems can be solved intelligently with general
mechanisms and without using cheap tricks, the better. The art is to choose the
right vocabulary and the right formal language in which to express visual
concepts. We hope that some of the constructions we proposed here might turn out
to be useful for other visual and cognitive tasks. In fact, we see this project
as part of a larger enterprise in cognitive science and AI to discover
appropriate representations for a domain-general probabilistic language of
thought \cite{Goodman_etal_2015,Piantadosi_etal_2016}.

But even the right language of vision will not be enough to explain how humans
solve Bongard problems. As Bongard problems can be considered communication
problems---Bongard tries to communicate a concept through the examples---we do
not only have to deal with the syntax and the semantics of visual concepts but
also with the pragmatics of the situation. Improving the vision module and the
formal language are certainly necessary if we want to solve more Bongard
problems in the future, or if we want to deal with other visual-cognitive tasks.
When we started working on Bongard problems, we thought these are the most
important aspects for progress on visual cognition. However, it turned out that
without also considering the pragmatic reasoning that allows us to prune away
large portions of the search space, we could not get our system to work. Many of
the remaining issues that we encounter with the solutions that our system
produces are of a pragmatic nature. For example, if Bongard chose very specific
examples, he probably did not want us to consider more general hypotheses, even
if they are much shorter. Similarly, for many problems there are implicit
presuppositions that apply to all images in the problem. Without these
presuppositions the solutions for each side simply do not make sense.

The pragmatic situation in Bongard problems is very particular and therefore
these pragmatic issues might seem irrelevant for other visual-cognitive tasks.
However, we suspect that any visual-cognitive task on which two humans interact
through language will involve a certain amount of pragmatic reasoning. For
example, if there is only one circle in a scene, it is unnecessary to be more
specific and refer to it as the outline circle \cite{Frank_Goodman_2012}. Hence,
any AI system that will interact with humans on a visual-cognitive task, will
also have to have some pragmatic reasoning abilities (which is currently missing
even from some of the most advanced systems
\cite[e.g.][]{Malinowski_Fritz_2014,Forbus_etal_2017}). Bongard problems are a
testbed to study this kind of interaction of image processing, conceptual
representations, and inductive and pragmatic reasoning. Further progress on
integrating these visual and cognitive processes into a working system for
solving Bongard problems is likely to inspire general architectures for visual
cognition that go beyond answering ``what is where''.

\section{Acknowledgments} CR was supported by the BMBF Project Bernstein Fokus:
Neurotechnologie Frankfurt, FKZ 01GQ0840.

\small{
\bibliographystyle{abbrv}
\bibliography{references}}
\appendix
\iffalse
% Report was automatically generated by get_results.py
% Commandline flag was: only_correct_rules
% ADDED BONGARD'S RULES BY HAND IN JANUARY 2018
\documentclass{article}
\usepackage{amsmath, amssymb, upgreek, graphicx,tabularx}
\b[a4paper,top=1cm,bottom=2cm,left=1cm,right=1cm]{geometry}
\input{grammar_operators.tex}

\begin{document}
\fi

\section*{BP \#1}
%==========================================
% BP #1
%       Samples: 299484 Runs: 6 Burn-in: 100000
%       Discarded rules with mistakes: 518
%       p = proportion of samples for this rule
%==========================================
\begin{minipage}[t]{0.4\textwidth}
\vspace{0pt}
\includegraphics[width=0.95\textwidth]{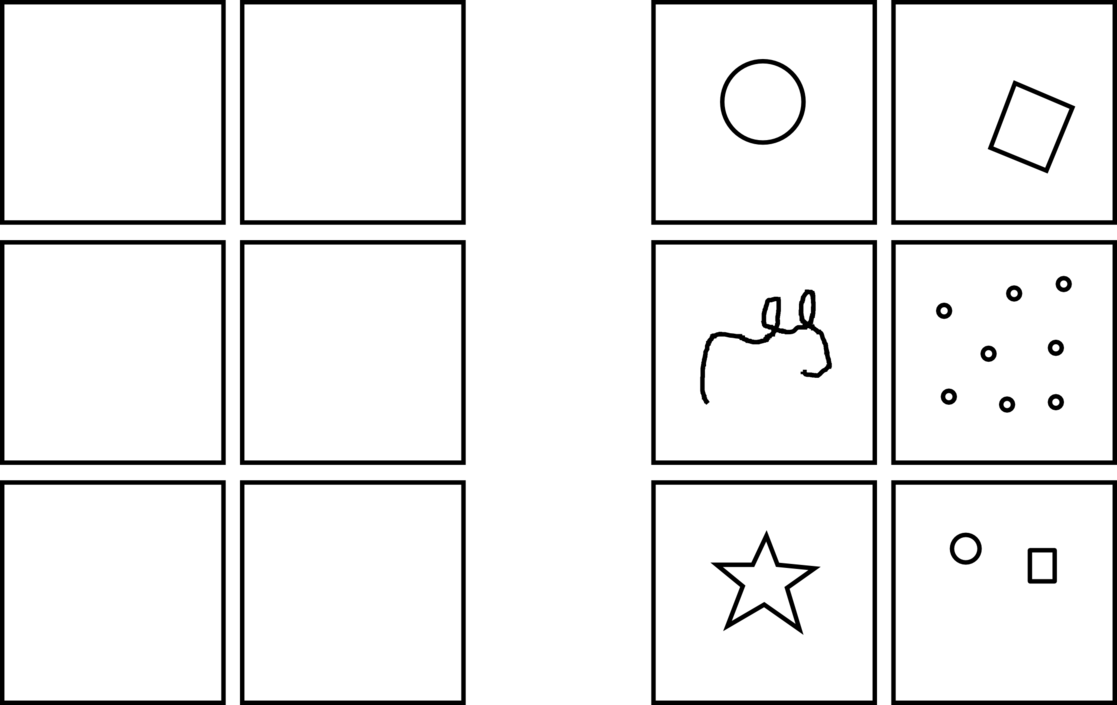}
\end{minipage}
\hfill
\begin{minipage}[t]{0.5\textwidth}
\vspace{0pt}
Samples: 299484\\
Runs: 6\\
Burn-in: 100000\\
Discarded rules with mistakes: 518\\
$p$ = proportion of samples for this rule\\
\end{minipage}

\bigskip

{\noindent\scriptsize
\begin{tabularx}{\textwidth}{Xr}
\hline%------------------------------------
$\LEFT$ empty figure               & $p$ \\
\hline%------------------------------------
\hline%------------------------------------
$\RIGHT$ not empty figure          & $p$ \\
\hline%------------------------------------
$\EXISTS(\FIGURES)$                & .84 \\
$\EXISTS(\OUTLINE(\FIGURES))$      & .05 \\
$\EXISTS(\GET(\FIGURES,\HULLS))$   & .02 \\
$\EXISTS(\GET(\FIGURES,\HOLES))$   & .02 \\
$\EXISTS(\CAP(\FIGURES,\FIGURES))$ & .01 \\
\hline%------------------------------------
Remaining rules                    & .07 \\
\hline%------------------------------------
\end{tabularx}}

\newpage
\section*{BP \#2}
%================================================
% BP #2
%       Samples: 296981 Runs: 6 Burn-in: 100000
%       Discarded rules with mistakes: 3020
%       p = proportion of samples for this rule
%================================================
\begin{minipage}[t]{0.4\textwidth}
\vspace{0pt}
\includegraphics[width=0.95\textwidth]{bpimgs/lowres-p002}
\end{minipage}
\hfill
\begin{minipage}[t]{0.5\textwidth}
\vspace{0pt}
Samples: 296981\\
Runs: 6\\
Burn-in: 100000\\
Discarded rules with mistakes: 3020\\
$p$ = proportion of samples for this rule\\
\end{minipage}

\bigskip

{\noindent\scriptsize
\begin{tabularx}{\textwidth}{Xr}
\hline%------------------------------------------
$\LEFT$ large figures                    & $p$ \\
\hline%------------------------------------------
$\GREATERLA(\FIGURES,\SIZE)$             & .44 \\
$\EXISTS(\BIG(\FIGURES))$                & .25 \\
$\EXACTLY(1,\BIG(\FIGURES))$             & .06 \\
$\EXISTS(\BIG(\BIG(\FIGURES)))$          & .03 \\
$\EXISTS(\HIGH(\FIGURES,\SIZE))$         & .02 \\
\hline%------------------------------------------
$\RIGHT$ small figures                   & $p$ \\
\hline%------------------------------------------
$\EXISTS(\SMALL(\FIGURES))$              & .03 \\
$\EXACTLY(1,\SMALL(\FIGURES))$           & .01 \\
$\EXISTS(\SMALL(\GET(\FIGURES,\HULLS)))$ & .01 \\
$\EXISTS(\LOW(\FIGURES,\SIZE))$          & .00 \\
$\EXISTS(\SMALL(\SMALL(\FIGURES)))$      & .00 \\
\hline%------------------------------------------
Remaining rules                          & .14 \\
\hline%------------------------------------------
\end{tabularx}}

\newpage
\section*{BP \#3}
%===============================================
% BP #3
%       Samples: 299873 Runs: 6 Burn-in: 100000
%       Discarded rules with mistakes: 129
%       p = proportion of samples for this rule
%===============================================
\begin{minipage}[t]{0.4\textwidth}
\vspace{0pt}
\includegraphics[width=0.95\textwidth]{bpimgs/lowres-p003}
\end{minipage}
\hfill
\begin{minipage}[t]{0.5\textwidth}
\vspace{0pt}
Samples: 299873\\
Runs: 6\\
Burn-in: 100000\\
Discarded rules with mistakes: 129\\
$p$ = proportion of samples for this rule\\
\end{minipage}

\bigskip

{\noindent\scriptsize
\begin{tabularx}{\textwidth}{Xr}
\hline%-----------------------------------------
$\LEFT$ outline figures                 & $p$ \\
\hline%-----------------------------------------
$\EXISTS(\OUTLINE(\FIGURES))$           & .16 \\
$\EXISTS(\GET(\FIGURES,\HOLES))$        & .07 \\
$\EXACTLY(1,\OUTLINE(\FIGURES))$        & .04 \\
$\EXACTLY(1,\GET(\FIGURES,\HOLES))$     & .02 \\
$\EXISTS(\OUTLINE(\OUTLINE(\FIGURES)))$ & .02 \\
\hline%-----------------------------------------
$\RIGHT$ solid figures                  & $p$ \\
\hline%-----------------------------------------
$\GREATERLA(\FIGURES,\COLOR)$           & .29 \\
$\EXISTS(\SOLID(\FIGURES))$             & .18 \\
$\EXACTLY(1,\SOLID(\FIGURES))$          & .04 \\
$\EXISTS(\SOLID(\SOLID(\FIGURES)))$     & .02 \\
$\EXISTS(\HIGH(\FIGURES,\COLOR))$       & .01 \\
\hline%-----------------------------------------
Remaining rules                         & .15 \\
\hline%-----------------------------------------
\end{tabularx}}

\newpage
\section*{BP \#4}
%=============================================================================
% BP #4
%       Samples: 299707 Runs: 6 Burn-in: 100000
%       Discarded rules with mistakes: 293
%       p = proportion of samples for this rule
%=============================================================================
\begin{minipage}[t]{0.4\textwidth}
\vspace{0pt}
\includegraphics[width=0.95\textwidth]{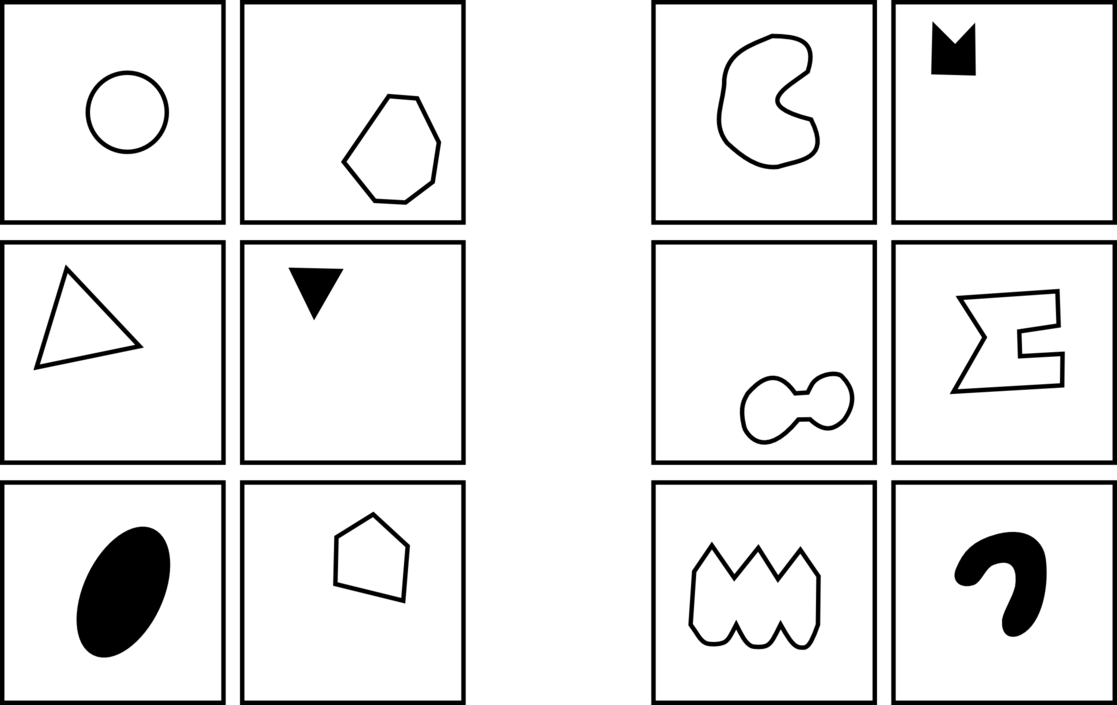}
\end{minipage}
\hfill
\begin{minipage}[t]{0.5\textwidth}
\vspace{0pt}
Samples: 299707\\
Runs: 6\\
Burn-in: 100000\\
Discarded rules with mistakes: 293\\
$p$ = proportion of samples for this rule\\
\end{minipage}

\bigskip

{\noindent\scriptsize
\begin{tabularx}{\textwidth}{Xr}
\hline%-----------------------------------------------------------------------
$\LEFT$ convex figures                                                & $p$ \\
\hline%-----------------------------------------------------------------------
$\GREATERLA(\FIGURES,\CONVEXITY)$                                     & .96 \\
$\GREATERLA(\CAP(\FIGURES,\FIGURES),\CONVEXITY)$                      & .01 \\
$\GREATERLA(\CUP(\FIGURES,\FIGURES),\CONVEXITY)$                      & .01 \\
$\GREATERLA(\CUP(\FIGURES,\TRIANGLES),\CONVEXITY)$                    & .00 \\
$\GREATERLA(\CUP(\CIRCLES,\FIGURES),\CONVEXITY)$                      & .00 \\
\hline%-----------------------------------------------------------------------
$\RIGHT$ nonconvex figures                                            & $p$ \\
\hline%-----------------------------------------------------------------------
$\GREATERLLA(\GET(\FIGURES,\HULLS),\FIGURES,\CONVEXITY)$              & .00 \\
$\GREATERLLA(\FIGURES,\GET(\FIGURES,\HULLS),\NCORNERS)$               & .00 \\
$\GREATERLLA(\SOLID(\GET(\FIGURES,\HULLS)),\FIGURES,\CONVEXITY)$      & .00 \\
$\GREATERLLA(\GET(\GET(\FIGURES,\HULLS),\HULLS),\FIGURES,\CONVEXITY)$ & .00 \\
$\GREATERLLA(\GET(\GET(\FIGURES,\HULLS),\HOLES),\FIGURES,\CONVEXITY)$ & .00 \\
\hline%-----------------------------------------------------------------------
Remaining rules                                                       & .02 \\
\hline%-----------------------------------------------------------------------
\end{tabularx}}

\newpage
\section*{BP \#6}
%=========================================================
% BP #6
%       Samples: 299620 Runs: 6 Burn-in: 100000
%       Discarded rules with mistakes: 382
%       p = proportion of samples for this rule
%=========================================================
\begin{minipage}[t]{0.4\textwidth}
\vspace{0pt}
\includegraphics[width=0.95\textwidth]{bpimgs/lowres-p006}
\end{minipage}
\hfill
\begin{minipage}[t]{0.5\textwidth}
\vspace{0pt}
Samples: 299620\\
Runs: 6\\
Burn-in: 100000\\
Discarded rules with mistakes: 382\\
$p$ = proportion of samples for this rule\\
\end{minipage}

\bigskip

{\noindent\scriptsize
\begin{tabularx}{\textwidth}{Xr}
\hline%---------------------------------------------------
$\LEFT$ triangles                                 & $p$ \\
\hline%---------------------------------------------------
$\EXISTS(\TRIANGLES)$                             & .69 \\
$\EXACTLY(1,\TRIANGLES)$                          & .17 \\
$\EXISTS(\GET(\TRIANGLES,\HULLS))$                & .02 \\
$\EXISTS(\CUP(\TRIANGLES,\TRIANGLES))$            & .00 \\
$\EXACTLY(1,\GET(\TRIANGLES,\HULLS))$             & .00 \\
\hline%---------------------------------------------------
$\RIGHT$ quadrangles                              & $p$ \\
\hline%---------------------------------------------------
$\GREATERLA(\FIGURES,\NCORNERS)$                  & .07 \\
$\EXISTS(\HIGH(\FIGURES,\NCORNERS))$              & .00 \\
$\GREATERLA(\CUP(\FIGURES,\FIGURES),\NCORNERS)$   & .00 \\
$\GREATERLA(\CAP(\FIGURES,\FIGURES),\NCORNERS)$   & .00 \\
$\GREATERLA(\CUP(\TRIANGLES,\FIGURES),\NCORNERS)$ & .00 \\
\hline%---------------------------------------------------
Remaining rules                                   & .04 \\
\hline%---------------------------------------------------
\end{tabularx}}

\newpage
\section*{BP \#7}
%======================================================================
% BP #7
%       Samples: 298734 Runs: 6 Burn-in: 100000
%       Discarded rules with mistakes: 1267
%       p = proportion of samples for this rule
%======================================================================
\begin{minipage}[t]{0.4\textwidth}
\vspace{0pt}
\includegraphics[width=0.95\textwidth]{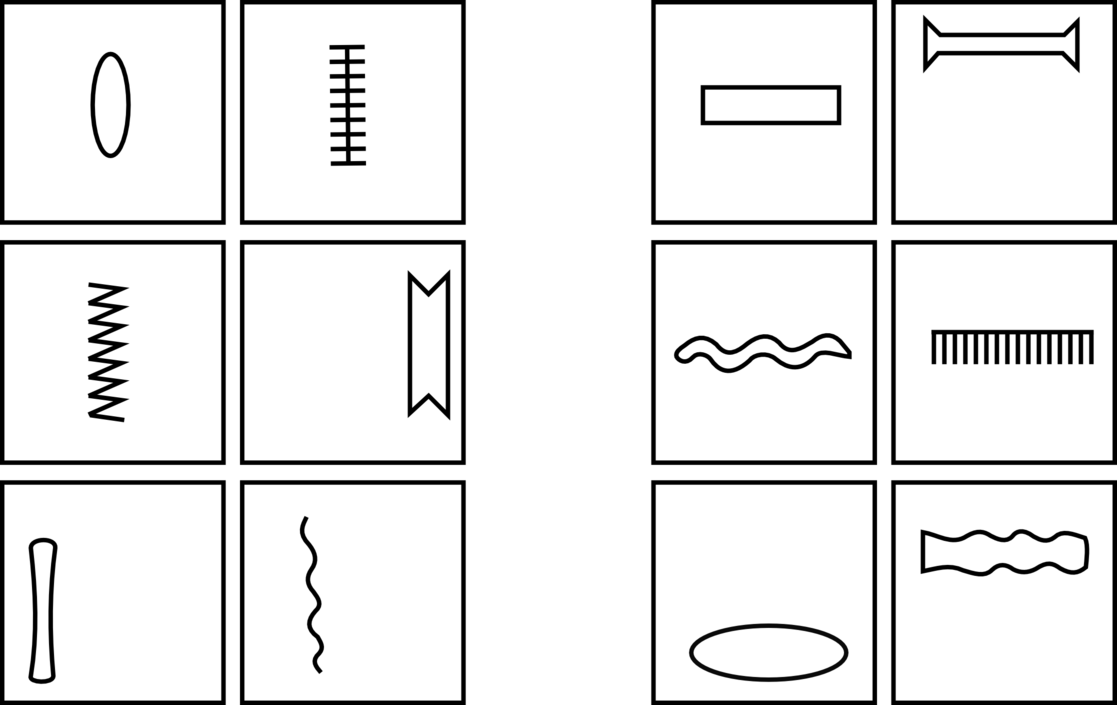}
\end{minipage}
\hfill
\begin{minipage}[t]{0.5\textwidth}
\vspace{0pt}
Samples: 298734\\
Runs: 6\\
Burn-in: 100000\\
Discarded rules with mistakes: 1267\\
$p$ = proportion of samples for this rule\\
\end{minipage}

\bigskip

{\noindent\scriptsize
\begin{tabularx}{\textwidth}{Xr}
\hline%----------------------------------------------------------------
$\LEFT$ figures elongated vertically                           & $p$ \\
\hline%----------------------------------------------------------------
$\GREATERLA(\FIGURES,\ORIENTATION)$                            & .86 \\
$\EXISTS(\HIGH(\FIGURES,\ORIENTATION))$                        & .04 \\
$\GREATERLA(\GET(\FIGURES,\HULLS),\ORIENTATION)$               & .02 \\
$\EXACTLY(1,\HIGH(\FIGURES,\ORIENTATION))$                     & .01 \\
$\GREATERLA(\CAP(\FIGURES,\FIGURES),\ORIENTATION)$             & .01 \\
\hline%----------------------------------------------------------------
$\RIGHT$ figures elongated horizontally                        & $p$ \\
\hline%----------------------------------------------------------------
$\EXISTS(\LOW(\FIGURES,\ORIENTATION))$                         & .01 \\
$\EXISTS(\LOW(\GET(\FIGURES,\HULLS),\ORIENTATION))$            & .00 \\
$\EXACTLY(1,\LOW(\FIGURES,\ORIENTATION))$                      & .00 \\
$\EXISTS(\BIG(\CUP(\LOW(\FIGURES,\ORIENTATION),\RECTANGLES)))$ & .00 \\
$\EXISTS(\LOW(\BIG(\FIGURES),\ORIENTATION))$                   & .00 \\
\hline%----------------------------------------------------------------
Remaining rules                                                & .04 \\
\hline%----------------------------------------------------------------
\end{tabularx}}

\newpage
\section*{BP \#8}
%====================================================
% BP #8
%       Samples: 299785 Runs: 6 Burn-in: 100000
%       Discarded rules with mistakes: 215
%       p = proportion of samples for this rule
%====================================================
\begin{minipage}[t]{0.4\textwidth}
\vspace{0pt}
\includegraphics[width=0.95\textwidth]{bpimgs/lowres-p008}
\end{minipage}
\hfill
\begin{minipage}[t]{0.5\textwidth}
\vspace{0pt}
Samples: 299785\\
Runs: 6\\
Burn-in: 100000\\
Discarded rules with mistakes: 215\\
$p$ = proportion of samples for this rule\\
\end{minipage}

\bigskip

{\noindent\scriptsize
\begin{tabularx}{\textwidth}{Xr}
\hline%----------------------------------------------
$\LEFT$ figures on the right side            & $p$ \\
\hline%----------------------------------------------
$\GREATERLA(\FIGURES,\XPOS)$                 & .76 \\
$\GREATERLA(\OUTLINE(\FIGURES),\XPOS)$       & .04 \\
$\EXISTS(\HIGH(\FIGURES,\XPOS))$             & .02 \\
$\GREATERLA(\GET(\FIGURES,\HOLES),\XPOS)$    & .02 \\
$\GREATERLA(\GET(\FIGURES,\HULLS),\XPOS)$    & .02 \\
\hline%----------------------------------------------
$\RIGHT$ figures on the left side            & $p$ \\
\hline%----------------------------------------------
$\EXISTS(\LOW(\FIGURES,\XPOS))$              & .03 \\
$\EXACTLY(1,\LOW(\FIGURES,\XPOS))$           & .01 \\
$\EXACTLY(1,\OUTLINE(\LOW(\FIGURES,\XPOS)))$ & .00 \\
$\EXISTS(\LOW(\OUTLINE(\FIGURES),\XPOS))$    & .00 \\
$\EXISTS(\GET(\LOW(\FIGURES,\XPOS),\HULLS))$ & .00 \\
\hline%----------------------------------------------
Remaining rules                              & .09 \\
\hline%----------------------------------------------
\end{tabularx}}

\newpage
\section*{BP \#9}
%===============================================================
% BP #9
%       Samples: 288291 Runs: 6 Burn-in: 100000
%       Discarded rules with mistakes: 11709
%       p = proportion of samples for this rule
%===============================================================
\begin{minipage}[t]{0.4\textwidth}
\vspace{0pt}
\includegraphics[width=0.95\textwidth]{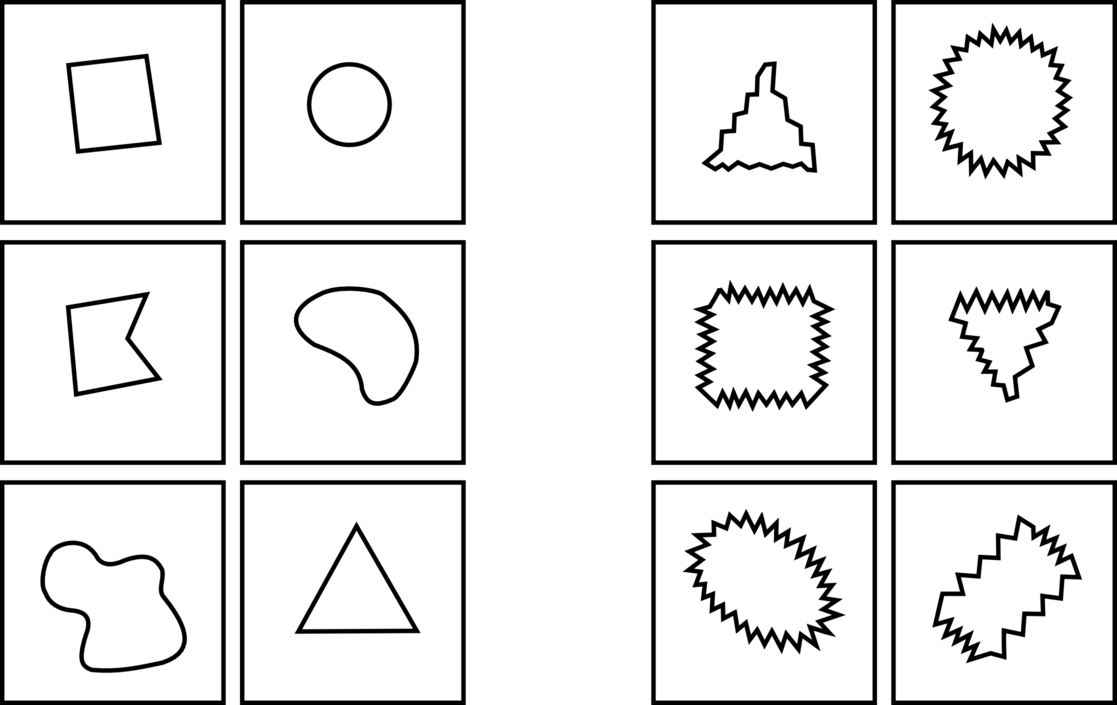}
\end{minipage}
\hfill
\begin{minipage}[t]{0.5\textwidth}
\vspace{0pt}
Samples: 288291\\
Runs: 6\\
Burn-in: 100000\\
Discarded rules with mistakes: 11709\\
$p$ = proportion of samples for this rule\\
\end{minipage}

\bigskip

{\noindent\scriptsize
\begin{tabularx}{\textwidth}{Xr}
\hline%---------------------------------------------------------
$\LEFT$ smooth contour figures                          & $p$ \\
\hline%---------------------------------------------------------
$\GREATERLA(\FIGURES,\COMPACTNESS)$                     & .88 \\
$\GREATERLA(\OUTLINE(\FIGURES),\COMPACTNESS)$           & .05 \\
$\GREATERLA(\GET(\FIGURES,\HOLES),\COMPACTNESS)$        & .02 \\
$\GREATERLA(\OUTLINE(\OUTLINE(\FIGURES)),\COMPACTNESS)$ & .01 \\
$\GREATERLA(\CAP(\FIGURES,\FIGURES),\COMPACTNESS)$      & .01 \\
\hline%---------------------------------------------------------
$\RIGHT$ twisting contour figures                       & $p$ \\
\hline%---------------------------------------------------------
\hline%---------------------------------------------------------
Remaining rules                                         & .04 \\
\hline%---------------------------------------------------------
\end{tabularx}}

\newpage
\section*{BP \#11}
%===============================================================
% BP #11
%       Samples: 299971 Runs: 6 Burn-in: 100000
%       Discarded rules with mistakes: 30
%       p = proportion of samples for this rule
%===============================================================
\begin{minipage}[t]{0.4\textwidth}
\vspace{0pt}
\includegraphics[width=0.95\textwidth]{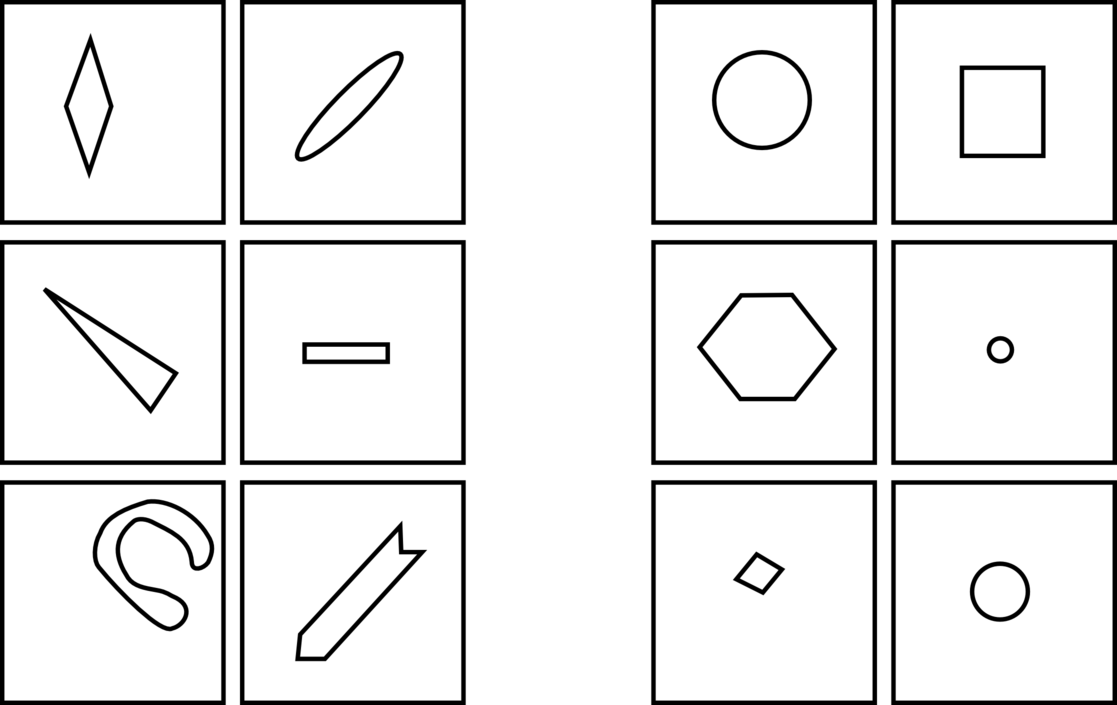}
\end{minipage}
\hfill
\begin{minipage}[t]{0.5\textwidth}
\vspace{0pt}
Samples: 299971\\
Runs: 6\\
Burn-in: 100000\\
Discarded rules with mistakes: 30\\
$p$ = proportion of samples for this rule\\
\end{minipage}

\bigskip

{\noindent\scriptsize
\begin{tabularx}{\textwidth}{Xr}
\hline%---------------------------------------------------------
$\LEFT$ elongated figures                               & $p$ \\
\hline%---------------------------------------------------------
\hline%---------------------------------------------------------
$\RIGHT$ compact figures                                & $p$ \\
\hline%---------------------------------------------------------
$\GREATERLA(\FIGURES,\COMPACTNESS)$                     & .87 \\
$\GREATERLA(\OUTLINE(\FIGURES),\COMPACTNESS)$           & .05 \\
$\GREATERLA(\GET(\FIGURES,\HOLES),\COMPACTNESS)$        & .02 \\
$\EXISTS(\HIGH(\FIGURES,\COMPACTNESS))$                 & .01 \\
$\GREATERLA(\OUTLINE(\OUTLINE(\FIGURES)),\COMPACTNESS)$ & .01 \\
\hline%---------------------------------------------------------
Remaining rules                                         & .05 \\
\hline%---------------------------------------------------------
\end{tabularx}}

\newpage
\section*{BP \#12}
%==================================================================================
% BP #12
%       Samples: 159414 Runs: 6 Burn-in: 100000
%       Discarded rules with mistakes: 140588
%       p = proportion of samples for this rule
%==================================================================================
\begin{minipage}[t]{0.4\textwidth}
\vspace{0pt}
\includegraphics[width=0.95\textwidth]{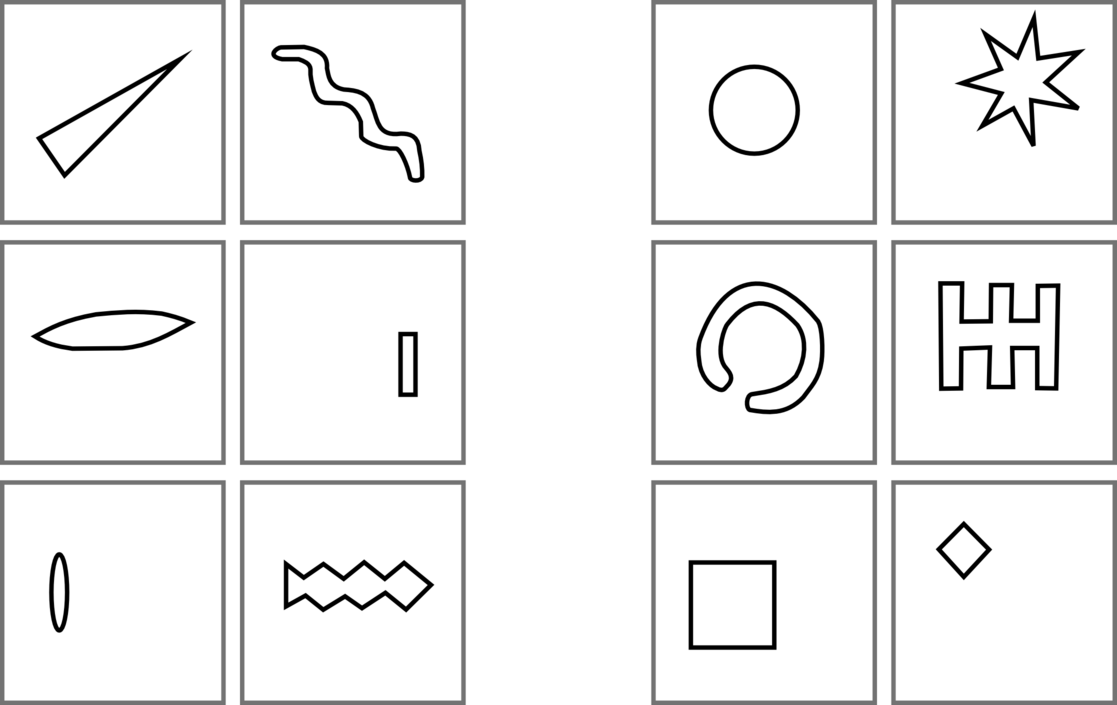}
\end{minipage}
\hfill
\begin{minipage}[t]{0.5\textwidth}
\vspace{0pt}
Samples: 159414\\
Runs: 6\\
Burn-in: 100000\\
Discarded rules with mistakes: 140588\\
$p$ = proportion of samples for this rule\\
\end{minipage}

\bigskip

{\noindent\scriptsize
\begin{tabularx}{\textwidth}{Xr}
\hline%----------------------------------------------------------------------------
$\LEFT$ convex figure shell elongated                                      & $p$ \\
\hline%----------------------------------------------------------------------------
\hline%----------------------------------------------------------------------------
$\RIGHT$ convex figure shell compact                                       & $p$ \\
\hline%----------------------------------------------------------------------------
$\GREATERLA(\GET(\FIGURES,\HULLS),\COMPACTNESS)$                           & .77 \\
$\GREATERLA(\GET(\GET(\FIGURES,\HULLS),\HULLS),\COMPACTNESS)$              & .05 \\
$\GREATERLA(\GET(\OUTLINE(\FIGURES),\HULLS),\COMPACTNESS)$                 & .04 \\
$\GREATERLA(\GET(\GET(\FIGURES,\HULLS),\HOLES),\COMPACTNESS)$              & .03 \\
$\GREATERLA(\GET(\GET(\GET(\FIGURES,\HULLS),\HULLS),\HULLS),\COMPACTNESS)$ & .01 \\
\hline%----------------------------------------------------------------------------
Remaining rules                                                            & .10 \\
\hline%----------------------------------------------------------------------------
\end{tabularx}}

\newpage
\section*{BP \#21}
%===================================================
% BP #21
%       Samples: 299339 Runs: 6 Burn-in: 100000
%       Discarded rules with mistakes: 662
%       p = proportion of samples for this rule
%===================================================
\begin{minipage}[t]{0.4\textwidth}
\vspace{0pt}
\includegraphics[width=0.95\textwidth]{bpimgs/lowres-p0021}
\end{minipage}
\hfill
\begin{minipage}[t]{0.5\textwidth}
\vspace{0pt}
Samples: 299339\\
Runs: 6\\
Burn-in: 100000\\
Discarded rules with mistakes: 662\\
$p$ = proportion of samples for this rule\\
\end{minipage}

\bigskip

{\noindent\scriptsize
\begin{tabularx}{\textwidth}{Xr}
\hline%---------------------------------------------
$\LEFT$ small figure present                & $p$ \\
\hline%---------------------------------------------
$\EXISTS(\SMALL(\FIGURES))$                 & .47 \\
$\EXISTS(\LOW(\FIGURES,\SIZE))$             & .05 \\
$\MORE(\FIGURES,\BIG(\FIGURES))$            & .04 \\
$\EXISTS(\SMALL(\OUTLINE(\FIGURES)))$       & .02 \\
$\EXISTS(\OUTLINE(\SMALL(\FIGURES)))$       & .02 \\
\hline%---------------------------------------------
$\RIGHT$ no small figure present            & $p$ \\
\hline%---------------------------------------------
$\EQUALNUM(\FIGURES,\BIG(\FIGURES))$        & .05 \\
$\EQUALNUM(\BIG(\FIGURES),\FIGURES)$        & .04 \\
$\EQUALNUM(\FIGURES,\HIGH(\FIGURES,\SIZE))$ & .01 \\
$\EQUALNUM(\FIGURES,\BIG(\BIG(\FIGURES)))$  & .00 \\
$\EQUALNUM(\BIG(\BIG(\FIGURES)),\FIGURES)$  & .00 \\
\hline%---------------------------------------------
Remaining rules                             & .28 \\
\hline%---------------------------------------------
\end{tabularx}}

\newpage
\section*{BP \#22}
%===================================================
% BP #22
%       Samples: 299994 Runs: 6 Burn-in: 100000
%       Discarded rules with mistakes: 9
%       p = proportion of samples for this rule
%===================================================
\begin{minipage}[t]{0.4\textwidth}
\vspace{0pt}
\includegraphics[width=0.95\textwidth]{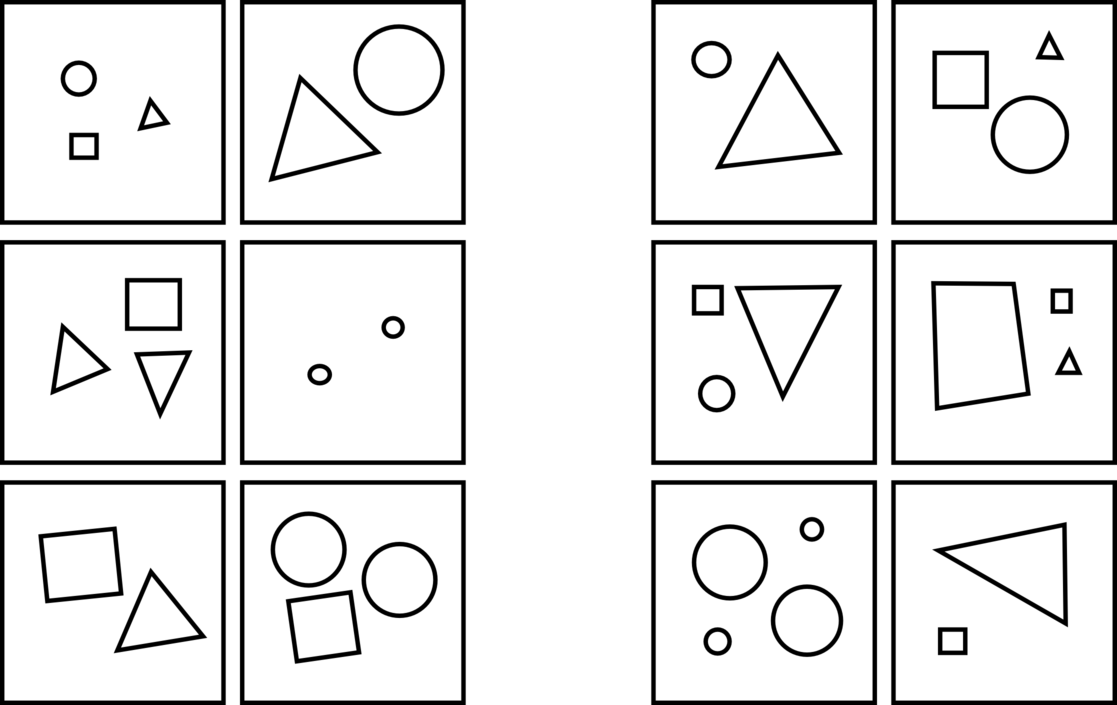}
\end{minipage}
\hfill
\begin{minipage}[t]{0.5\textwidth}
\vspace{0pt}
Samples: 299994\\
Runs: 6\\
Burn-in: 100000\\
Discarded rules with mistakes: 9\\
$p$ = proportion of samples for this rule\\
\end{minipage}

\bigskip

{\noindent\scriptsize
\begin{tabularx}{\textwidth}{Xr}
\hline%---------------------------------------------
$\LEFT$ areas of figures approximately equal& $p$ \\
\hline%---------------------------------------------
$\MORESIMLA(\FIGURES,\SIZE)$                & .82 \\
$\MORESIMLA(\OUTLINE(\FIGURES),\SIZE)$      & .05 \\
$\MORESIMLA(\GET(\FIGURES,\HULLS),\SIZE)$   & .02 \\
$\MORESIMLA(\GET(\FIGURES,\HOLES),\SIZE)$   & .02 \\
$\MORESIMLA(\CUP(\FIGURES,\FIGURES),\SIZE)$ & .00 \\
\hline%---------------------------------------------
$\RIGHT$ areas of figures differ greatly    & $p$ \\
\hline%---------------------------------------------
\hline%---------------------------------------------
Remaining rules                             & .08 \\
\hline%---------------------------------------------
\end{tabularx}}

\newpage
\section*{BP \#23}
%============================================
% BP #23
%       Samples: 299937 Runs: 6 Burn-in: 100000
%       Discarded rules with mistakes: 63
%       p = proportion of samples for this rule
%============================================
\begin{minipage}[t]{0.4\textwidth}
\vspace{0pt}
\includegraphics[width=0.95\textwidth]{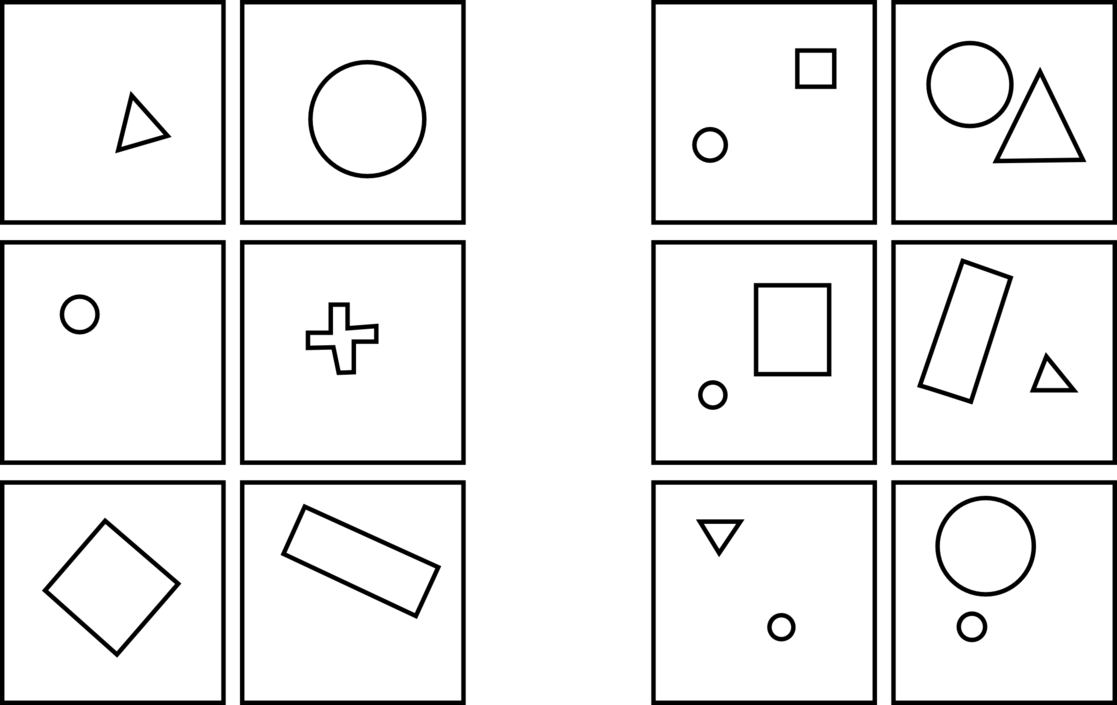}
\end{minipage}
\hfill
\begin{minipage}[t]{0.5\textwidth}
\vspace{0pt}
Samples: 299937\\
Runs: 6\\
Burn-in: 100000\\
Discarded rules with mistakes: 63\\
$p$ = proportion of samples for this rule\\
\end{minipage}

\bigskip

{\noindent\scriptsize
\begin{tabularx}{\textwidth}{Xr}
\hline%--------------------------------------
$\LEFT$ one figure                   & $p$ \\
\hline%--------------------------------------
$\EXACTLY(1,\FIGURES)$               & .40 \\
$\EXACTLY(1,\OUTLINE(\FIGURES))$     & .02 \\
$\EXACTLY(1,\GET(\FIGURES,\HOLES))$  & .01 \\
$\EXACTLY(1,\GET(\FIGURES,\HULLS))$  & .01 \\
$\EXISTS(\LOW(\FIGURES,\DISTANCE))$  & .01 \\
\hline%--------------------------------------
$\RIGHT$ two figures                 & $p$ \\
\hline%--------------------------------------
$\EXACTLY(2,\FIGURES)$               & .40 \\
$\EXACTLY(2,\OUTLINE(\FIGURES))$     & .02 \\
$\EXACTLY(2,\GET(\FIGURES,\HULLS))$  & .01 \\
$\EXACTLY(2,\GET(\FIGURES,\HOLES))$  & .01 \\
$\EXISTS(\HIGH(\FIGURES,\DISTANCE))$ & .01 \\
\hline%--------------------------------------
Remaining rules                      & .10 \\
\hline%--------------------------------------
\end{tabularx}}

\newpage
\section*{BP \#24}
%==================================================================
% BP #24
%       Samples: 299921 Runs: 6 Burn-in: 100000
%       Discarded rules with mistakes: 81
%       p = proportion of samples for this rule
%==================================================================
\begin{minipage}[t]{0.4\textwidth}
\vspace{0pt}
\includegraphics[width=0.95\textwidth]{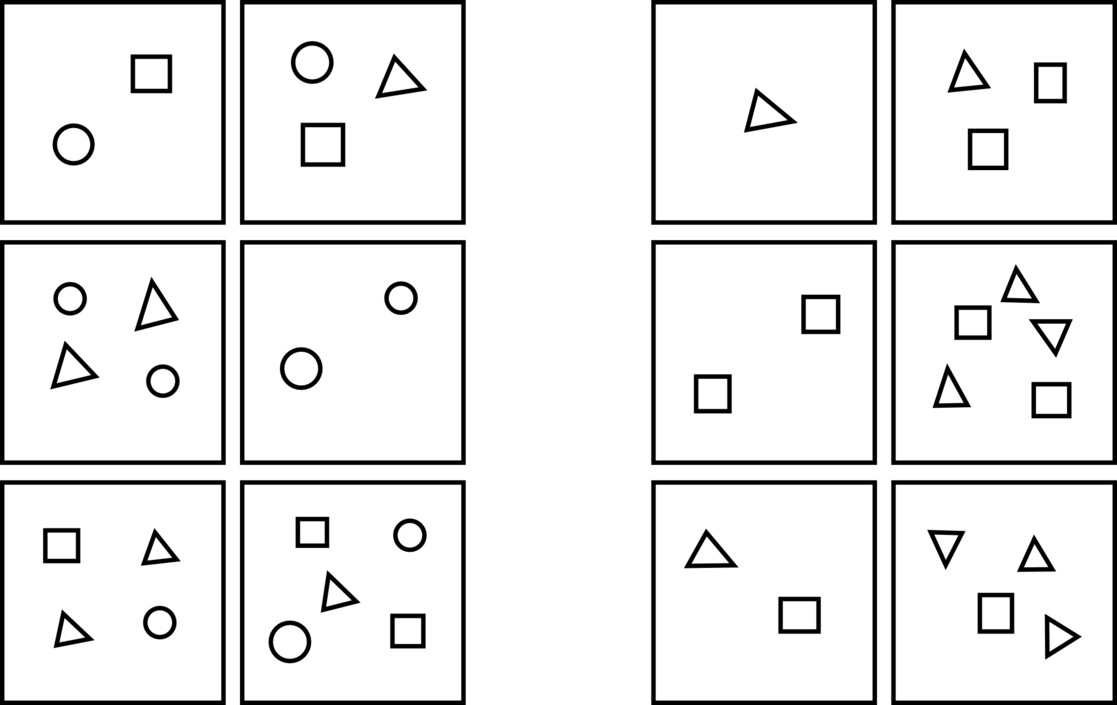}
\end{minipage}
\hfill
\begin{minipage}[t]{0.5\textwidth}
\vspace{0pt}
Samples: 299921\\
Runs: 6\\
Burn-in: 100000\\
Discarded rules with mistakes: 81\\
$p$ = proportion of samples for this rule\\
\end{minipage}

\bigskip

{\noindent\scriptsize
\begin{tabularx}{\textwidth}{Xr}
\hline%------------------------------------------------------------
$\LEFT$ a circle                                           & $p$ \\
\hline%------------------------------------------------------------
$\EXISTS(\CIRCLES)$                                        & .83 \\
$\EXISTS(\OUTLINE(\CIRCLES))$                              & .05 \\
$\EXISTS(\GET(\CIRCLES,\HULLS))$                           & .02 \\
$\EXISTS(\GET(\CIRCLES,\HOLES))$                           & .02 \\
$\EXISTS(\OUTLINE(\OUTLINE(\CIRCLES)))$                    & .01 \\
\hline%------------------------------------------------------------
$\RIGHT$ no circle                                         & $p$ \\
\hline%------------------------------------------------------------
$\EQUALNUM(\FIGURES,\CUP(\FIGURES,\GET(\CIRCLES,\HOLES)))$ & .00 \\
$\EQUALNUM(\FIGURES,\CUP(\FIGURES,\GET(\CIRCLES,\HULLS)))$ & .00 \\
$\EQUALNUM(\CUP(\RECTANGLES,\TRIANGLES),\FIGURES)$         & .00 \\
$\EQUALNUM(\CUP(\TRIANGLES,\RECTANGLES),\FIGURES)$         & .00 \\
$\EQUALNUM(\FIGURES,\CUP(\TRIANGLES,\RECTANGLES))$         & .00 \\
\hline%------------------------------------------------------------
Remaining rules                                            & .08 \\
\hline%------------------------------------------------------------
\end{tabularx}}

\newpage
\section*{BP \#25}
%================================================
% BP #25
%       Samples: 299778 Runs: 6 Burn-in: 100000
%       Discarded rules with mistakes: 225
%       p = proportion of samples for this rule
%================================================
\begin{minipage}[t]{0.4\textwidth}
\vspace{0pt}
\includegraphics[width=0.95\textwidth]{bpimgs/lowres-p0025}
\end{minipage}
\hfill
\begin{minipage}[t]{0.5\textwidth}
\vspace{0pt}
Samples: 299778\\
Runs: 6\\
Burn-in: 100000\\
Discarded rules with mistakes: 225\\
$p$ = proportion of samples for this rule\\
\end{minipage}

\bigskip

{\noindent\scriptsize
\begin{tabularx}{\textwidth}{Xr}
\hline%------------------------------------------
$\LEFT$ black figure is a triangle       & $p$ \\
\hline%------------------------------------------
$\EXISTS(\SOLID(\TRIANGLES))$            & .20 \\
$\EXACTLY(1,\SOLID(\TRIANGLES))$         & .05 \\
$\EXISTS(\SOLID(\SOLID(\TRIANGLES)))$    & .02 \\
$\EXISTS(\HIGH(\TRIANGLES,\COLOR))$      & .02 \\
$\EQUALNUM(\OUTLINE(\CIRCLES),\CIRCLES)$ & .02 \\
\hline%------------------------------------------
$\RIGHT$ black figure is a circle        & $p$ \\
\hline%------------------------------------------
$\EXISTS(\SOLID(\CIRCLES))$              & .19 \\
$\EXACTLY(1,\SOLID(\CIRCLES))$           & .06 \\
$\MORE(\CIRCLES,\OUTLINE(\CIRCLES))$     & .02 \\
$\EXISTS(\SOLID(\SOLID(\CIRCLES)))$      & .02 \\
$\EXISTS(\HIGH(\CIRCLES,\COLOR))$        & .02 \\
\hline%------------------------------------------
Remaining rules                          & .38 \\
\hline%------------------------------------------
\end{tabularx}}

\newpage
\section*{BP \#26}
%=======================================================
% BP #26
%       Samples: 298585 Runs: 6 Burn-in: 100000
%       Discarded rules with mistakes: 1415
%       p = proportion of samples for this rule
%=======================================================
\begin{minipage}[t]{0.4\textwidth}
\vspace{0pt}
\includegraphics[width=0.95\textwidth]{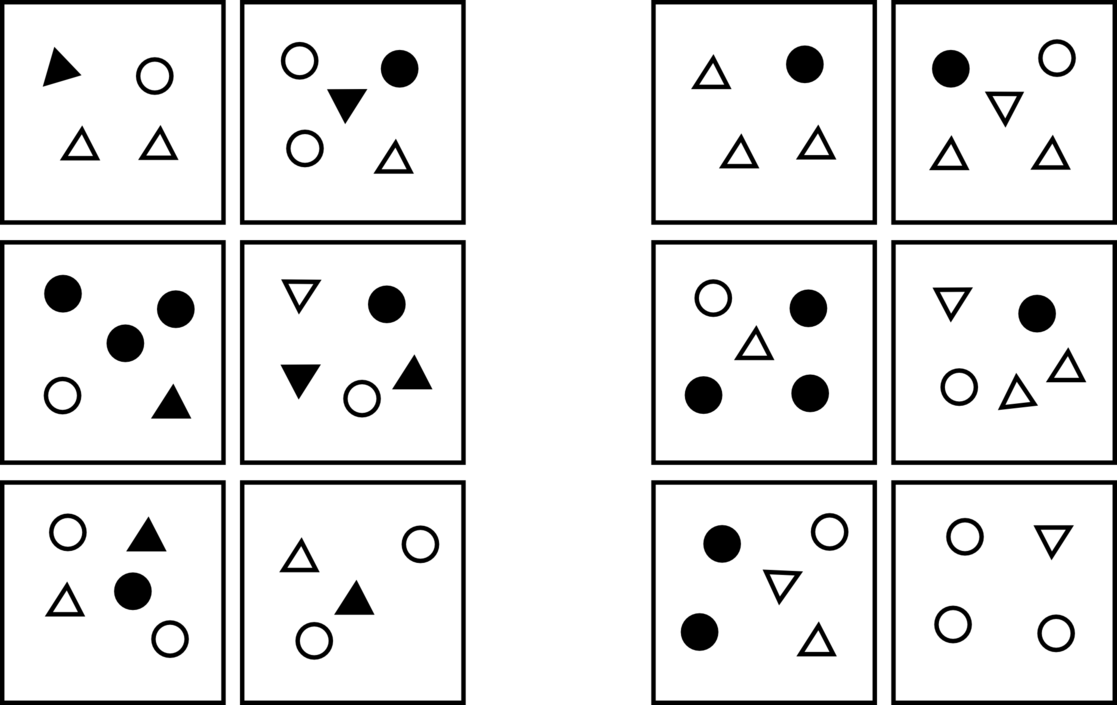}
\end{minipage}
\hfill
\begin{minipage}[t]{0.5\textwidth}
\vspace{0pt}
Samples: 298585\\
Runs: 6\\
Burn-in: 100000\\
Discarded rules with mistakes: 1415\\
$p$ = proportion of samples for this rule\\
\end{minipage}

\bigskip

{\noindent\scriptsize
\begin{tabularx}{\textwidth}{Xr}
\hline%-------------------------------------------------
$\LEFT$ solid black triangle                    & $p$ \\
\hline%-------------------------------------------------
$\EXISTS(\SOLID(\TRIANGLES))$                   & .42 \\
$\MORE(\TRIANGLES,\OUTLINE(\TRIANGLES))$        & .05 \\
$\EXISTS(\SOLID(\SOLID(\TRIANGLES)))$           & .04 \\
$\EXISTS(\HIGH(\TRIANGLES,\COLOR))$             & .03 \\
$\MORE(\TRIANGLES,\GET(\TRIANGLES,\HOLES))$     & .02 \\
\hline%-------------------------------------------------
$\RIGHT$ no solid black triangle                & $p$ \\
\hline%-------------------------------------------------
$\EQUALNUM(\OUTLINE(\TRIANGLES),\TRIANGLES)$    & .04 \\
$\EQUALNUM(\TRIANGLES,\OUTLINE(\TRIANGLES))$    & .04 \\
$\EQUALNUM(\GET(\TRIANGLES,\HOLES),\TRIANGLES)$ & .02 \\
$\EQUALNUM(\TRIANGLES,\GET(\TRIANGLES,\HOLES))$ & .02 \\
$\EQUALNUM(\LOW(\TRIANGLES,\COLOR),\TRIANGLES)$ & .01 \\
\hline%-------------------------------------------------
Remaining rules                                 & .31 \\
\hline%-------------------------------------------------
\end{tabularx}}

\newpage
\section*{BP \#27}
%=================================================================
% BP #27
%       Samples: 288897 Runs: 6 Burn-in: 100000
%       Discarded rules with mistakes: 11103
%       p = proportion of samples for this rule
%=================================================================
\begin{minipage}[t]{0.4\textwidth}
\vspace{0pt}
\includegraphics[width=0.95\textwidth]{bpimgs/lowres-p0027}
\end{minipage}
\hfill
\begin{minipage}[t]{0.5\textwidth}
\vspace{0pt}
Samples: 288897\\
Runs: 6\\
Burn-in: 100000\\
Discarded rules with mistakes: 11103\\
$p$ = proportion of samples for this rule\\
\end{minipage}

\bigskip

{\noindent\scriptsize
\begin{tabularx}{\textwidth}{Xr}
\hline%-----------------------------------------------------------
$\LEFT$ more solid black figures                          & $p$ \\
\hline%-----------------------------------------------------------
$\MORESIMLA(\CUP(\GET(\FIGURES,\HULLS),\FIGURES),\COLOR)$ & .02 \\
$\MORESIMLA(\CUP(\GET(\FIGURES,\HOLES),\FIGURES),\COLOR)$ & .02 \\
$\MORE(\SOLID(\FIGURES),\OUTLINE(\FIGURES))$              & .02 \\
$\MORESIMLA(\CUP(\FIGURES,\GET(\FIGURES,\HOLES)),\COLOR)$ & .02 \\
$\MORESIMLA(\CUP(\FIGURES,\GET(\FIGURES,\HULLS)),\COLOR)$ & .02 \\
\hline%-----------------------------------------------------------
$\RIGHT$ more outline figures                             & $p$ \\
\hline%-----------------------------------------------------------
$\MORE(\OUTLINE(\FIGURES),\CIRCLES)$                      & .21 \\
$\MORE(\GET(\FIGURES,\HOLES),\CIRCLES)$                   & .09 \\
$\MORE(\OUTLINE(\FIGURES),\SOLID(\FIGURES))$              & .02 \\
$\MORE(\OUTLINE(\OUTLINE(\FIGURES)),\CIRCLES)$            & .02 \\
$\MORE(\LOW(\FIGURES,\COLOR),\CIRCLES)$                   & .02 \\
\hline%-----------------------------------------------------------
Remaining rules                                           & .56 \\
\hline%-----------------------------------------------------------
\end{tabularx}}

\newpage
\section*{BP \#28}
%==============================================================
% BP #28
%       Samples: 240756 Runs: 6 Burn-in: 100000
%       Discarded rules with mistakes: 59244
%       p = proportion of samples for this rule
%==============================================================
\begin{minipage}[t]{0.4\textwidth}
\vspace{0pt}
\includegraphics[width=0.95\textwidth]{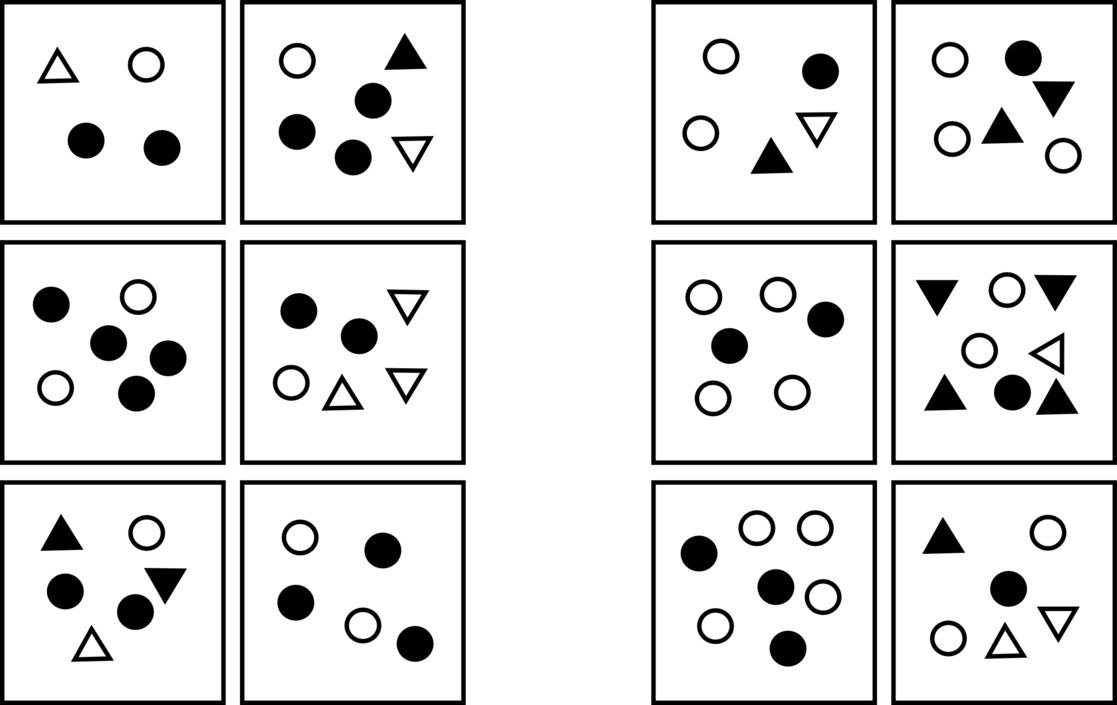}
\end{minipage}
\hfill
\begin{minipage}[t]{0.5\textwidth}
\vspace{0pt}
Samples: 240756\\
Runs: 6\\
Burn-in: 100000\\
Discarded rules with mistakes: 59244\\
$p$ = proportion of samples for this rule\\
\end{minipage}

\bigskip

{\noindent\scriptsize
\begin{tabularx}{\textwidth}{Xr}
\hline%--------------------------------------------------------
$\LEFT$ more solid black circles                       & $p$ \\
\hline%--------------------------------------------------------
$\MORE(\SOLID(\CIRCLES),\OUTLINE(\CIRCLES))$           & .07 \\
$\MORE(\SOLID(\CIRCLES),\GET(\CIRCLES,\HOLES))$        & .03 \\
$\MORE(\HIGH(\CIRCLES,\COLOR),\OUTLINE(\CIRCLES))$     & .01 \\
$\MORE(\SOLID(\SOLID(\CIRCLES)),\OUTLINE(\CIRCLES))$   & .01 \\
$\MORE(\SOLID(\CIRCLES),\OUTLINE(\OUTLINE(\CIRCLES)))$ & .01 \\
\hline%--------------------------------------------------------
$\RIGHT$ more outline circles                          & $p$ \\
\hline%--------------------------------------------------------
$\MORE(\OUTLINE(\CIRCLES),\SOLID(\CIRCLES))$           & .24 \\
$\MORE(\GET(\CIRCLES,\HOLES),\SOLID(\CIRCLES))$        & .10 \\
$\MORE(\OUTLINE(\CIRCLES),\SOLID(\SOLID(\CIRCLES)))$   & .02 \\
$\MORE(\OUTLINE(\OUTLINE(\CIRCLES)),\SOLID(\CIRCLES))$ & .02 \\
$\MORE(\OUTLINE(\CIRCLES),\HIGH(\CIRCLES,\COLOR))$     & .02 \\
\hline%--------------------------------------------------------
Remaining rules                                        & .47 \\
\hline%--------------------------------------------------------
\end{tabularx}}

\newpage
\section*{BP \#29}
%==========================================================================================
% BP #29
%       Samples: 141261 Runs: 6 Burn-in: 100000
%       Discarded rules with mistakes: 158740
%       p = proportion of samples for this rule
%==========================================================================================
\begin{minipage}[t]{0.4\textwidth}
\vspace{0pt}
\includegraphics[width=0.95\textwidth]{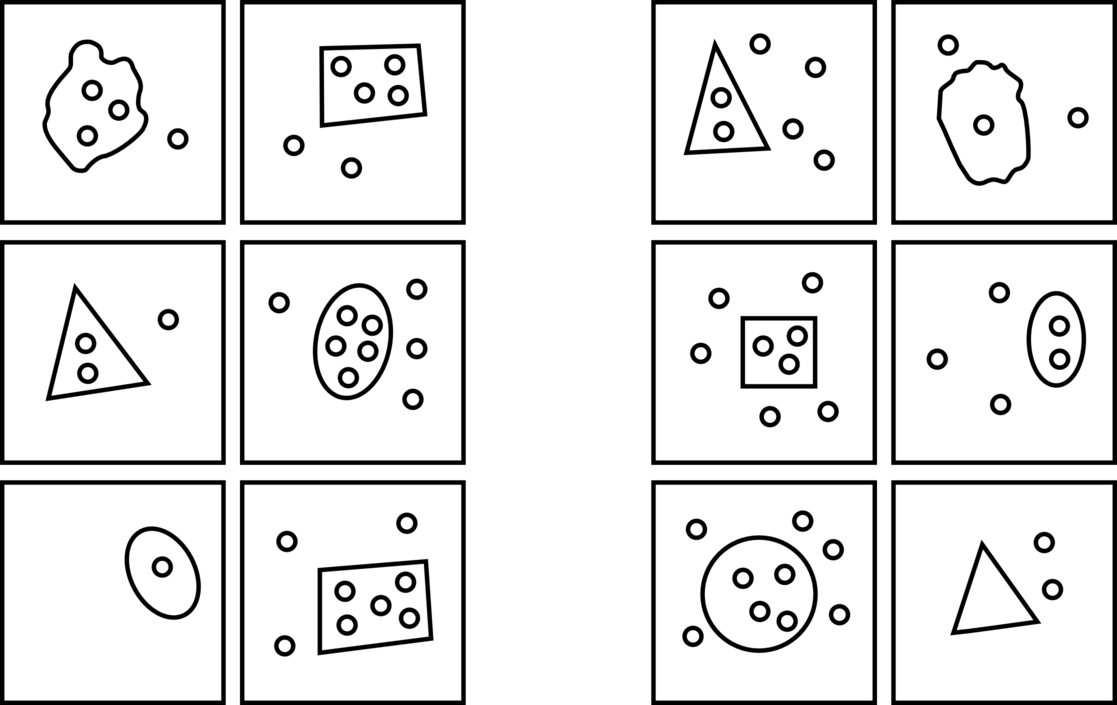}
\end{minipage}
\hfill
\begin{minipage}[t]{0.5\textwidth}
\vspace{0pt}
Samples: 141261\\
Runs: 6\\
Burn-in: 100000\\
Discarded rules with mistakes: 158740\\
$p$ = proportion of samples for this rule\\
\end{minipage}

\bigskip

{\noindent\scriptsize
\begin{tabularx}{\textwidth}{Xr}
\hline%------------------------------------------------------------------------------------
$\LEFT$ there are more small circles inside the figure outline than outside        & $p$ \\
\hline%------------------------------------------------------------------------------------
$\MORE(\INSIDE(\FIGURES),\SETMINUS(\SMALL(\FIGURES),\INSIDE(\FIGURES)))$           & .03 \\
$\MORE(\INSIDE(\FIGURES),\SETMINUS(\SMALL(\CIRCLES),\INSIDE(\FIGURES)))$           & .01 \\
$\MORE(\INSIDE(\FIGURES),\SETMINUS(\LOW(\FIGURES,\SIZE),\INSIDE(\FIGURES)))$       & .00 \\
$\MORE(\INSIDE(\FIGURES),\SETMINUS(\OUTLINE(\SMALL(\FIGURES)),\INSIDE(\FIGURES)))$ & .00 \\
$\MORE(\INSIDE(\FIGURES),\SETMINUS(\SMALL(\FIGURES),\INSIDE(\BIG(\FIGURES))))$     & .00 \\
\hline%------------------------------------------------------------------------------------
$\RIGHT$ there are fewer small circles inside the figure outline than outside      & $p$ \\
\hline%------------------------------------------------------------------------------------
$\MORE(\SETMINUS(\FIGURES,\INSIDE(\FIGURES)),\INSIDE(\FIGURES))$                   & .07 \\
$\MORE(\SETMINUS(\CIRCLES,\INSIDE(\FIGURES)),\INSIDE(\FIGURES))$                   & .02 \\
$\MORE(\CUP(\TRIANGLES,\SETMINUS(\FIGURES,\INSIDE(\FIGURES))),\INSIDE(\FIGURES))$  & .00 \\
$\MORE(\CUP(\RECTANGLES,\SETMINUS(\FIGURES,\INSIDE(\FIGURES))),\INSIDE(\FIGURES))$ & .00 \\
$\MORE(\SETMINUS(\FIGURES,\INSIDE(\FIGURES)),\INSIDE(\BIG(\FIGURES)))$             & .00 \\
\hline%------------------------------------------------------------------------------------
Remaining rules                                                                    & .84 \\
\hline%------------------------------------------------------------------------------------
\end{tabularx}}

\newpage
\section*{BP \#34}
%==============================================================
% BP #34
%       Samples: 298520 Runs: 6 Burn-in: 100000
%       Discarded rules with mistakes: 1480
%       p = proportion of samples for this rule
%==============================================================
\begin{minipage}[t]{0.4\textwidth}
\vspace{0pt}
\includegraphics[width=0.95\textwidth]{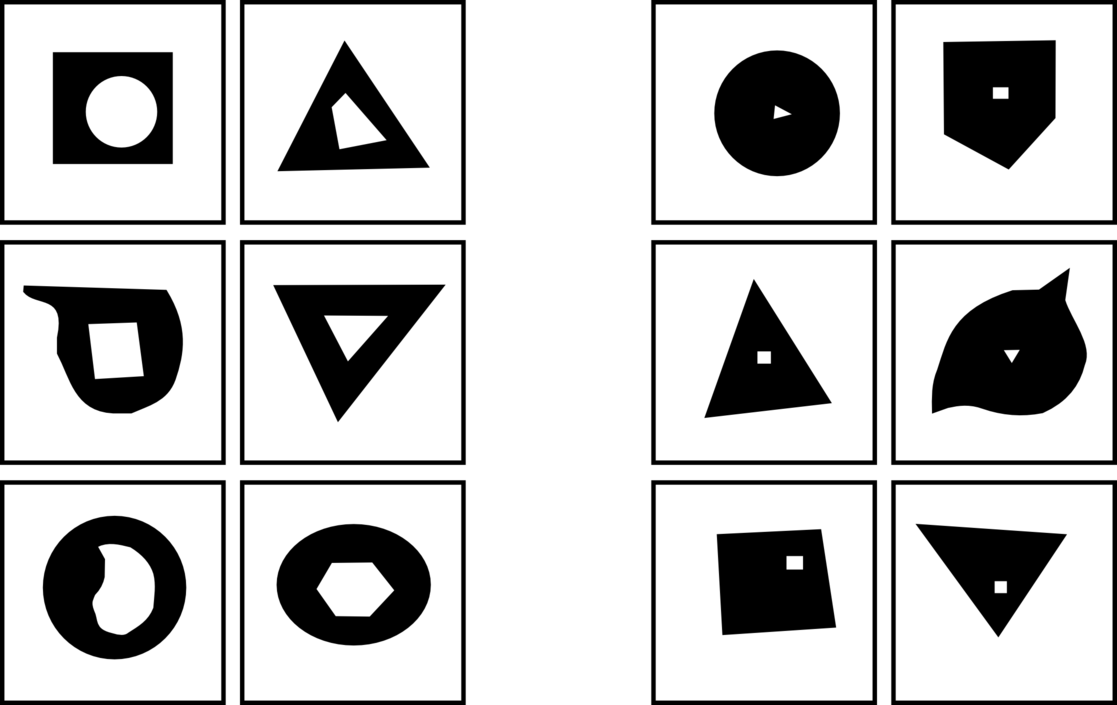}
\end{minipage}
\hfill
\begin{minipage}[t]{0.5\textwidth}
\vspace{0pt}
Samples: 298520\\
Runs: 6\\
Burn-in: 100000\\
Discarded rules with mistakes: 1480\\
$p$ = proportion of samples for this rule\\
\end{minipage}

\bigskip

{\noindent\scriptsize
\begin{tabularx}{\textwidth}{Xr}
\hline%--------------------------------------------------------
$\LEFT$ a large hole                                   & $p$ \\
\hline%--------------------------------------------------------
$\GREATERLA(\GET(\FIGURES,\HOLES),\SIZE)$              & .39 \\
$\EXISTS(\BIG(\GET(\FIGURES,\HOLES)))$                 & .10 \\
$\EXACTLY(1,\BIG(\GET(\FIGURES,\HOLES)))$              & .04 \\
$\GREATERLA(\GET(\GET(\FIGURES,\HOLES),\HOLES),\SIZE)$ & .03 \\
$\GREATERLA(\GET(\OUTLINE(\FIGURES),\HOLES),\SIZE)$    & .02 \\
\hline%--------------------------------------------------------
$\RIGHT$ a small hole                                  & $p$ \\
\hline%--------------------------------------------------------
$\EXISTS(\SMALL(\GET(\FIGURES,\HOLES)))$               & .07 \\
$\EXACTLY(1,\SMALL(\GET(\FIGURES,\HOLES)))$            & .06 \\
$\EXISTS(\SMALL(\SMALL(\GET(\FIGURES,\HOLES))))$       & .01 \\
$\EXISTS(\LOW(\GET(\FIGURES,\HOLES),\SIZE))$           & .01 \\
$\EXACTLY(1,\SMALL(\SMALL(\GET(\FIGURES,\HOLES))))$    & .01 \\
\hline%--------------------------------------------------------
Remaining rules                                        & .27 \\
\hline%--------------------------------------------------------
\end{tabularx}}

\newpage
\section*{BP \#35}
%====================================================================================
% BP #35
%       Samples: 141435 Runs: 6 Burn-in: 100000
%       Discarded rules with mistakes: 158566
%       p = proportion of samples for this rule
%====================================================================================
\begin{minipage}[t]{0.4\textwidth}
\vspace{0pt}
\includegraphics[width=0.95\textwidth]{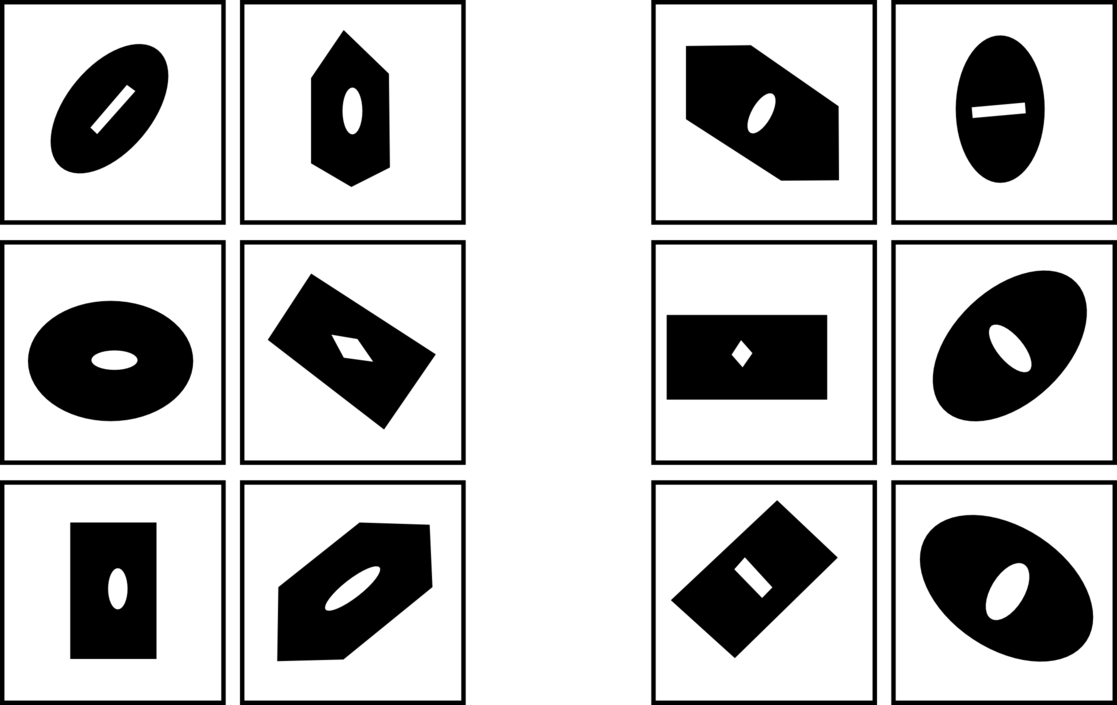}
\end{minipage}
\hfill
\begin{minipage}[t]{0.5\textwidth}
\vspace{0pt}
Samples: 141435\\
Runs: 6\\
Burn-in: 100000\\
Discarded rules with mistakes: 158566\\
$p$ = proportion of samples for this rule\\
\end{minipage}

\bigskip

{\noindent\scriptsize
\begin{tabularx}{\textwidth}{Xr}
\hline%------------------------------------------------------------------------------
$\LEFT$ the axis of the hole is parallel to the figure axis                  & $p$ \\
\hline%------------------------------------------------------------------------------
$\MORESIMLA(\CUP(\FIGURES,\GET(\FIGURES,\HOLES)),\ORIENTATION)$              & .29 \\
$\MORESIMLA(\CUP(\GET(\FIGURES,\HOLES),\FIGURES),\ORIENTATION)$              & .29 \\
$\MORESIMLA(\CUP(\FIGURES,\GET(\GET(\FIGURES,\HOLES),\HOLES)),\ORIENTATION)$ & .02 \\
$\MORESIMLA(\CUP(\GET(\GET(\FIGURES,\HOLES),\HOLES),\FIGURES),\ORIENTATION)$ & .02 \\
$\MORESIMLA(\CUP(\FIGURES,\GET(\OUTLINE(\FIGURES),\HOLES)),\ORIENTATION)$    & .01 \\
\hline%------------------------------------------------------------------------------
$\RIGHT$ the axis of the hole is perpendicular to the figure axis            & $p$ \\
\hline%------------------------------------------------------------------------------
\hline%------------------------------------------------------------------------------
Remaining rules                                                              & .38 \\
\hline%------------------------------------------------------------------------------
\end{tabularx}}

\newpage
\section*{BP \#36}
%==========================================================
% BP #36
%       Samples: 229631 Runs: 6 Burn-in: 100000
%       Discarded rules with mistakes: 70369
%       p = proportion of samples for this rule
%==========================================================
\begin{minipage}[t]{0.4\textwidth}
\vspace{0pt}
\includegraphics[width=0.95\textwidth]{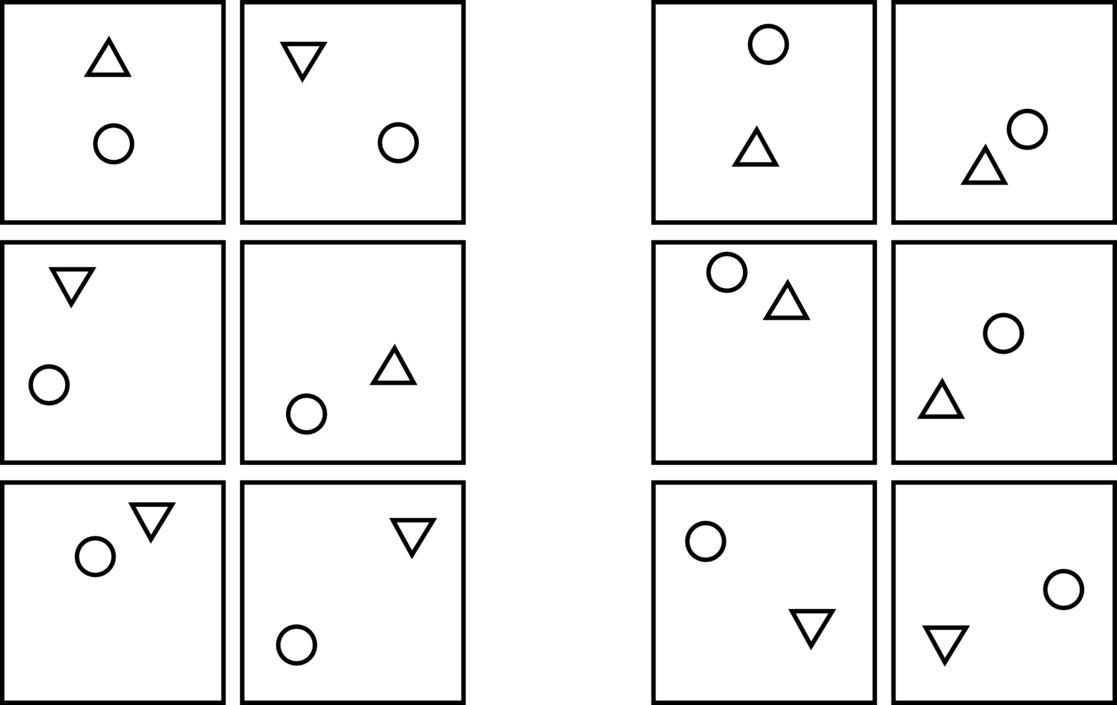}
\end{minipage}
\hfill
\begin{minipage}[t]{0.5\textwidth}
\vspace{0pt}
Samples: 229631\\
Runs: 6\\
Burn-in: 100000\\
Discarded rules with mistakes: 70369\\
$p$ = proportion of samples for this rule\\
\end{minipage}

\bigskip

{\noindent\scriptsize
\begin{tabularx}{\textwidth}{Xr}
\hline%----------------------------------------------------
$\LEFT$ triangle above circle                      & $p$ \\
\hline%----------------------------------------------------
$\GREATERLLA(\CIRCLES,\TRIANGLES,\YPOS)$           & .29 \\
$\GREATERLLA(\CIRCLES,\SMALL(\FIGURES),\YPOS)$     & .02 \\
$\GREATERLLA(\CIRCLES,\OUTLINE(\TRIANGLES),\YPOS)$ & .02 \\
$\GREATERLLA(\OUTLINE(\CIRCLES),\TRIANGLES,\YPOS)$ & .02 \\
$\GREATERLLA(\BIG(\FIGURES),\TRIANGLES,\YPOS)$     & .01 \\
\hline%----------------------------------------------------
$\RIGHT$ circle above triangle                     & $p$ \\
\hline%----------------------------------------------------
$\GREATERLLA(\TRIANGLES,\CIRCLES,\YPOS)$           & .31 \\
$\GREATERLLA(\TRIANGLES,\OUTLINE(\CIRCLES),\YPOS)$ & .02 \\
$\GREATERLLA(\TRIANGLES,\BIG(\FIGURES),\YPOS)$     & .02 \\
$\GREATERLLA(\OUTLINE(\TRIANGLES),\CIRCLES,\YPOS)$ & .02 \\
$\GREATERLLA(\SMALL(\FIGURES),\CIRCLES,\YPOS)$     & .02 \\
\hline%----------------------------------------------------
Remaining rules                                    & .28 \\
\hline%----------------------------------------------------
\end{tabularx}}

\newpage
\section*{BP \#37}
%=============================================================
% BP #37
%       Samples: 231512 Runs: 6 Burn-in: 100000
%       Discarded rules with mistakes: 68489
%       p = proportion of samples for this rule
%=============================================================
\begin{minipage}[t]{0.4\textwidth}
\vspace{0pt}
\includegraphics[width=0.95\textwidth]{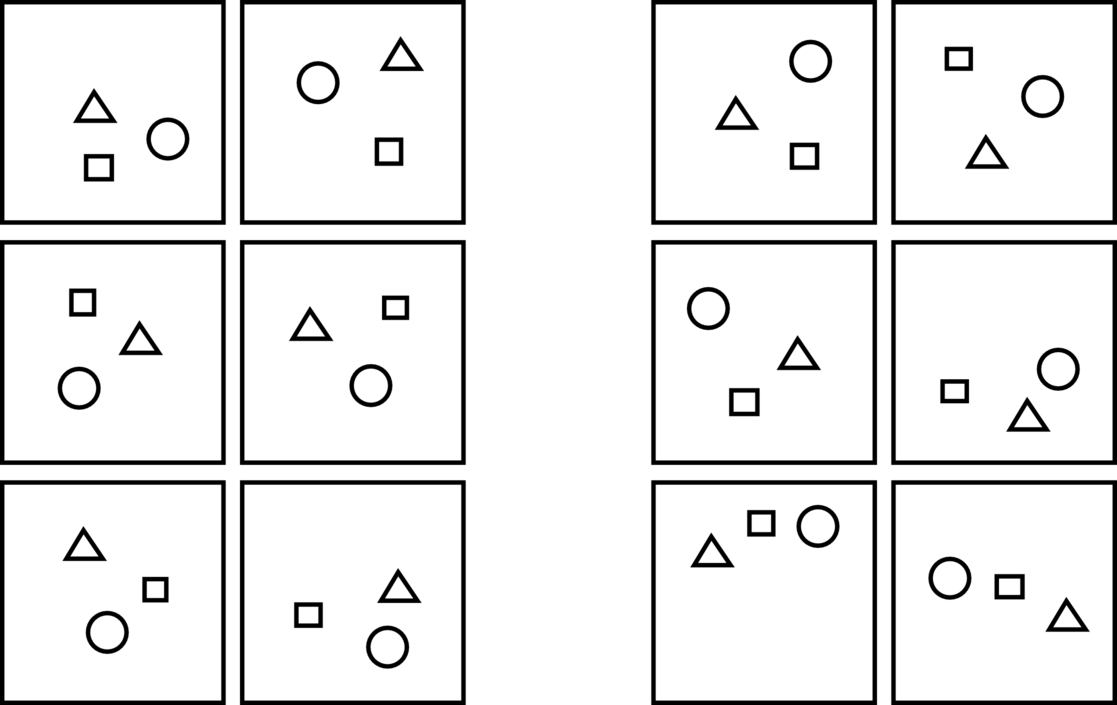}
\end{minipage}
\hfill
\begin{minipage}[t]{0.5\textwidth}
\vspace{0pt}
Samples: 231512\\
Runs: 6\\
Burn-in: 100000\\
Discarded rules with mistakes: 68489\\
$p$ = proportion of samples for this rule\\
\end{minipage}

\bigskip

{\noindent\scriptsize
\begin{tabularx}{\textwidth}{Xr}
\hline%-------------------------------------------------------
$\LEFT$ triangle above circle                         & $p$ \\
\hline%-------------------------------------------------------
$\GREATERLLA(\CIRCLES,\TRIANGLES,\YPOS)$              & .31 \\
$\GREATERLLA(\OUTLINE(\CIRCLES),\TRIANGLES,\YPOS)$    & .02 \\
$\GREATERLLA(\CIRCLES,\OUTLINE(\TRIANGLES),\YPOS)$    & .02 \\
$\GREATERLLA(\BIG(\FIGURES),\TRIANGLES,\YPOS)$        & .02 \\
$\GREATERLLA(\CIRCLES,\GET(\TRIANGLES,\HOLES),\YPOS)$ & .01 \\
\hline%-------------------------------------------------------
$\RIGHT$ circle above triangle                        & $p$ \\
\hline%-------------------------------------------------------
$\GREATERLLA(\TRIANGLES,\CIRCLES,\YPOS)$              & .33 \\
$\GREATERLLA(\OUTLINE(\TRIANGLES),\CIRCLES,\YPOS)$    & .02 \\
$\GREATERLLA(\TRIANGLES,\BIG(\FIGURES),\YPOS)$        & .02 \\
$\GREATERLLA(\TRIANGLES,\OUTLINE(\CIRCLES),\YPOS)$    & .02 \\
$\GREATERLLA(\GET(\TRIANGLES,\HOLES),\CIRCLES,\YPOS)$ & .01 \\
\hline%-------------------------------------------------------
Remaining rules                                       & .24 \\
\hline%-------------------------------------------------------
\end{tabularx}}

\newpage
\section*{BP \#38}
%=============================================================
% BP #38
%       Samples: 225658 Runs: 6 Burn-in: 100000
%       Discarded rules with mistakes: 74343
%       p = proportion of samples for this rule
%=============================================================
\begin{minipage}[t]{0.4\textwidth}
\vspace{0pt}
\includegraphics[width=0.95\textwidth]{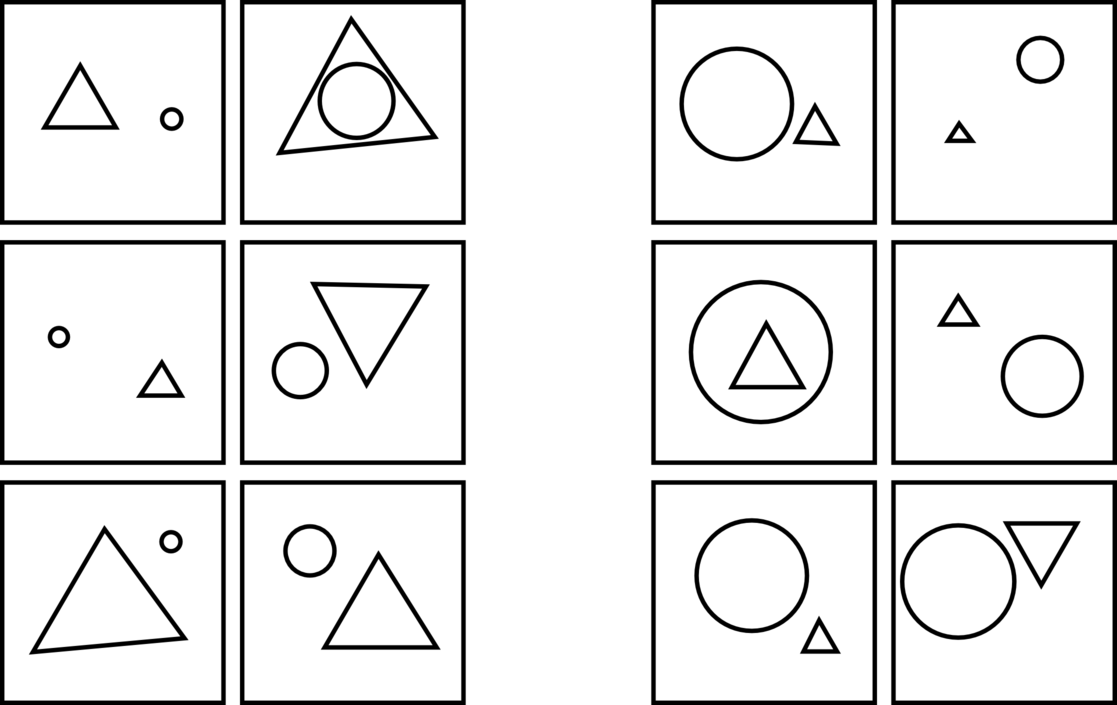}
\end{minipage}
\hfill
\begin{minipage}[t]{0.5\textwidth}
\vspace{0pt}
Samples: 225658\\
Runs: 6\\
Burn-in: 100000\\
Discarded rules with mistakes: 74343\\
$p$ = proportion of samples for this rule\\
\end{minipage}

\bigskip

{\noindent\scriptsize
\begin{tabularx}{\textwidth}{Xr}
\hline%-------------------------------------------------------
$\LEFT$ triangle larger than circle                   & $p$ \\
\hline%-------------------------------------------------------
$\GREATERLLA(\TRIANGLES,\CIRCLES,\SIZE)$              & .25 \\
$\GREATERLLA(\TRIANGLES,\OUTLINE(\CIRCLES),\SIZE)$    & .01 \\
$\GREATERLLA(\OUTLINE(\TRIANGLES),\CIRCLES,\SIZE)$    & .01 \\
$\EXISTS(\SETMINUS(\BIG(\FIGURES),\CIRCLES))$         & .01 \\
$\GREATERLLA(\GET(\TRIANGLES,\HOLES),\CIRCLES,\SIZE)$ & .01 \\
\hline%-------------------------------------------------------
$\RIGHT$ triangle smaller than circle                 & $p$ \\
\hline%-------------------------------------------------------
$\GREATERLLA(\CIRCLES,\TRIANGLES,\SIZE)$              & .34 \\
$\GREATERLLA(\CIRCLES,\OUTLINE(\TRIANGLES),\SIZE)$    & .02 \\
$\GREATERLLA(\OUTLINE(\CIRCLES),\TRIANGLES,\SIZE)$    & .02 \\
$\GREATERLLA(\FIGURES,\GET(\TRIANGLES,\HOLES),\SIZE)$ & .01 \\
$\GREATERLLA(\CIRCLES,\GET(\TRIANGLES,\HOLES),\SIZE)$ & .01 \\
\hline%-------------------------------------------------------
Remaining rules                                       & .32 \\
\hline%-------------------------------------------------------
\end{tabularx}}

\newpage
\section*{BP \#39}
%=====================================================================
% BP #39
%       Samples: 280166 Runs: 6 Burn-in: 100000
%       Discarded rules with mistakes: 19834
%       p = proportion of samples for this rule
%=====================================================================
\begin{minipage}[t]{0.4\textwidth}
\vspace{0pt}
\includegraphics[width=0.95\textwidth]{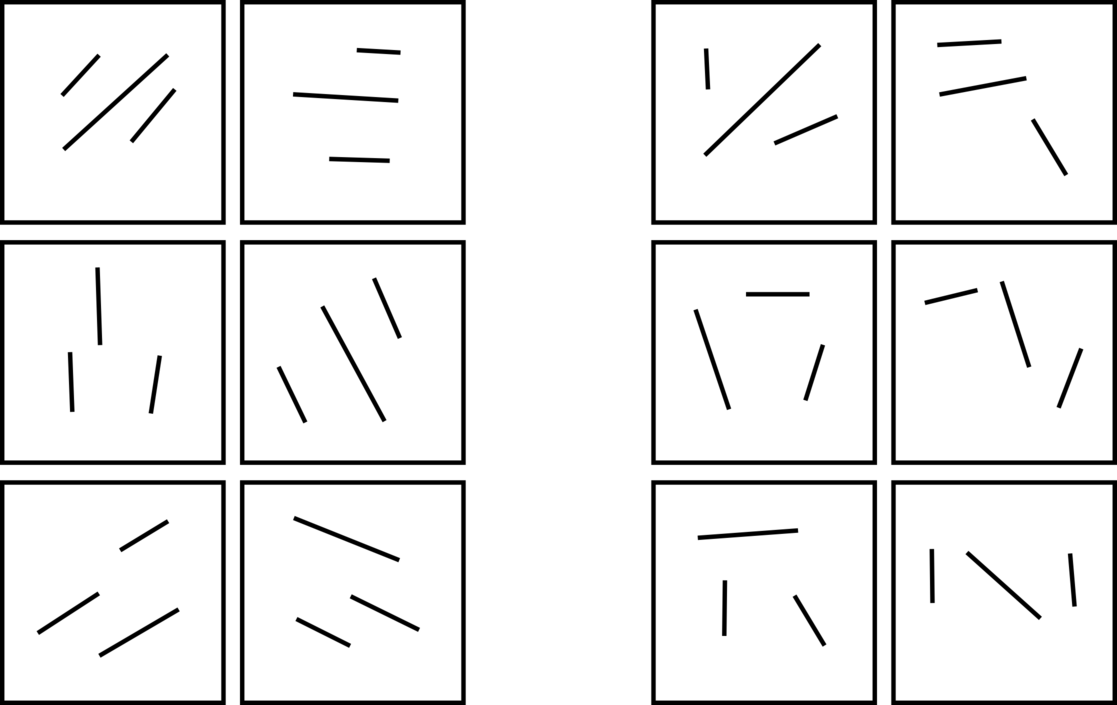}
\end{minipage}
\hfill
\begin{minipage}[t]{0.5\textwidth}
\vspace{0pt}
Samples: 280166\\
Runs: 6\\
Burn-in: 100000\\
Discarded rules with mistakes: 19834\\
$p$ = proportion of samples for this rule\\
\end{minipage}

\bigskip

{\noindent\scriptsize
\begin{tabularx}{\textwidth}{Xr}
\hline%---------------------------------------------------------------
$\LEFT$ segments almost parallel to each other                & $p$ \\
\hline%---------------------------------------------------------------
$\MORESIMLA(\FIGURES,\ORIENTATION)$                           & .74 \\
$\MORESIMLA(\GET(\FIGURES,\HULLS),\ORIENTATION)$              & .09 \\
$\MORESIMLA(\SOLID(\FIGURES),\ORIENTATION)$                   & .04 \\
$\MORESIMLA(\BIG(\FIGURES),\ORIENTATION)$                     & .02 \\
$\MORESIMLA(\GET(\GET(\FIGURES,\HULLS),\HULLS),\ORIENTATION)$ & .01 \\
\hline%---------------------------------------------------------------
$\RIGHT$ large angles between segments                        & $p$ \\
\hline%---------------------------------------------------------------
\hline%---------------------------------------------------------------
Remaining rules                                               & .10 \\
\hline%---------------------------------------------------------------
\end{tabularx}}

\newpage
\section*{BP \#40}
%====================================================================
% BP #40
%       Samples: 299433 Runs: 6 Burn-in: 100000
%       Discarded rules with mistakes: 568
%       p = proportion of samples for this rule
%====================================================================
\begin{minipage}[t]{0.4\textwidth}
\vspace{0pt}
\includegraphics[width=0.95\textwidth]{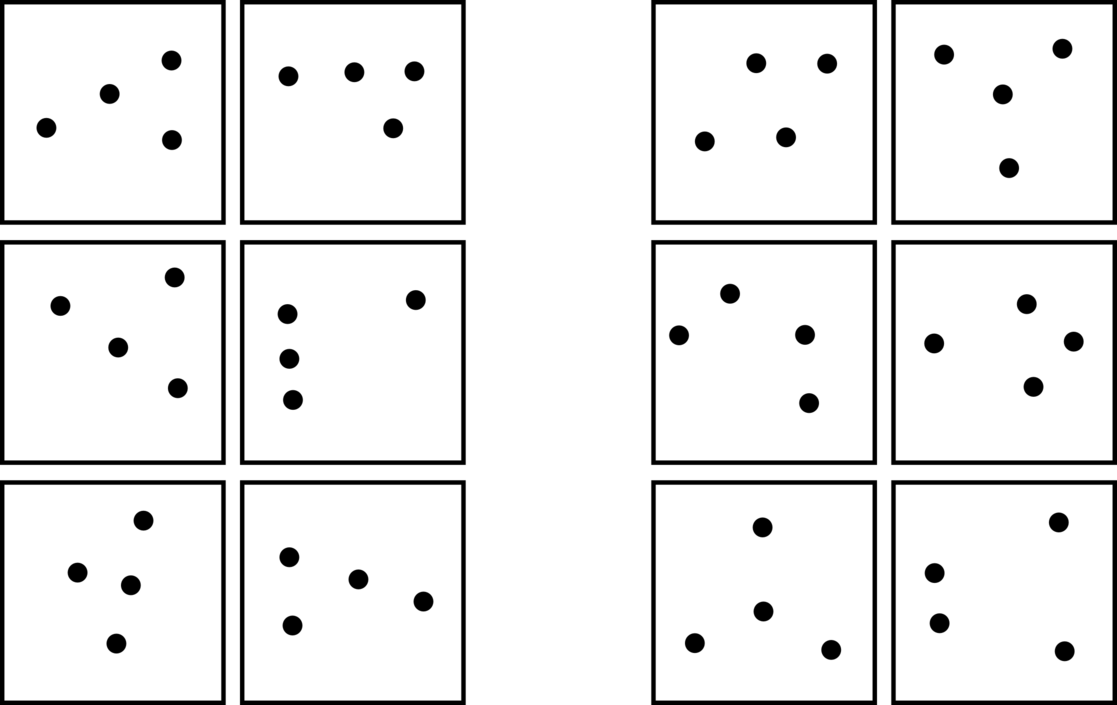}
\end{minipage}
\hfill
\begin{minipage}[t]{0.5\textwidth}
\vspace{0pt}
Samples: 299433\\
Runs: 6\\
Burn-in: 100000\\
Discarded rules with mistakes: 568\\
$p$ = proportion of samples for this rule\\
\end{minipage}

\bigskip

{\noindent\scriptsize
\begin{tabularx}{\textwidth}{Xr}
\hline%--------------------------------------------------------------
$\LEFT$ three points on a straight line                      & $p$ \\
\hline%--------------------------------------------------------------
$\EXISTS(\ALIGNED(\FIGURES))$                                & .26 \\
$\EXISTS(\ALIGNED(\CIRCLES))$                                & .26 \\
$\EXACTLY(3,\ALIGNED(\CIRCLES))$                             & .06 \\
$\EXACTLY(3,\ALIGNED(\FIGURES))$                             & .06 \\
$\EXISTS(\ALIGNED(\ALIGNED(\CIRCLES)))$                      & .03 \\
\hline%--------------------------------------------------------------
$\RIGHT$ no three points on a straight line                  & $p$ \\
\hline%--------------------------------------------------------------
$\EQUALNUM(\FIGURES,\SETMINUS(\FIGURES,\ALIGNED(\FIGURES)))$ & .00 \\
$\EQUALNUM(\CIRCLES,\SETMINUS(\CIRCLES,\ALIGNED(\CIRCLES)))$ & .00 \\
$\EXACTLY(4,\BIG(\SETMINUS(\FIGURES,\ALIGNED(\FIGURES))))$   & .00 \\
$\EQUALNUM(\FIGURES,\SETMINUS(\CIRCLES,\ALIGNED(\CIRCLES)))$ & .00 \\
$\EQUALNUM(\FIGURES,\SETMINUS(\FIGURES,\ALIGNED(\CIRCLES)))$ & .00 \\
\hline%--------------------------------------------------------------
Remaining rules                                              & .33 \\
\hline%--------------------------------------------------------------
\end{tabularx}}

\newpage
\section*{BP \#41}
%==============================================================================
% BP #41
%       Samples: 299669 Runs: 6 Burn-in: 100000
%       Discarded rules with mistakes: 331
%       p = proportion of samples for this rule
%==============================================================================
\begin{minipage}[t]{0.4\textwidth}
\vspace{0pt}
\includegraphics[width=0.95\textwidth]{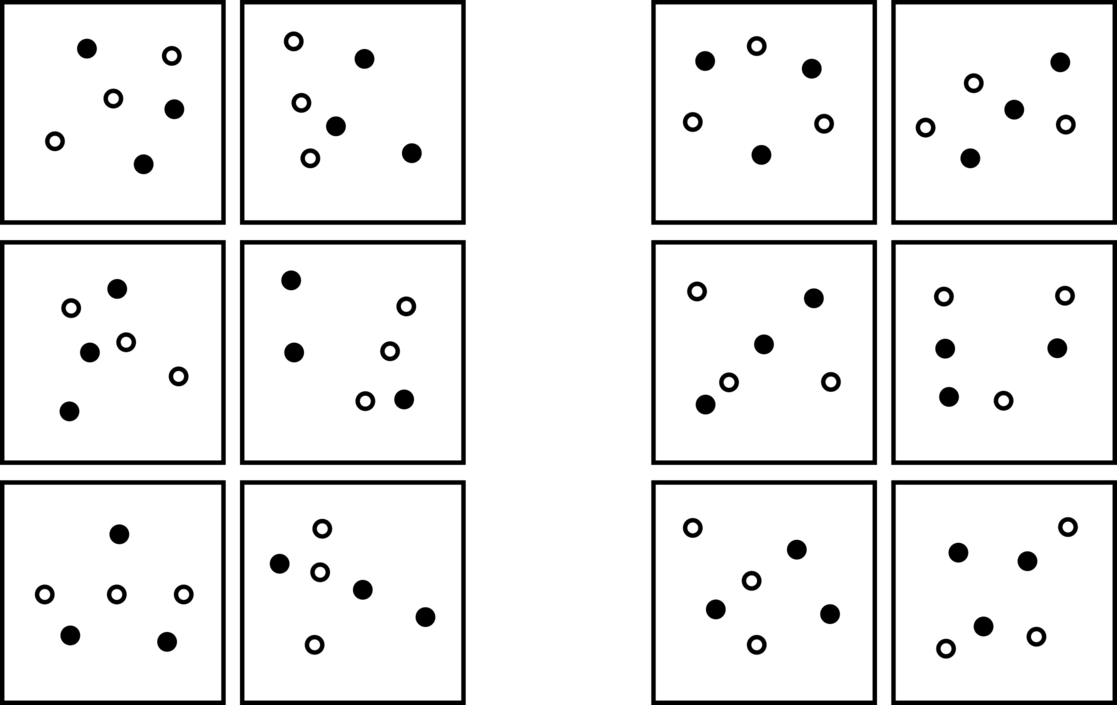}
\end{minipage}
\hfill
\begin{minipage}[t]{0.5\textwidth}
\vspace{0pt}
Samples: 299669\\
Runs: 6\\
Burn-in: 100000\\
Discarded rules with mistakes: 331\\
$p$ = proportion of samples for this rule\\
\end{minipage}

\bigskip

{\noindent\scriptsize
\begin{tabularx}{\textwidth}{Xr}
\hline%------------------------------------------------------------------------
$\LEFT$ outline circles on a straight line                             & $p$ \\
\hline%------------------------------------------------------------------------
$\EXISTS(\ALIGNED(\OUTLINE(\FIGURES)))$                                & .11 \\
$\EXISTS(\ALIGNED(\OUTLINE(\CIRCLES)))$                                & .11 \\
$\EXISTS(\ALIGNED(\GET(\CIRCLES,\HOLES)))$                             & .05 \\
$\EXISTS(\ALIGNED(\GET(\FIGURES,\HOLES)))$                             & .05 \\
$\EXACTLY(3,\ALIGNED(\OUTLINE(\FIGURES)))$                             & .03 \\
\hline%------------------------------------------------------------------------
$\RIGHT$ outline circles not on a straight line                        & $p$ \\
\hline%------------------------------------------------------------------------
$\EXISTS(\OUTLINE(\SETMINUS(\CIRCLES,\ALIGNED(\OUTLINE(\CIRCLES)))))$  & .00 \\
$\EXISTS(\OUTLINE(\SETMINUS(\FIGURES,\ALIGNED(\OUTLINE(\FIGURES)))))$  & .00 \\
$\EQUALNUM(\CIRCLES,\CUP(\CIRCLES,\ALIGNED(\GET(\CIRCLES,\HOLES))))$   & .00 \\
$\MORE(\OUTLINE(\CIRCLES),\ALIGNED(\OUTLINE(\CIRCLES)))$               & .00 \\
$\EQUALNUM(\FIGURES,\SETMINUS(\FIGURES,\ALIGNED(\OUTLINE(\FIGURES))))$ & .00 \\
\hline%------------------------------------------------------------------------
Remaining rules                                                        & .63 \\
\hline%------------------------------------------------------------------------
\end{tabularx}}

\newpage
\section*{BP \#42}
%================================================================
% BP #42
%       Samples: 294199 Runs: 6 Burn-in: 100000
%       Discarded rules with mistakes: 5801
%       p = proportion of samples for this rule
%================================================================
\begin{minipage}[t]{0.4\textwidth}
\vspace{0pt}
\includegraphics[width=0.95\textwidth]{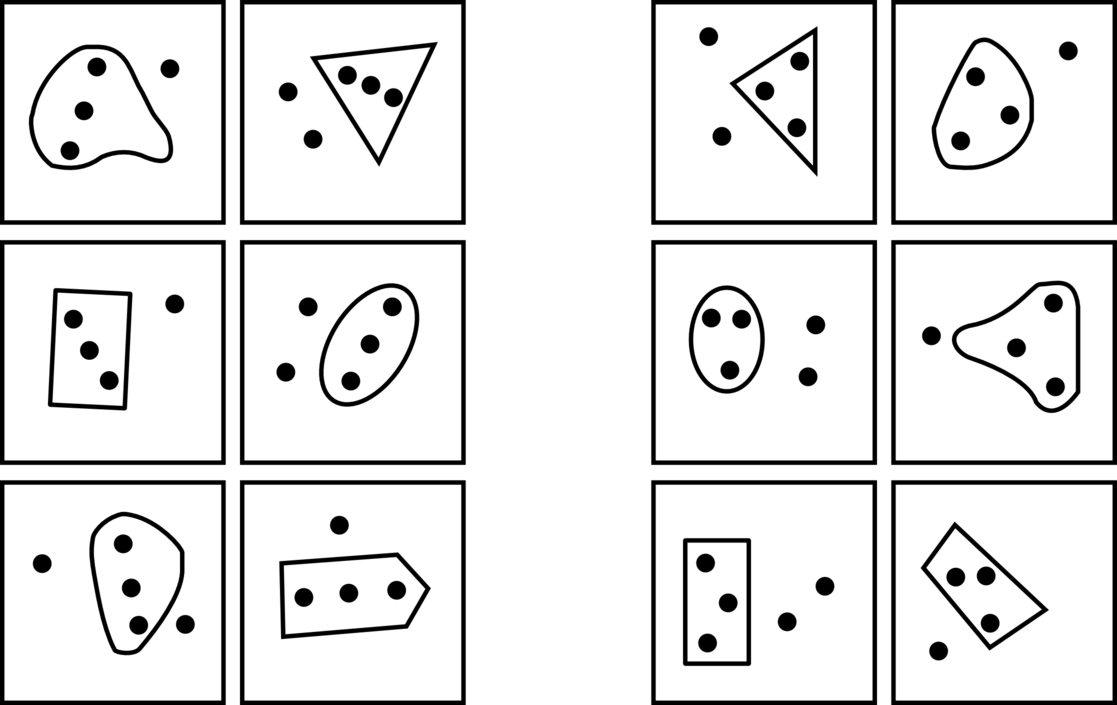}
\end{minipage}
\hfill
\begin{minipage}[t]{0.5\textwidth}
\vspace{0pt}
Samples: 294199\\
Runs: 6\\
Burn-in: 100000\\
Discarded rules with mistakes: 5801\\
$p$ = proportion of samples for this rule\\
\end{minipage}

\bigskip

{\noindent\scriptsize
\begin{tabularx}{\textwidth}{Xr}
\hline%----------------------------------------------------------
$\LEFT$ points inside the figure outline are on a straight line & $p$ \\
\hline%----------------------------------------------------------
$\EXISTS(\ALIGNED(\INSIDE(\FIGURES)))$                   & .22 \\
$\EXACTLY(3,\ALIGNED(\INSIDE(\FIGURES)))$                & .08 \\
$\EXISTS(\ALIGNED(\ALIGNED(\INSIDE(\FIGURES))))$         & .02 \\
$\EXISTS(\ALIGNED(\SOLID(\INSIDE(\FIGURES))))$           & .01 \\
$\EXISTS(\SOLID(\ALIGNED(\INSIDE(\FIGURES))))$           & .01 \\
\hline%----------------------------------------------------------
$\RIGHT$ points inside the figure outline are not on a straight line & $p$ \\
\hline%----------------------------------------------------------
$\MORE(\INSIDE(\FIGURES),\ALIGNED(\INSIDE(\FIGURES)))$   & .01 \\
$\MORE(\BIG(\FIGURES),\ALIGNED(\INSIDE(\FIGURES)))$      & .00 \\
$\MORE(\OUTLINE(\FIGURES),\ALIGNED(\INSIDE(\FIGURES)))$  & .00 \\
$\MORE(\CONTAINS(\FIGURES),\ALIGNED(\INSIDE(\FIGURES)))$ & .00 \\
$\MORE(\CONTAINS(\CIRCLES),\ALIGNED(\INSIDE(\FIGURES)))$ & .00 \\
\hline%----------------------------------------------------------
Remaining rules                                          & .64 \\
\hline%----------------------------------------------------------
\end{tabularx}}

\newpage
\section*{BP \#47}
%==================================================
% BP #47
%       Samples: 298539 Runs: 6 Burn-in: 100000
%       Discarded rules with mistakes: 1463
%       p = proportion of samples for this rule
%==================================================
\begin{minipage}[t]{0.4\textwidth}
\vspace{0pt}
\includegraphics[width=0.95\textwidth]{bpimgs/lowres-p0047}
\end{minipage}
\hfill
\begin{minipage}[t]{0.5\textwidth}
\vspace{0pt}
Samples: 298539\\
Runs: 6\\
Burn-in: 100000\\
Discarded rules with mistakes: 1463\\
$p$ = proportion of samples for this rule\\
\end{minipage}

\bigskip

{\noindent\scriptsize
\begin{tabularx}{\textwidth}{Xr}
\hline%--------------------------------------------
$\LEFT$ triangle inside of the circle      & $p$ \\
\hline%--------------------------------------------
$\EXISTS(\CONTAINS(\TRIANGLES))$           & .10 \\
$\EXISTS(\INSIDE(\CIRCLES))$               & .10 \\
$\EXACTLY(1,\INSIDE(\CIRCLES))$            & .03 \\
$\EXACTLY(1,\CONTAINS(\TRIANGLES))$        & .03 \\
$\EXISTS(\OUTLINE(\CONTAINS(\TRIANGLES)))$ & .01 \\
\hline%--------------------------------------------
$\RIGHT$ circle inside of the triangle     & $p$ \\
\hline%--------------------------------------------
$\EXISTS(\CONTAINS(\CIRCLES))$             & .12 \\
$\EXISTS(\INSIDE(\TRIANGLES))$             & .12 \\
$\EXACTLY(1,\INSIDE(\TRIANGLES))$          & .03 \\
$\EXACTLY(1,\CONTAINS(\CIRCLES))$          & .03 \\
$\EXISTS(\CONTAINS(\INSIDE(\TRIANGLES)))$  & .01 \\
\hline%--------------------------------------------
Remaining rules                            & .43 \\
\hline%--------------------------------------------
\end{tabularx}}

\newpage
\section*{BP \#48}
%======================================================================================================================
% BP #48
%       Samples: 267885 Runs: 8 Burn-in: 100000
%       Discarded rules with mistakes: 132118
%       p = proportion of samples for this rule
%======================================================================================================================
\begin{minipage}[t]{0.4\textwidth}
\vspace{0pt}
\includegraphics[width=0.95\textwidth]{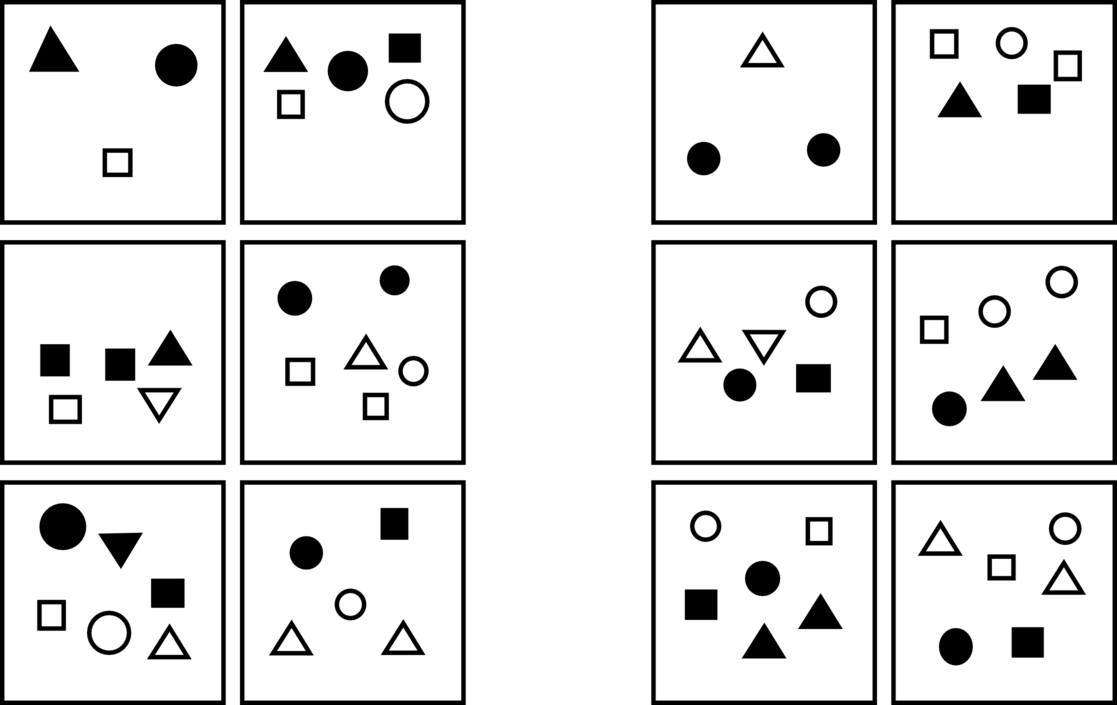}
\end{minipage}
\hfill
\begin{minipage}[t]{0.5\textwidth}
\vspace{0pt}
Samples: 267885\\
Runs: 8\\
Burn-in: 100000\\
Discarded rules with mistakes: 132118\\
$p$ = proportion of samples for this rule\\
\end{minipage}

\bigskip

{\noindent\scriptsize
\begin{tabularx}{\textwidth}{Xr}
\hline%----------------------------------------------------------------------------------------------------------------
$\LEFT$ solid dark figures above the outline figures                                                           & $p$ \\
\hline%----------------------------------------------------------------------------------------------------------------
$\GREATERLA(\GET(\SMALL(\FIGURES),\HOLES),\YPOS)$                                                              & .10 \\
$\GREATERLLA(\OUTLINE(\FIGURES),\SOLID(\FIGURES),\YPOS)$                                                       & .10 \\
$\GREATERLA(\SMALL(\GET(\FIGURES,\HOLES)),\YPOS)$                                                              & .08 \\
$\GREATERLLA(\GET(\FIGURES,\HOLES),\SOLID(\FIGURES),\YPOS)$                                                    & .05 \\
$\GREATERLA(\OUTLINE(\SMALL(\FIGURES)),\YPOS)$                                                                 & .02 \\
\hline%----------------------------------------------------------------------------------------------------------------
$\RIGHT$ outline figures above the solid dark figures                                                          & $p$ \\
\hline%----------------------------------------------------------------------------------------------------------------
$\MORE(\GET(\LOW(\TRIANGLES,\SIZE),\HULLS),\BIG(\CUP(\FIGURES,\BIG(\BIG(\GET(\TRIANGLES,\HOLES))))))$          & .00 \\
$\MORE(\GET(\GET(\TRIANGLES,\HULLS),\HULLS),\BIG(\CUP(\FIGURES,\BIG(\BIG(\GET(\TRIANGLES,\HOLES))))))$         & .00 \\
$\MORE(\GET(\GET(\TRIANGLES,\HULLS),\HOLES),\BIG(\CUP(\FIGURES,\BIG(\BIG(\GET(\TRIANGLES,\HOLES))))))$         & .00 \\
$\MORE(\GET(\TRIANGLES,\HULLS),\BIG(\CUP(\FIGURES,\BIG(\BIG(\GET(\TRIANGLES,\HOLES))))))$                      & .00 \\
$\MORE(\GET(\GET(\TRIANGLES,\HULLS),\HULLS),\BIG(\CUP(\FIGURES,\BIG(\BIG(\GET(\SMALL(\TRIANGLES),\HOLES))))))$ & .00 \\
\hline%----------------------------------------------------------------------------------------------------------------
Remaining rules                                                                                                & .65 \\
\hline%----------------------------------------------------------------------------------------------------------------
\end{tabularx}}

\newpage
\section*{BP \#49}
%==================================================
% BP #49
%       Samples: 298979 Runs: 6 Burn-in: 100000
%       Discarded rules with mistakes: 1022
%       p = proportion of samples for this rule
%==================================================
\begin{minipage}[t]{0.4\textwidth}
\vspace{0pt}
\includegraphics[width=0.95\textwidth]{bpimgs/lowres-p0049}
\end{minipage}
\hfill
\begin{minipage}[t]{0.5\textwidth}
\vspace{0pt}
Samples: 298979\\
Runs: 6\\
Burn-in: 100000\\
Discarded rules with mistakes: 1022\\
$p$ = proportion of samples for this rule\\
\end{minipage}

\bigskip

{\noindent\scriptsize
\begin{tabularx}{\textwidth}{Xr}
\hline%--------------------------------------------
$\LEFT$ points inside the figure outline are grouped more densely than outside the contour & $p$ \\
\hline%--------------------------------------------
\hline%--------------------------------------------
$\RIGHT$ points outside the figure outline are grouped more densely than inside the contour                                  & $p$ \\
\hline%--------------------------------------------
$\MORESIMLA(\FIGURES,\DISTANCE)$           & .40 \\
$\MORESIMLA(\CIRCLES,\DISTANCE)$           & .29 \\
$\GREATERLA(\INSIDE(\FIGURES),\DISTANCE)$  & .02 \\
$\MORESIMLA(\OUTLINE(\FIGURES),\DISTANCE)$ & .02 \\
$\MORESIMLA(\SMALL(\CIRCLES),\DISTANCE)$   & .02 \\
\hline%--------------------------------------------
Remaining rules                            & .25 \\
\hline%--------------------------------------------
\end{tabularx}}

\newpage
\section*{BP \#51}
%========================================================
% BP #51
%       Samples: 293581 Runs: 6 Burn-in: 100000
%       Discarded rules with mistakes: 6421
%       p = proportion of samples for this rule
%========================================================
\begin{minipage}[t]{0.4\textwidth}
\vspace{0pt}
\includegraphics[width=0.95\textwidth]{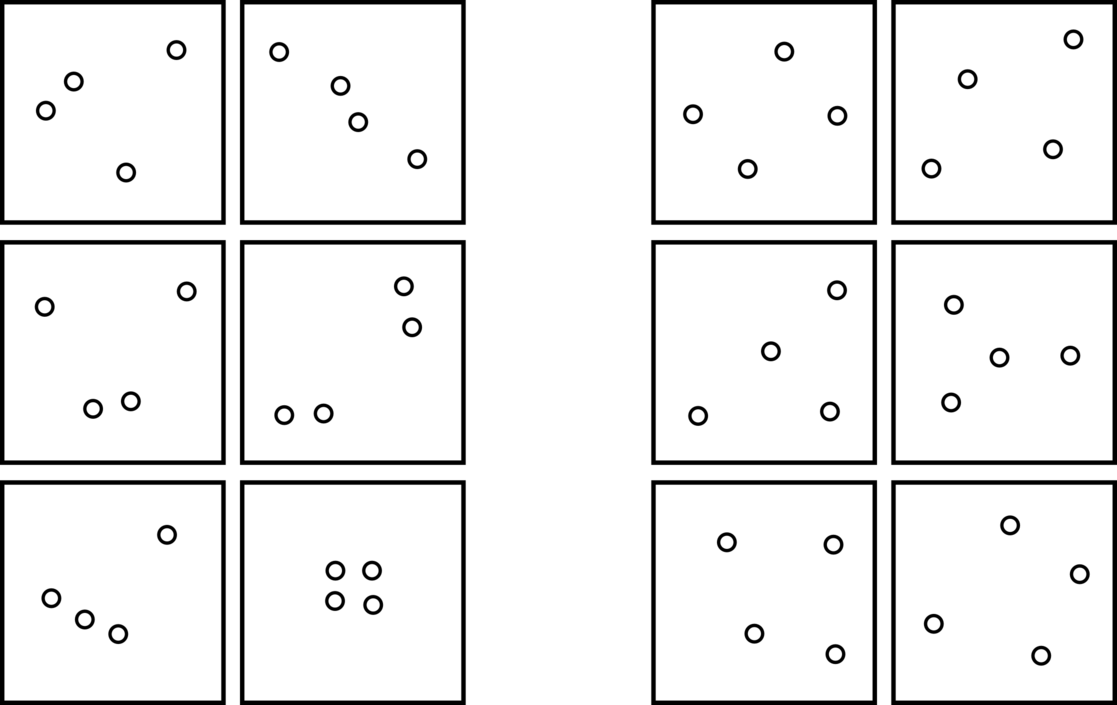}
\end{minipage}
\hfill
\begin{minipage}[t]{0.5\textwidth}
\vspace{0pt}
Samples: 293581\\
Runs: 6\\
Burn-in: 100000\\
Discarded rules with mistakes: 6421\\
$p$ = proportion of samples for this rule\\
\end{minipage}

\bigskip

{\noindent\scriptsize
\begin{tabularx}{\textwidth}{Xr}
\hline%--------------------------------------------------
$\LEFT$ two circles close to each other          & $p$ \\
\hline%--------------------------------------------------
$\EXISTS(\LOW(\FIGURES,\DISTANCE))$              & .13 \\
$\EXISTS(\LOW(\CIRCLES,\DISTANCE))$              & .13 \\
$\EXISTS(\LOW(\GET(\FIGURES,\HOLES),\DISTANCE))$ & .03 \\
$\EXISTS(\LOW(\GET(\CIRCLES,\HOLES),\DISTANCE))$ & .03 \\
$\EXISTS(\LOW(\GET(\FIGURES,\HULLS),\DISTANCE))$ & .03 \\
\hline%--------------------------------------------------
$\RIGHT$ no two circles close to each other      & $p$ \\
\hline%--------------------------------------------------
$\EXACTLY(4,\HIGH(\CIRCLES,\DISTANCE))$          & .05 \\
$\EXACTLY(4,\HIGH(\FIGURES,\DISTANCE))$          & .05 \\
$\EQUALNUM(\HIGH(\FIGURES,\DISTANCE),\FIGURES)$  & .03 \\
$\EQUALNUM(\HIGH(\CIRCLES,\DISTANCE),\CIRCLES)$  & .03 \\
$\EQUALNUM(\CIRCLES,\HIGH(\CIRCLES,\DISTANCE))$  & .02 \\
\hline%--------------------------------------------------
Remaining rules                                  & .49 \\
\hline%--------------------------------------------------
\end{tabularx}}

\newpage
\section*{BP \#53}
%=======================================================================
% BP #53
%       Samples: 116809 Runs: 6 Burn-in: 100000
%       Discarded rules with mistakes: 183192
%       p = proportion of samples for this rule
%=======================================================================
\begin{minipage}[t]{0.4\textwidth}
\vspace{0pt}
\includegraphics[width=0.95\textwidth]{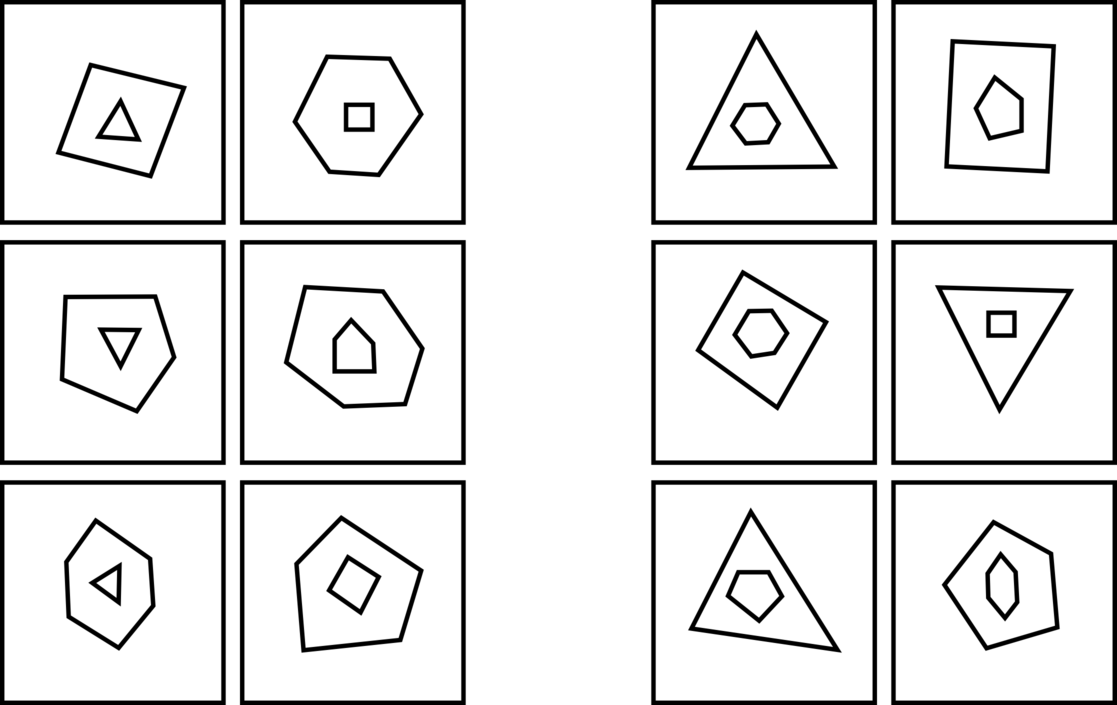}
\end{minipage}
\hfill
\begin{minipage}[t]{0.5\textwidth}
\vspace{0pt}
Samples: 116809\\
Runs: 6\\
Burn-in: 100000\\
Discarded rules with mistakes: 183192\\
$p$ = proportion of samples for this rule\\
\end{minipage}

\bigskip

{\noindent\scriptsize
\begin{tabularx}{\textwidth}{Xr}
\hline%-----------------------------------------------------------------
$\LEFT$ inside figure has fewer angles than outside figure      & $p$ \\
\hline%-----------------------------------------------------------------
$\GREATERLLA(\BIG(\FIGURES),\INSIDE(\FIGURES),\NCORNERS)$       & .08 \\
$\GREATERLLA(\BIG(\FIGURES),\SMALL(\FIGURES),\NCORNERS)$        & .08 \\
$\GREATERLLA(\CONTAINS(\FIGURES),\INSIDE(\FIGURES),\NCORNERS)$  & .07 \\
$\GREATERLLA(\CONTAINS(\FIGURES),\SMALL(\FIGURES),\NCORNERS)$   & .07 \\
$\GREATERLA(\GET(\INSIDE(\FIGURES),\HULLS),\XPOS)$              & .03 \\
\hline%-----------------------------------------------------------------
$\RIGHT$ inside figure has more angles than outside figure      & $p$ \\
\hline%-----------------------------------------------------------------
$\GREATERLLA(\SMALL(\FIGURES),\BIG(\FIGURES),\NCORNERS)$        & .02 \\
$\GREATERLLA(\INSIDE(\FIGURES),\BIG(\FIGURES),\NCORNERS)$       & .01 \\
$\GREATERLLA(\INSIDE(\FIGURES),\CONTAINS(\FIGURES),\NCORNERS)$  & .01 \\
$\GREATERLLA(\SMALL(\FIGURES),\CONTAINS(\FIGURES),\NCORNERS)$   & .01 \\
$\GREATERLLA(\SMALL(\FIGURES),\HIGH(\FIGURES,\SIZE),\NCORNERS)$ & .00 \\
\hline%-----------------------------------------------------------------
Remaining rules                                                 & .62 \\
\hline%-----------------------------------------------------------------
\end{tabularx}}

\newpage
\section*{BP \#56}
%======================================================
% BP #56
%       Samples: 300001 Runs: 6 Burn-in: 100000
%       Discarded rules with mistakes: 0
%       p = proportion of samples for this rule
%======================================================
\begin{minipage}[t]{0.4\textwidth}
\vspace{0pt}
\includegraphics[width=0.95\textwidth]{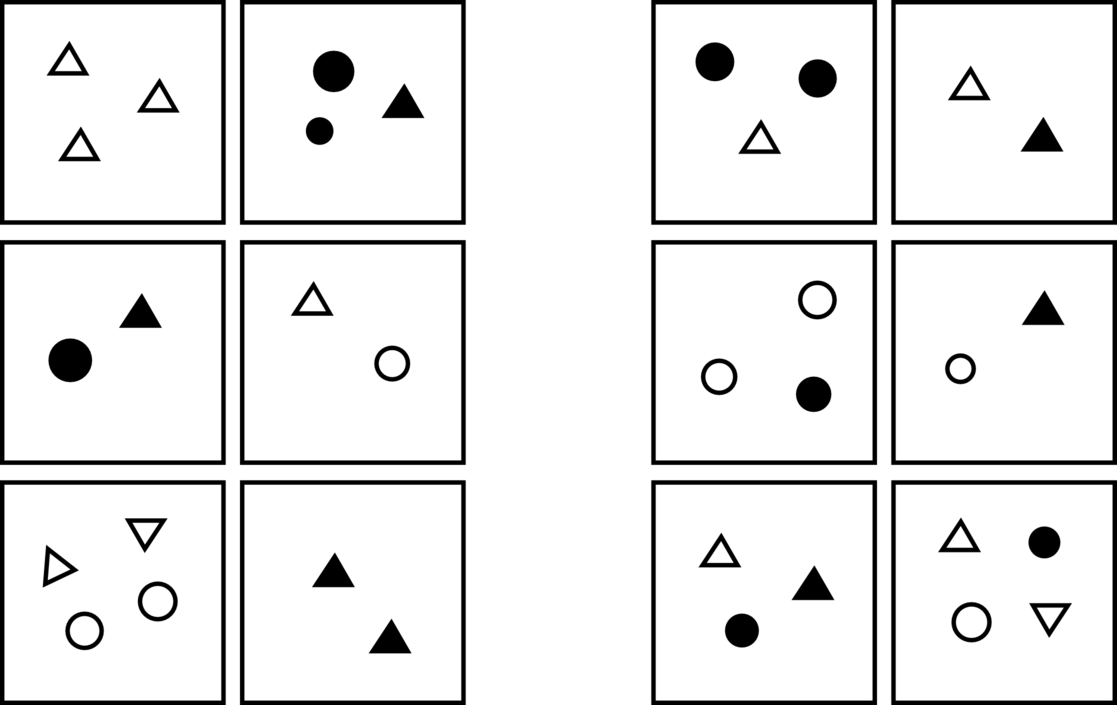}
\end{minipage}
\hfill
\begin{minipage}[t]{0.5\textwidth}
\vspace{0pt}
Samples: 300001\\
Runs: 6\\
Burn-in: 100000\\
Discarded rules with mistakes: 0\\
$p$ = proportion of samples for this rule\\
\end{minipage}

\bigskip

{\noindent\scriptsize
\begin{tabularx}{\textwidth}{Xr}
\hline%------------------------------------------------
$\LEFT$ all figures of the same color          & $p$ \\
\hline%------------------------------------------------
$\MORESIMLA(\FIGURES,\COLOR)$                  & .93 \\
$\MORESIMLA(\BIG(\FIGURES),\COLOR)$            & .02 \\
$\MORESIMLA(\CUP(\FIGURES,\FIGURES),\COLOR)$   & .01 \\
$\MORESIMLA(\CAP(\FIGURES,\FIGURES),\COLOR)$   & .00 \\
$\MORESIMLA(\CUP(\TRIANGLES,\FIGURES),\COLOR)$ & .00 \\
\hline%------------------------------------------------
$\RIGHT$ figures of different colors           & $p$ \\
\hline%------------------------------------------------
\hline%------------------------------------------------
Remaining rules                                & .05 \\
\hline%------------------------------------------------
\end{tabularx}}

\newpage
\section*{BP \#58}
%=======================================================
% BP #58
%       Samples: 298164 Runs: 6 Burn-in: 100000
%       Discarded rules with mistakes: 1836
%       p = proportion of samples for this rule
%=======================================================
\begin{minipage}[t]{0.4\textwidth}
\vspace{0pt}
\includegraphics[width=0.95\textwidth]{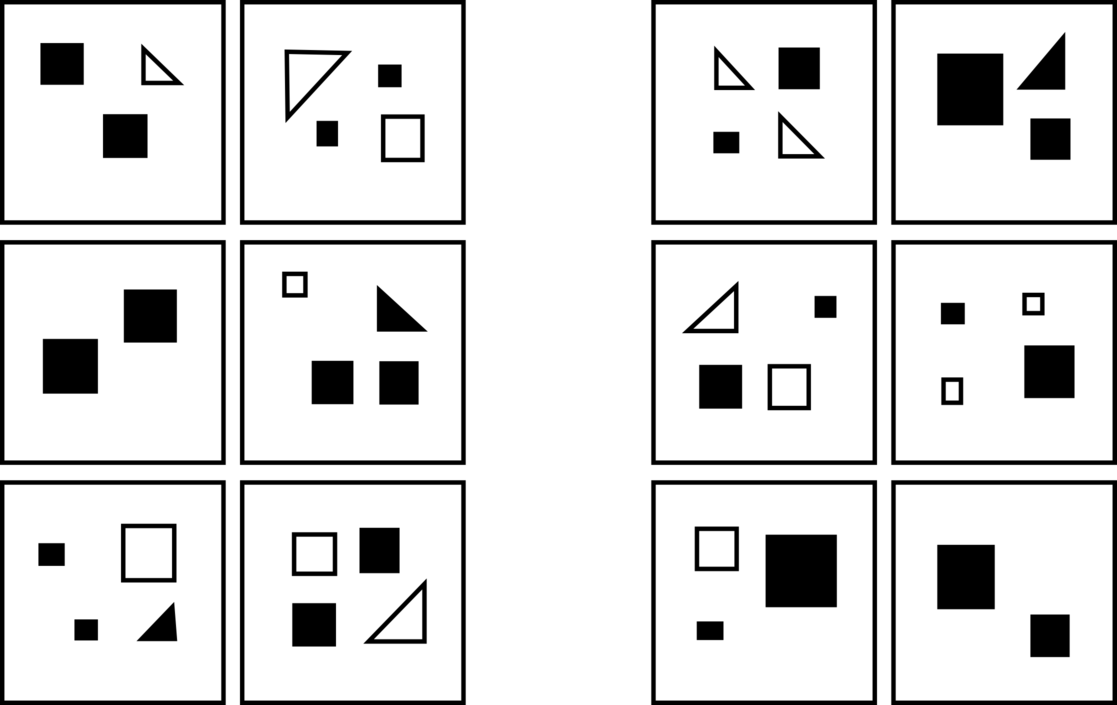}
\end{minipage}
\hfill
\begin{minipage}[t]{0.5\textwidth}
\vspace{0pt}
Samples: 298164\\
Runs: 6\\
Burn-in: 100000\\
Discarded rules with mistakes: 1836\\
$p$ = proportion of samples for this rule\\
\end{minipage}

\bigskip

{\noindent\scriptsize
\begin{tabularx}{\textwidth}{Xr}
\hline%-------------------------------------------------
$\LEFT$ solid dark quadrangles are identical    & $p$ \\
\hline%-------------------------------------------------
$\MORESIMLA(\SOLID(\FIGURES),\SIZE)$            & .37 \\
$\MORESIMLA(\SOLID(\RECTANGLES),\SIZE)$         & .37 \\
$\MORESIMLA(\SOLID(\SOLID(\RECTANGLES)),\SIZE)$ & .04 \\
$\MORESIMLA(\SOLID(\SOLID(\FIGURES)),\SIZE)$    & .04 \\
$\MORESIMLA(\HIGH(\FIGURES,\COLOR),\SIZE)$      & .01 \\
\hline%-------------------------------------------------
$\RIGHT$ solid dark quadrangles are different   & $p$ \\
\hline%-------------------------------------------------
\hline%-------------------------------------------------
Remaining rules                                 & .16 \\
\hline%-------------------------------------------------
\end{tabularx}}

\newpage
\section*{BP \#65}
%===================================================
% BP #65
%       Samples: 299856 Runs: 6 Burn-in: 100000
%       Discarded rules with mistakes: 145
%       p = proportion of samples for this rule
%===================================================
\begin{minipage}[t]{0.4\textwidth}
\vspace{0pt}
\includegraphics[width=0.95\textwidth]{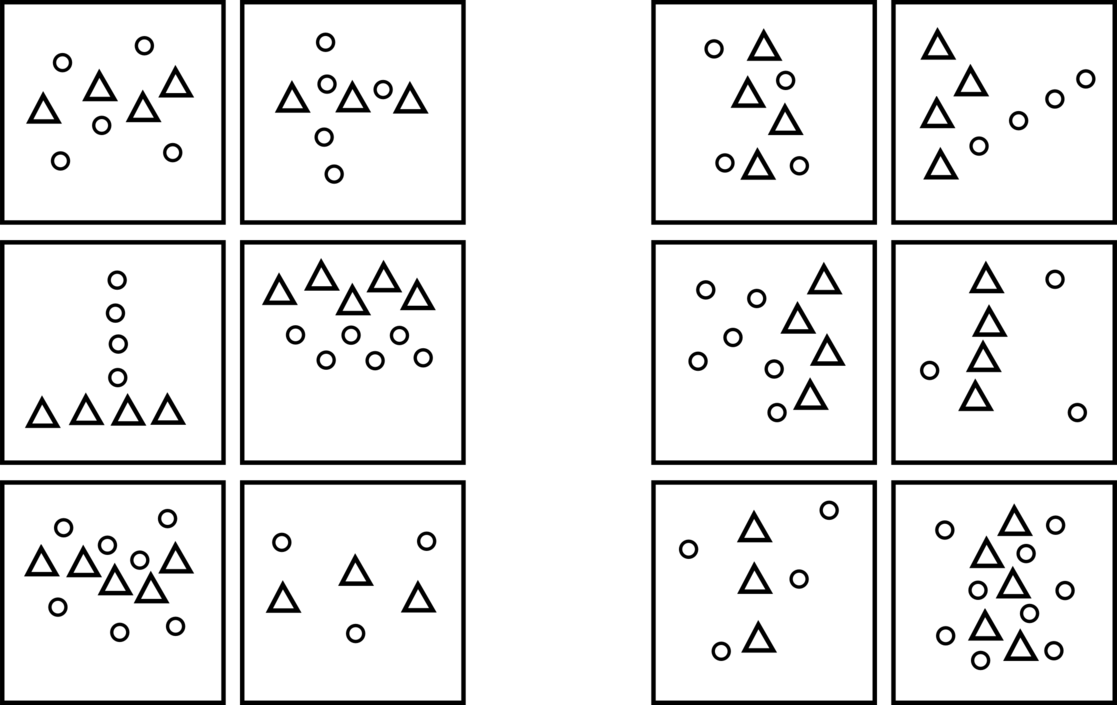}
\end{minipage}
\hfill
\begin{minipage}[t]{0.5\textwidth}
\vspace{0pt}
Samples: 299856\\
Runs: 6\\
Burn-in: 100000\\
Discarded rules with mistakes: 145\\
$p$ = proportion of samples for this rule\\
\end{minipage}

\bigskip

{\noindent\scriptsize
\begin{tabularx}{\textwidth}{Xr}
\hline%---------------------------------------------
$\LEFT$ a set of triangles elongated horizontally & $p$ \\
\hline%---------------------------------------------
$\MORESIMLA(\TRIANGLES,\YPOS)$              & .40 \\
$\MORESIMLA(\BIG(\FIGURES),\YPOS)$          & .02 \\
$\MORESIMLA(\OUTLINE(\TRIANGLES),\YPOS)$    & .02 \\
$\MORESIMLA(\GET(\TRIANGLES,\HULLS),\YPOS)$ & .01 \\
$\MORESIMLA(\GET(\TRIANGLES,\HOLES),\YPOS)$ & .01 \\
\hline%---------------------------------------------
$\RIGHT$ a set of triangles elongated vertically & $p$ \\
\hline%---------------------------------------------
$\MORESIMLA(\TRIANGLES,\XPOS)$              & .36 \\
$\MORESIMLA(\OUTLINE(\TRIANGLES),\XPOS)$    & .02 \\
$\MORESIMLA(\BIG(\FIGURES),\XPOS)$          & .02 \\
$\MORESIMLA(\GET(\TRIANGLES,\HULLS),\XPOS)$ & .01 \\
$\MORESIMLA(\GET(\TRIANGLES,\HOLES),\XPOS)$ & .01 \\
\hline%---------------------------------------------
Remaining rules                             & .11 \\
\hline%---------------------------------------------
\end{tabularx}}

\newpage
\section*{BP \#66}
%==============================================
% BP #66
%       Samples: 298393 Runs: 6 Burn-in: 100000
%       Discarded rules with mistakes: 1609
%       p = proportion of samples for this rule
%==============================================
\begin{minipage}[t]{0.4\textwidth}
\vspace{0pt}
\includegraphics[width=0.95\textwidth]{bpimgs/lowres-p0066}
\end{minipage}
\hfill
\begin{minipage}[t]{0.5\textwidth}
\vspace{0pt}
Samples: 298393\\
Runs: 6\\
Burn-in: 100000\\
Discarded rules with mistakes: 1609\\
$p$ = proportion of samples for this rule\\
\end{minipage}

\bigskip

{\noindent\scriptsize
\begin{tabularx}{\textwidth}{Xr}
\hline%----------------------------------------
$\LEFT$ unconnected circles on horizontal line & $p$ \\
\hline%----------------------------------------
$\MORESIMLA(\FIGURES,\YPOS)$           & .24 \\
$\MORESIMLA(\CIRCLES,\YPOS)$           & .23 \\
$\MORESIMLA(\SMALL(\FIGURES),\YPOS)$   & .01 \\
$\MORESIMLA(\ALIGNED(\FIGURES),\YPOS)$ & .01 \\
$\MORESIMLA(\OUTLINE(\CIRCLES),\YPOS)$ & .01 \\
\hline%----------------------------------------
$\RIGHT$ unconnected circles on vertical line & $p$ \\
\hline%----------------------------------------
$\MORESIMLA(\CIRCLES,\XPOS)$           & .23 \\
$\MORESIMLLA(\CIRCLES,\FIGURES,\XPOS)$ & .01 \\
$\MORESIMLA(\OUTLINE(\CIRCLES),\XPOS)$ & .01 \\
$\MORESIMLA(\ALIGNED(\CIRCLES),\XPOS)$ & .01 \\
$\MORESIMLA(\SMALL(\FIGURES),\XPOS)$   & .01 \\
\hline%----------------------------------------
Remaining rules                        & .21 \\
\hline%----------------------------------------
\end{tabularx}}

\newpage
\section*{BP \#71}
%=========================================================================
% BP #71
%       Samples: 298402 Runs: 6 Burn-in: 100000
%       Discarded rules with mistakes: 1599
%       p = proportion of samples for this rule
%=========================================================================
\begin{minipage}[t]{0.4\textwidth}
\vspace{0pt}
\includegraphics[width=0.95\textwidth]{bpimgs/lowres-p0071}
\end{minipage}
\hfill
\begin{minipage}[t]{0.5\textwidth}
\vspace{0pt}
Samples: 298402\\
Runs: 6\\
Burn-in: 100000\\
Discarded rules with mistakes: 1599\\
$p$ = proportion of samples for this rule\\
\end{minipage}

\bigskip

{\noindent\scriptsize
\begin{tabularx}{\textwidth}{Xr}
\hline%-------------------------------------------------------------------
$\LEFT$ there are inside figures of the second order              & $p$ \\
\hline%-------------------------------------------------------------------
$\EXISTS(\INSIDE(\INSIDE(\FIGURES)))$                             & .20 \\
$\EXISTS(\CONTAINS(\CONTAINS(\FIGURES)))$                         & .17 \\
$\EXACTLY(1,\CONTAINS(\CONTAINS(\FIGURES)))$                      & .05 \\
$\EXACTLY(1,\INSIDE(\INSIDE(\FIGURES)))$                          & .04 \\
$\EXISTS(\INSIDE(\OUTLINE(\INSIDE(\FIGURES))))$                   & .01 \\
\hline%-------------------------------------------------------------------
$\RIGHT$ there are no inside figures of the second order          & $p$ \\
\hline%-------------------------------------------------------------------
$\EXACTLY(1,\LOW(\CONTAINS(\FIGURES),\XPOS))$                     & .00 \\
$\EXACTLY(1,\LOW(\CONTAINS(\SMALL(\FIGURES)),\XPOS))$             & .00 \\
$\EXACTLY(1,\LOW(\CONTAINS(\LOW(\FIGURES,\SIZE)),\XPOS))$         & .00 \\
$\EXACTLY(1,\LOW(\CONTAINS(\OUTLINE(\FIGURES)),\XPOS))$           & .00 \\
$\EXACTLY(1,\LOW(\CONTAINS(\INSIDE(\CONTAINS(\FIGURES))),\XPOS))$ & .00 \\
\hline%-------------------------------------------------------------------
Remaining rules                                                   & .53 \\
\hline%-------------------------------------------------------------------
\end{tabularx}}

\newpage
\section*{BP \#79}
%================================================================================
% BP #79
%       Samples: 74860 Runs: 6 Burn-in: 100000
%       Discarded rules with mistakes: 225140
%       p = proportion of samples for this rule
%================================================================================
\begin{minipage}[t]{0.4\textwidth}
\vspace{0pt}
\includegraphics[width=0.95\textwidth]{bpimgs/lowres-p0079}
\end{minipage}
\hfill
\begin{minipage}[t]{0.5\textwidth}
\vspace{0pt}
Samples: 74860\\
Runs: 6\\
Burn-in: 100000\\
Discarded rules with mistakes: 225140\\
$p$ = proportion of samples for this rule\\
\end{minipage}

\bigskip

{\noindent\scriptsize
\begin{tabularx}{\textwidth}{Xr}
\hline%--------------------------------------------------------------------------
$\LEFT$ a dark circle is closer to the outline circle than to the triangle & $p$ \\
\hline%--------------------------------------------------------------------------
\hline%--------------------------------------------------------------------------
$\RIGHT$ a dark circle is closer to the triangle than to the outline circle & $p$ \\
\hline%--------------------------------------------------------------------------
$\GREATERLLA(\CIRCLES,\CUP(\SOLID(\CIRCLES),\TRIANGLES),\DISTANCE)$      & .12 \\
$\GREATERLLA(\CIRCLES,\CUP(\TRIANGLES,\SOLID(\CIRCLES)),\DISTANCE)$      & .09 \\
$\GREATERLLA(\CIRCLES,\SETMINUS(\FIGURES,\OUTLINE(\CIRCLES)),\DISTANCE)$ & .06 \\
$\GREATERLLA(\CIRCLES,\CUP(\SOLID(\FIGURES),\TRIANGLES),\DISTANCE)$      & .06 \\
$\GREATERLLA(\CIRCLES,\CUP(\TRIANGLES,\SOLID(\FIGURES)),\DISTANCE)$      & .05 \\
\hline%--------------------------------------------------------------------------
Remaining rules                                                          & .61 \\
\hline%--------------------------------------------------------------------------
\end{tabularx}}

%\end{document}

\end{document}